\documentclass{article}

\PassOptionsToPackage{numbers,compress}{natbib}
\usepackage[main, final]{neurips_2026}
\usepackage[utf8]{inputenc} 
\usepackage[T1]{fontenc}    
\usepackage{url}            
\usepackage{booktabs}       
\usepackage{amsfonts}       
\usepackage{nicefrac}       
\usepackage{microtype}      
\usepackage{graphicx}
\usepackage{colortbl}
\usepackage{array}
\usepackage{geometry}
\usepackage{caption}
\usepackage{makecell}
\usepackage{wrapfig}
\usepackage{subcaption} 

\usepackage[dvipsnames]{xcolor}

\usepackage{pifont} 
\newcommand{\cmark}{\ding{51}}
\newcommand{\xmark}{\ding{55}}

\usepackage{wrapfig}
\usepackage{multirow}
\usepackage{multicol}
\usepackage{listings} 
\usepackage{titletoc} 
\usepackage{fancyvrb}
\usepackage{tcolorbox} 

\definecolor{uptri}{RGB}{38,114,168}
\definecolor{dntri}{RGB}{178,34,52}
\newcommand{\up}{\textcolor{uptri}{\ding{115}}}
\newcommand{\dn}{\textcolor{dntri}{\ding{116}}}
\definecolor{lightgray}{rgb}{0.83, 0.83, 0.83}
\definecolor{Gray}{gray}{0.6}
\definecolor{aliceblue}{rgb}{0.94, 0.97, 1.0}
\definecolor{mistyrose}{rgb}{1.0, 0.89, 0.88}
\definecolor{backcolour}{rgb}{0.95,0.95,0.92}

\newcommand{\newpara}[1]{\vspace{-1pt}\noindent\textbf{#1}}

\usepackage[accsupp]{axessibility}  
\usepackage{hyperref}
\usepackage{cleveref}

\crefname{equation}{Eq.}{Eqs.}
\Crefname{equation}{Equation}{Equations}

\crefname{figure}{Fig.}{Figs.}
\Crefname{figure}{Figure}{Figures}

\crefname{table}{Tab.}{Tabs.}
\Crefname{table}{Table}{Tables}

\crefname{section}{Sec.}{Secs.}
\Crefname{section}{Section}{Sections}

\crefname{algorithm}{Alg.}{Algs.}
\Crefname{algorithm}{Algorithm}{Algorithms}

\title{Who Says What: Symbolic Trimodal Binding \\ Mechanisms in Audio-Visual LLMs}

\author{
  Jihoo Jung$^{1}$ \quad
  Youngjoon Jang$^{1,2}$ \quad
  Joon Son Chung$^1$ \\
  $^1$KAIST\quad $^2$VGG, University of Oxford \\
}

\begin{document}

\maketitle

\begin{abstract}
Current Audio-Visual LLMs (AVLLMs) struggle with reasoning over videos featuring multi-speaker dialogues. In such videos, resolving ``who says what'' is crucial, which necessitates trimodal (text-audio-visual) binding. Motivated by these challenges, we systematically investigate how this trimodal binding is achieved in AVLLMs. Specifically, we identify emergent symbolic trimodal binding mechanisms in AVLLMs that utilize modality-specific symbolic variables. By encoding auditory and visual components into symbolic variables-capturing temporal utterance sequences and spatial entity coordinates, respectively-the model establishes cross-modal linking within this abstract space. Crucially, we reveal that when trimodal binding fails, the breakdown predominantly stems from misaligned audio-visual connections. To overcome this bottleneck, we introduce an audio-visual prompting method utilizing an off-the-shelf Active Speaker Detection (ASD) model. By simply overlaying visual bounding boxes on active speakers, this training-free approach yields immediate performance gains across four conversation-centric benchmarks. Moreover, lightweight fine-tuning of fewer than 300 steps on these ASD-prompted-videos extends these gains to three general AV benchmarks, suggesting the generalizability of our method. Our code is available at \url{https://github.com/kaistmm/Who-Says-What}. 

\end{abstract}

\section{Introduction}

Conversation-rich videos, including films, television shows, and daily vlogs, account for a significant fraction of contemporary video data. Comprehending such videos relies on the ability to resolve ``who says what.'' This challenge can be framed as \textit{binding problem}-the ability to associate multiple distinct features that belong to the same entity~\cite{feng2024how, gur-arieh2026mixing}. For instance, resolving what ``the man in the gray shirt'' is saying requires the model to form robust in-context associations by jointly binding the textual description, the corresponding visual subject, and the target acoustic utterance.

The ability of neural networks to solve this binding problem has received substantial attention in the era of Large Language Models (LLMs)~\cite{davies2023discoveringvariablebindingcircuitry, prakash2024finetuning, prakash2025languagemodelsuselookbacks, feng2024how, feng2025monitoring}: in LLMs it involves associating entities with their attributes in lengthy texts, while in Vision-Language Models (VLMs) it poses a cross-modal challenge of aligning textual descriptions with specific visual objects. A key finding is that models utilise content-independent symbolic IDs to track and parse objects~\cite{feng2024how, feng2025monitoring, yang2025emergent, assouel2025visual}: rather than processing associations directly in semantic space, models convert raw content into abstract symbolic representations. These IDs allow features belonging to the same entity to be linked together, thereby enabling the model to retrieve the necessary semantic details. For instance, \cite{izadi2025visual} demonstrates that VLMs represent visual objects using symbolic IDs denoting spatial layout (e.g., left or right), independent of visual traits such as color. These symbolic IDs are invoked when processing textual descriptions to direct attention toward the corresponding spatial locations. 

Despite recent progress in understanding unimodal and bimodal binding in LLMs and VLMs, it remains an open question how Audio-Visual LLMs (AVLLMs) resolve the more complex trimodal (text-audio-visual) binding. This challenge becomes markedly pronounced in conversation-rich, multi-speaker videos, where binding errors translate into speaker-attribution failures, such as assigning an utterance to the wrong visible person.
Given that current AVLLMs struggle with reasoning over such videos~\cite{tang2026d,chen2026diadem,nguyen2025see}, it is critical to uncover the mechanisms underlying trimodal binding. Understanding these mechanisms is essential for diagnosing the exact point of failure: whether the bottleneck stems from text-to-audio/visual grounding, or from direct audio-visual alignment.
\begin{figure}[t]
    \centering
    \captionsetup[subfigure]{aboveskip=-4pt}
    \scalebox{0.95}{%
    \begin{minipage}{\linewidth}
    \centering
    \begin{subfigure}[t]{0.40\textwidth}
        \centering
        \includegraphics[width=\textwidth]{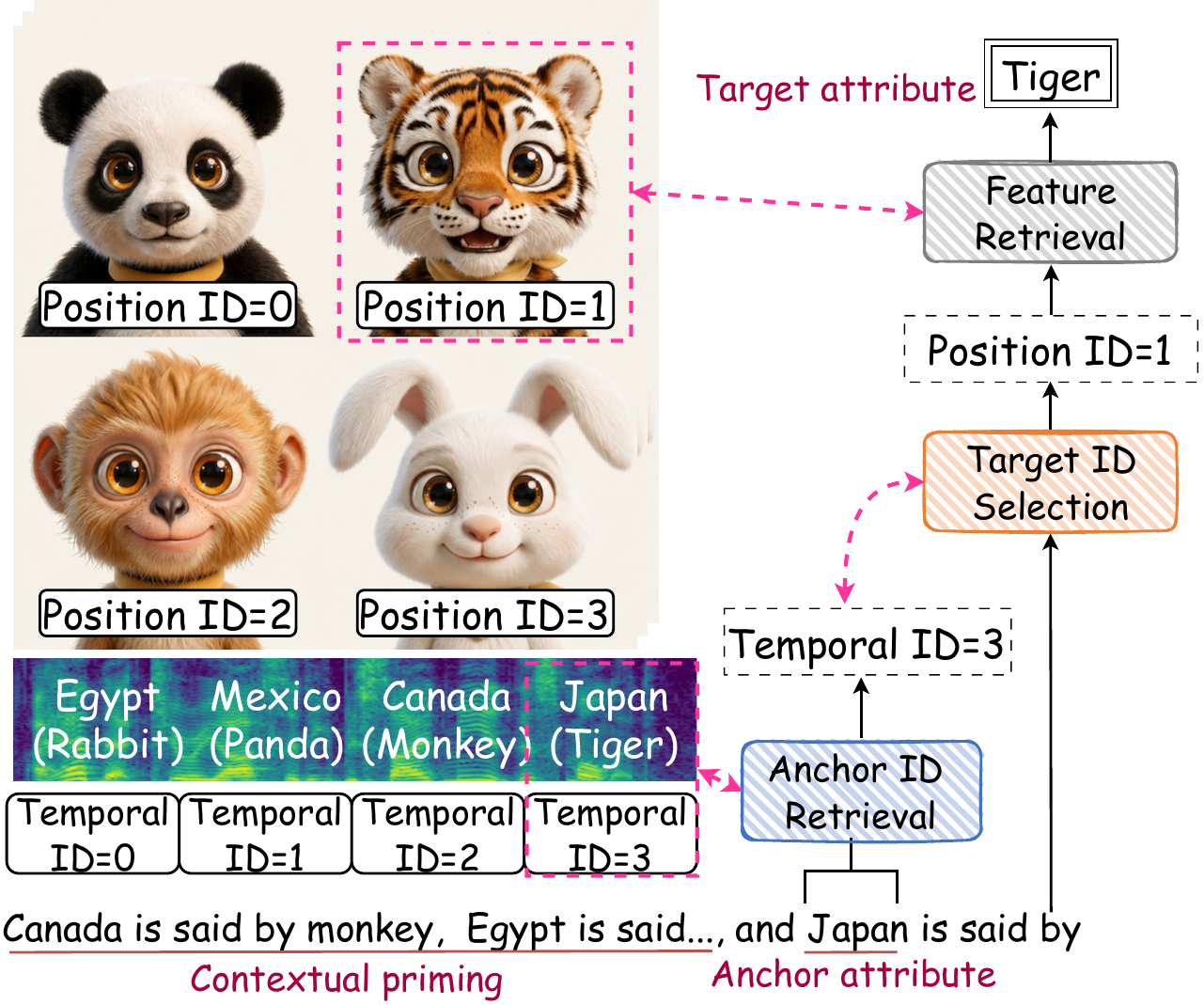}
        \vspace{-0.2cm}
        \caption{Symbolic trimodal binding mechanisms.}
        \label{fig:main_a}
    \end{subfigure}
    \hfill
    \begin{subfigure}[t]{0.58\textwidth}
        \centering
        \includegraphics[width=\textwidth]{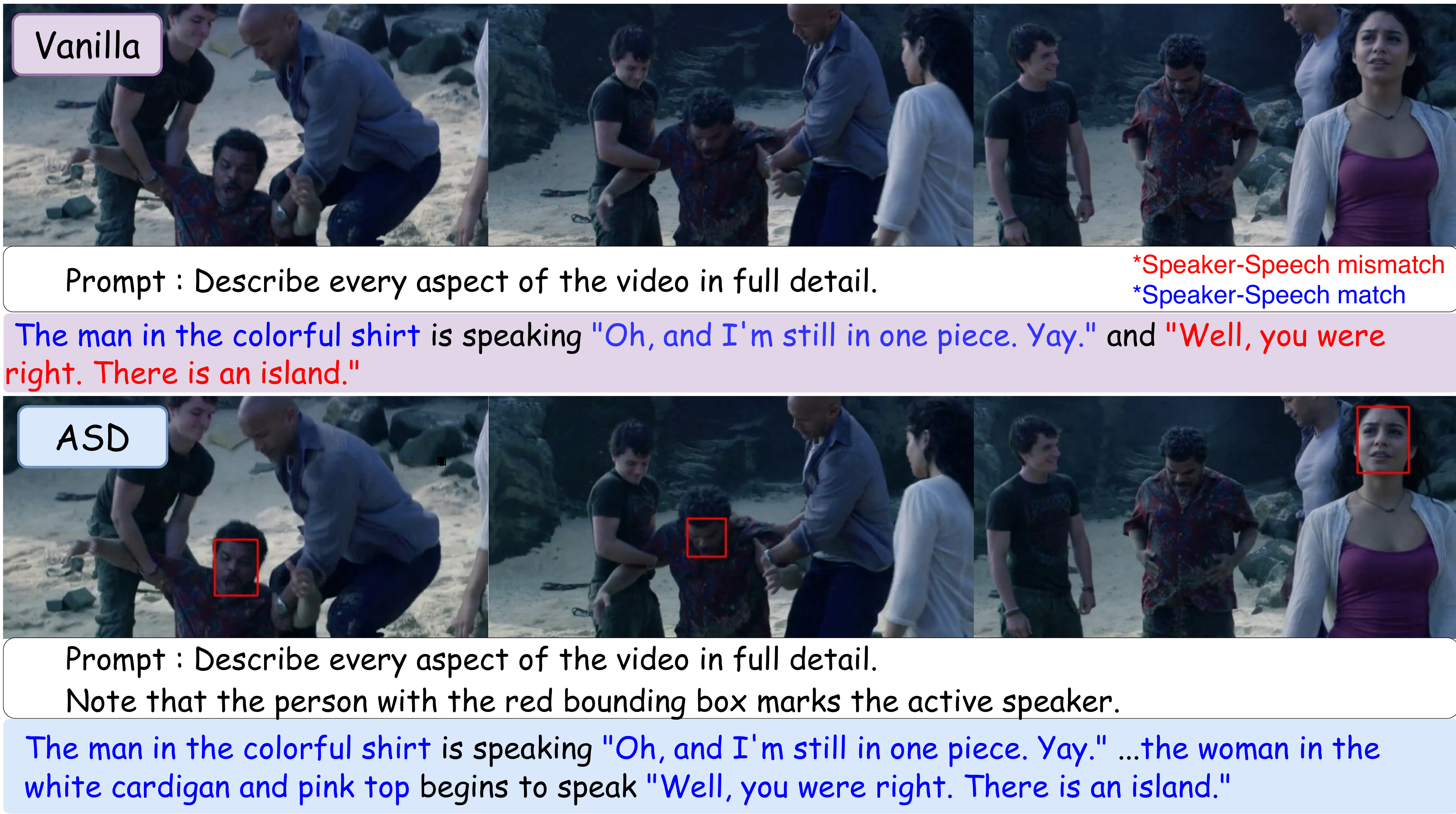}
        \vspace{-0.2cm}
        \caption{Audio-visual prompting with Active Speaker Detection.}
        \label{fig:main_b}
    \end{subfigure}
    \end{minipage}%
    }
    \vspace{-0.1cm}
    \caption{(a) Trimodal binding consists of three stages: anchor ID retrieval, target ID selection, and feature retrieval. (b) The vanilla model fails to accurately associate the speaker to their corresponding utterance, whereas the ASD method successfully establishes this connection.}
    \vspace{-0.5cm}
    \label{fig:main}
\end{figure}

In this work, we investigate the internal mechanisms of trimodal binding in conversation-rich multi-speaker videos using four recent AVLLMs: video-SALMONN2+ (7B)~\cite{tang2025video}, Qwen2.5-Omni (3B, 7B)~\cite{xu2025qwen25omnitechnicalreport}, and MiniCPM-o-4.5 (9B)~\cite{MiniCPM}. We specifically focus on ``single-shot'' videos-where active and inactive speakers appear simultaneously within each frame-as opposed to videos showing only a single active speaker at a time. Such scenarios are of particular interest because the co-occurrence of multiple candidates complicates the binding task, causing AVLLMs to struggle significantly~\cite{chen2026diadem}. To systematically explore this, we build a toy dataset by simulating single-shot multi-speaker videos featuring four animal speakers, each delivering a separate utterance (\cref{fig:main_a}). Using this setup, we formulate a trimodal binding task: given a text prompt describing either an acoustic attribute (e.g., a spoken utterance) or a visual attribute (e.g., animal species) as the \textit{anchor attribute}, the model is required to identify the corresponding \textit{target attribute} in the opposite modality.

Our findings are three-fold. First, we demonstrate that AVLLMs, much like LLMs and VLMs, employ abstract symbolic IDs to resolve the binding task. Crucially, this abstraction varies by modality: acoustic attributes map to temporal IDs encoding when utterances occur, while visual attributes map to spatial IDs encoding where speakers are located. Second, we identify a three-stage trimodal binding mechanism: (i)~anchor ID retrieval, where the text prompt describing an anchor attribute is mapped to its symbolic ID; (ii)~target ID selection, where this ID is mapped to the symbolic ID of the target attribute; and (iii)~feature retrieval, where the model extracts semantic features linked to the target symbolic ID. Finally, we show that binding failures primarily stem from the target ID selection stage, exposing a fundamental deficit in the audio-visual alignment of current AVLLMs.

Based on these findings, to compensate for the identified deficits, we propose a simple yet effective audio-visual prompting method using an off-the-shelf Active Speaker Detection (ASD) model. By leveraging the ASD model to overlay bounding boxes on active speakers before feeding the video into the AVLLMs (\cref{fig:main_b}), this training-free approach yields immediate performance gains across three conversation-centric benchmarks on Qwen2.5-Omni and MiniCPM-o-4.5. Furthermore, lightweight fine-tuning of fewer than 300 steps on these ASD-prompted-videos leads to further gains. Specifically, this tuning improves accuracy on conversation-centric datasets by 7.9\% (video-SALMONN2+), 9.7\% (Qwen2.5-Omni), and 3.5\% (MiniCPM-o-4.5). Notably, our method generalizes beyond conversational settings, consistently improving performance across three additional general benchmarks.

\section{Related Works}

\newpara{Audio-visual large language models.} 
Building upon the foundational capabilities of LLMs, Audio-Visual Large Language Models (AVLLMs) have emerged to expand text-centric reasoning into the realms of audio and visual perception~\cite{zhang2023video, cheng2024videollama2advancingspatialtemporal,chowdhury2024meerkat,lyu2023macaw,ye2024cat,tang2025video,guo2025aligned,xu2025qwen25omnitechnicalreport,xu2025qwen3omnitechnicalreport,ye2026omnivinci}. Within this rapid wave of diverse advancements, video-SALMONN2+~\cite{tang2025video} targets detailed, holistic audio-visual captioning, while Qwen2.5-Omni~\cite{xu2025qwen25omnitechnicalreport} and MiniCPM-o-4.5~\cite{MiniCPM} elevate the field by introducing natively generated text and speech outputs. More recently, Qwen3-Omni~\cite{xu2025qwen3omnitechnicalreport} advances the landscape further through a Mixture-of-Experts structure. Alongside these modeling advances, a growing body of work has begun to analyze how AVLLMs internally process and integrate audio-visual inputs~\cite{jung2026probing, jung2025avcd, yoo2026nature}. However, these models still frequently struggle in intricate, multi-speaker videos~\cite{tang2026d,chen2026diadem,nguyen2025see} that demand precise binding across modalities. Addressing these failures thus requires a closer examination of their internal cross-modal binding mechanisms.

\newpara{Binding mechanisms in LLMs/VLMs.} Recent studies have actively investigated how LLMs associate and bind textual entities with their corresponding attributes in long-context settings~\cite{feng2024how, dai-etal-2024-representational, gur-arieh2026mixing, feng2025monitoring, prakash2025languagemodelsuselookbacks}. 
\cite{feng2024how} first propose the existence of symbolic binding IDs that link entities to their attributes, while \cite{prakash2025languagemodelsuselookbacks} further examine how language models bind each character's beliefs about the state of a given entity.
This line of inquiry has also been extended to VLMs to explore multimodal binding mechanisms~\cite{assouel2025visual, buzeta2026seeing, hasani2025uncovering}. 
Recent works show that VLM failures on basic multi-object reasoning tasks, such as counting and localization, are closely linked to binding failures, and propose methods to mitigate them~\cite{campbell2024understanding, hasani2025uncovering, izadi2025visual}. Building on this, our work extends the scope from unimodal and bimodal contexts to investigate \emph{trimodal} binding-across text, audio, and vision-within AVLLMs.

\newpara{Visual prompting in VLMs.}
Visual prompting~\cite{vp_survey} enhances VLMs by providing fine-grained instructions through markers such as circles~\cite{ViP-LLaVA,SoM} and bounding boxes~\cite{Cityllava, Instructdet, Groma}. Recent works have explored the use of off-the-shelf vision modules to mitigate limitations of VLMs. For instance, to improve spatiotemporal reasoning, \cite{tang2025can} leverage object tracking models~\cite{cheng2024yolo, ravi2025sam}, providing the resulting object bounding-box coordinates as textual prompts. Similarly, to reduce hallucinations, \cite{marine} use object detection models~\cite{carion2020end, zhang2024recognize} as additional guidance. While often applied as training-free prompts~\cite{wu2025number,lee2026vikey,izadi2025visual}, these techniques can also be paired with fine-tuning to enhance the model's understanding of the augmented prompts~\cite{wu2025number, tang2025can, lin2024rethinking}. Inspired by these approaches, we introduce an audio-visual prompting method that leverages an off-the-shelf active speaker detection model to address the identified limitations of AVLLMs under both training-free and fine-tuning settings.
\section{Symbolic Trimodal Binding Mechanisms in AVLLMs}
\label{sec:symbolic}
In this section, we define the trimodal binding tasks (\cref{sec:binding_task}) and propose a three-stage symbolic mechanism (\cref{sec:symbolic_mech}), validated through representational (\cref{sec:rsa}) and causal mediation analyses (\cref{sec:cma}). We study four AVLLMs: video-SALMONN2+ (7B)~\cite{tang2025video}, Qwen2.5-Omni (7B, 3B)~\cite{xu2025qwen25omnitechnicalreport}, and MiniCPM-o-4.5~\cite{MiniCPM}. 
Unless otherwise specified, results are reported for video-SALMONN2+ in this section; results for other models are provided in \cref{app:rep} of the supp.\ mat.

\subsection{Trimodal Binding Task}
\label{sec:binding_task}
Trimodal binding task evaluates an ability to associate textual, acoustic and visual information within single-shot, multi-speaker videos. We define a video as a sequence of $n$ distinct speech events, where each event $i$ is characterized by a 4-tuple $(v_i, p_i, a_i, t_i)$. Specifically, the visual attribute $v_i \in V$ denotes the visual characteristics of the speaker, the spatial position $p_i \in P$ indicates their location on the screen (e.g., top-left, bottom-right etc.), the acoustic attribute $a_i \in A$ represents acoustic features such as the spoken content, and the temporal order $t_i \in T$ denotes the chronological speaking order of the utterance. The complete context of the video, $c$, is thus formulated as a set of bound tuples:
\begin{equation*}
c = \{(v_1, p_1, a_1, t_1), (v_2, p_2, a_2, t_2), \dots, (v_n, p_n, a_n, t_n)\}.
\end{equation*}
In our analysis setup, as illustrated in \cref{fig:main_a}, each video features four animal characters randomly placed in the four quadrants ($p \in \{0, 1, 2, 3\}$ each representing top-left, top-right, bottom-left and bottom-right position), each uttering a different country name in a random order ($t \in \{0, 1, 2, 3\}$). For instance, the video context in \cref{fig:main_a} is represented as, $c = \{(\text{panda}, 0, \text{Mexico},1), (\text{tiger}, 1, \text{Japan}, 3), (\text{monkey}, 2, \text{Canada}, 2), (\text{rabbit}, 3, \text{Egypt}, 0)\}.$

Given a context of video $c$, and a target event $k$, we provide the model with a text prompt describing one semantic attribute of the event, either the audio attribute $a_k$ or the visual attribute $v_k$. This attribute serves as the \emph{anchor}, and the model is asked to retrieve the corresponding \emph{target} attribute from the complementary modality. This yields two sub-tasks: (1) Acoustically-Anchored Visual Retrieval (AAVR), relying on an acoustic anchor $a_k$ to predict $v_k$, and (2) Visually-Anchored Audio Retrieval (VAAR), relying on a visual anchor $v_k$ to predict $a_k$. As illustrated in \cref{fig:main} for the AAVR task, the model is given the acoustic anchor $a_k = \text{``Japan''}$ in the text prompt ``\textit{Japan is said by the}\,'' and is expected to complete the sentence by predicting the target visual attribute $v_k = \text{``tiger''}$.

Following prior mechanistic interpretability work that conditions analyses on successful behaviors~\cite{meng2022locating, geva2023dissecting, fierro2025multilingual}, we restrict our analysis to correctly predicted samples. Since models perform poorly in the standard setting (\cref{fig:wo_hint_cma}), we use contextual priming (e.g., ``Canada is said by monkey, Egypt is said by rabbit, Mexico is said by panda, and Japan is said by'' for AAVR). This induces the correct binding and yields accurate predictions for controlled analysis (see \cref{app:hint_just} of the supp.\ mat. for further discussion).

\subsection{Symbolic Trimodal Binding Mechanisms}
\label{sec:symbolic_mech}
Consistent with recent findings in LLMs and VLMs~\cite{yang2025emergent, assouel2025visual}, we find that AVLLMs utilize symbolic IDs to perform trimodal binding. 
Rather than encoding semantic attributes $(a_i, v_i)$ based on their content, models abstract them into content-independent symbolic IDs and bind modalities by aligning these IDs. 
Specifically, AVLLMs derive symbolic IDs from modality-specific structural cues: sequential speaking order $(t_i)$ for parsing and tracking acoustic attributes, which we term \emph{Temporal IDs}, and spatial locations $(p_i)$ for tracking visual attributes, which we term \emph{Position IDs}.
We then identify a three-stage trimodal binding mechanism based on these modality-specific symbolic IDs.

\newpara{Stage 1: Anchor ID retrieval.} The text prompt describing the anchor attribute is converted into its corresponding symbolic ID (Temporal ID $t_k$ for AAVR, or Position ID $p_k$ for VAAR). We refer to the symbolic ID resulting from this stage as \textit{anchor ID}. For instance, in the AAVR shown in \cref{fig:main_a}, the model represents the text prompt ``Japan'' by mapping it to its temporal order within the utterance sequence (Temporal ID $t_k = 3$).

\newpara{Stage 2: Target ID selection.} The anchor ID is then transformed into the symbolic ID of the target attribute (Position ID $p_k$ for AAVR, or Temporal ID $t_k$ for VAAR). We refer to this as \textit{target ID}. In \cref{fig:main_a}, the Temporal ID $t_k = 3$ is converted into the corresponding Position ID $p_k = 1$.

\newpara{Stage 3: Feature retrieval.} Finally, the model uses the target ID to locate the target attribute and retrieve its semantic content. In \cref{fig:main_a}, the model attends to the spatial location indexed by the Position ID $p_k = 1$, recovers the corresponding visual attribute (``tiger''), and verbalizes it.

\begin{figure}[t]
    \centering
    \scalebox{0.95}{%
    \begin{minipage}{\linewidth}
    \centering
    \begin{subfigure}[b]{0.495\textwidth}
        \centering
        \includegraphics[width=\linewidth]{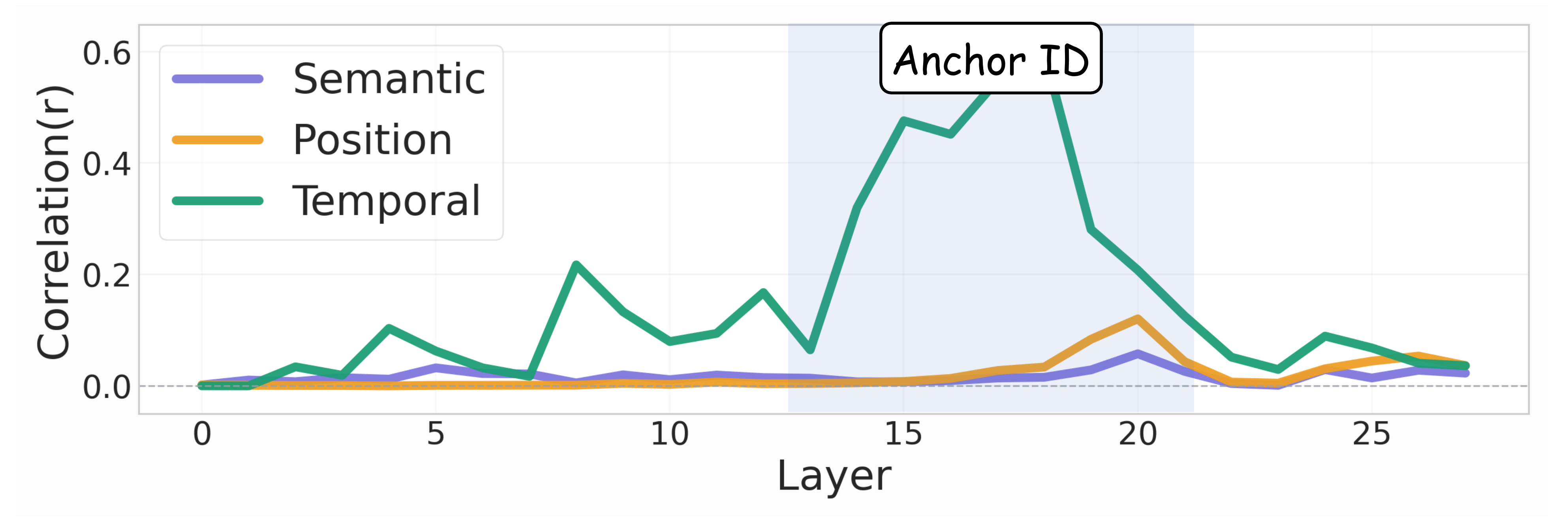}
        \vspace{-0.6cm}
        \caption{RSA at anchor attribute token in AAVR task}
        \label{fig:rsa_1}
    \end{subfigure}
    \hfill
    \begin{subfigure}[b]{0.495\textwidth}
        \centering
        \includegraphics[width=\linewidth]{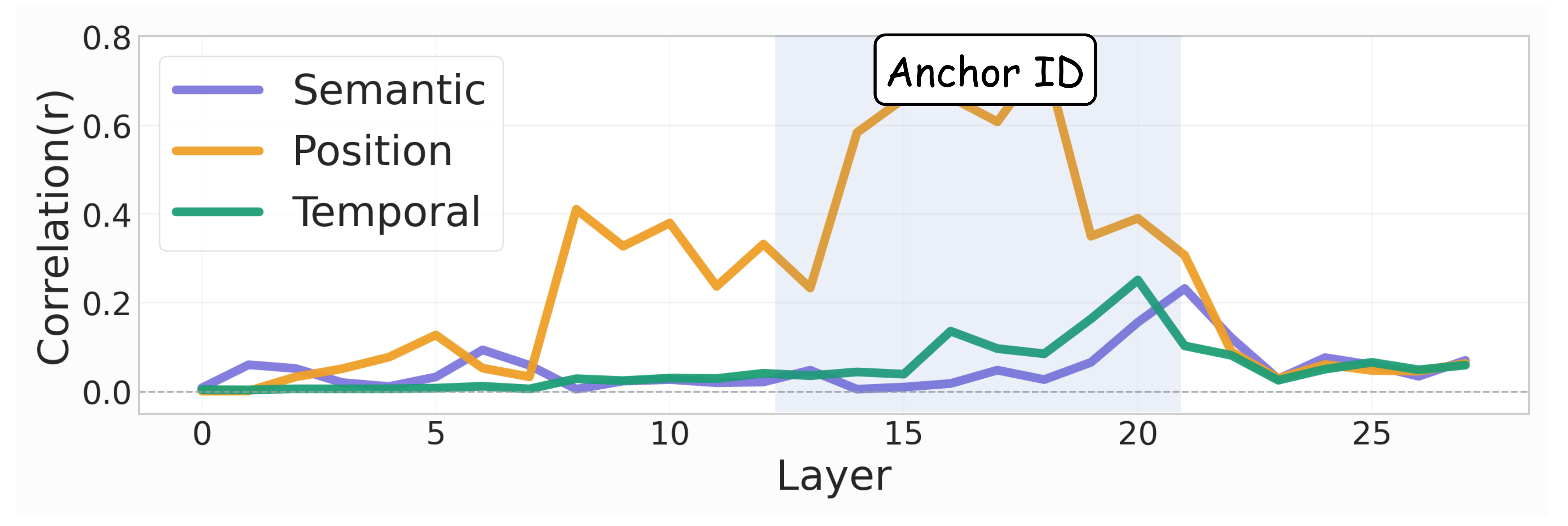}
        \vspace{-0.6cm}
        \caption{RSA at anchor attribute token in VAAR task}
        \label{fig:rsa_2}
    \end{subfigure}
    \vspace{-1mm}
    \begin{subfigure}[b]{0.495\textwidth}
        \centering
        \includegraphics[width=\linewidth]{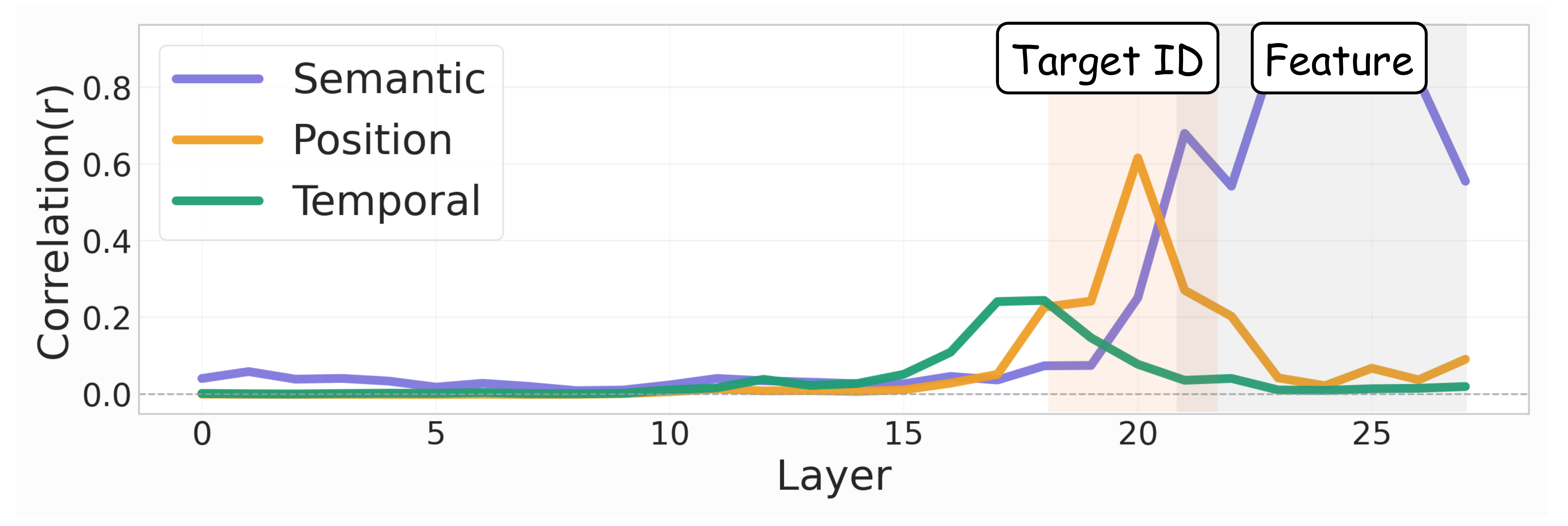}
        \vspace{-0.6cm}
        \caption{RSA at last token in AAVR task}
        \label{fig:rsa_3}
    \end{subfigure}
    \hfill
    \begin{subfigure}[b]{0.495\textwidth}
        \centering
        \includegraphics[width=\linewidth]{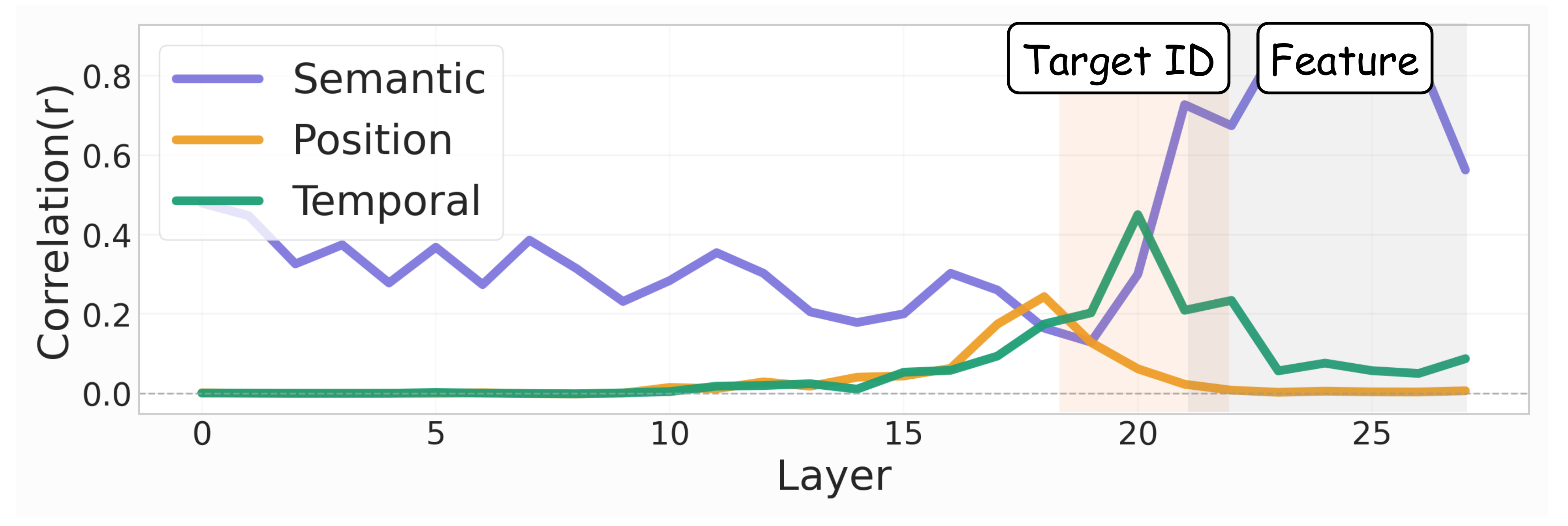}
        \vspace{-0.6cm}
        \caption{RSA at last token in VAAR task}
        \label{fig:rsa_4}
    \end{subfigure}
    \end{minipage}%
    }
    \caption{\textbf{RSA results at anchor attribute and last prompt tokens for AAVR and VAAR tasks.} At the anchor token, anchor attribute symbolic IDs strongly correlate in mid-to-late layers (anchor ID retrieval). At the last token, target attribute symbolic IDs correlate strongly in late layers (target ID selection), while semantic information peaks in the deepest layers (feature retrieval).}
    \vspace{-0.5cm}
\end{figure}
\subsection{Representational Analysis}
\label{sec:rsa}

To examine whether the embeddings of the model encode the proposed symbolic IDs, we perform Representational Similarity Analysis (RSA)~\cite{kriegeskorte2008representational}. RSA compares two embedding spaces by computing pairwise similarity patterns within each space and measuring their Pearson correlation. We compare the hidden states of the model against three hypothesized embedding spaces, each independently encoding the Temporal ID $t_k$, Position ID $p_k$, or semantic content of the target attribute ($a_k$ or $v_k$). Specifically, we extract hidden states from per-layer attention outputs at two key text prompt positions: the token specifying the anchor attribute and the last prompt token. Results are aggregated over 800 samples.

\newpara{Results.} \cref{fig:rsa_1,fig:rsa_2} present the RSA results at the anchor token for the AAVR and VAAR tasks, respectively. In the middle-to-late layers, temporal information dominates in AAVR, while positional information dominates in VAAR. These results suggest the emergence of modality-specific symbolic IDs and anchor ID retrieval stage. At the last prompt token (\cref{fig:rsa_3,fig:rsa_4}), late layers show prominent positional information in AAVR and temporal information in VAAR, reflecting target ID selection. In the deepest layers, semantic information becomes dominant, reflecting the feature retrieval stage. Similar patterns for other models are reported in \cref{app:rep}  of the supp. mat.

\begin{figure}[t]
    \centering
    \includegraphics[width=0.95\linewidth]{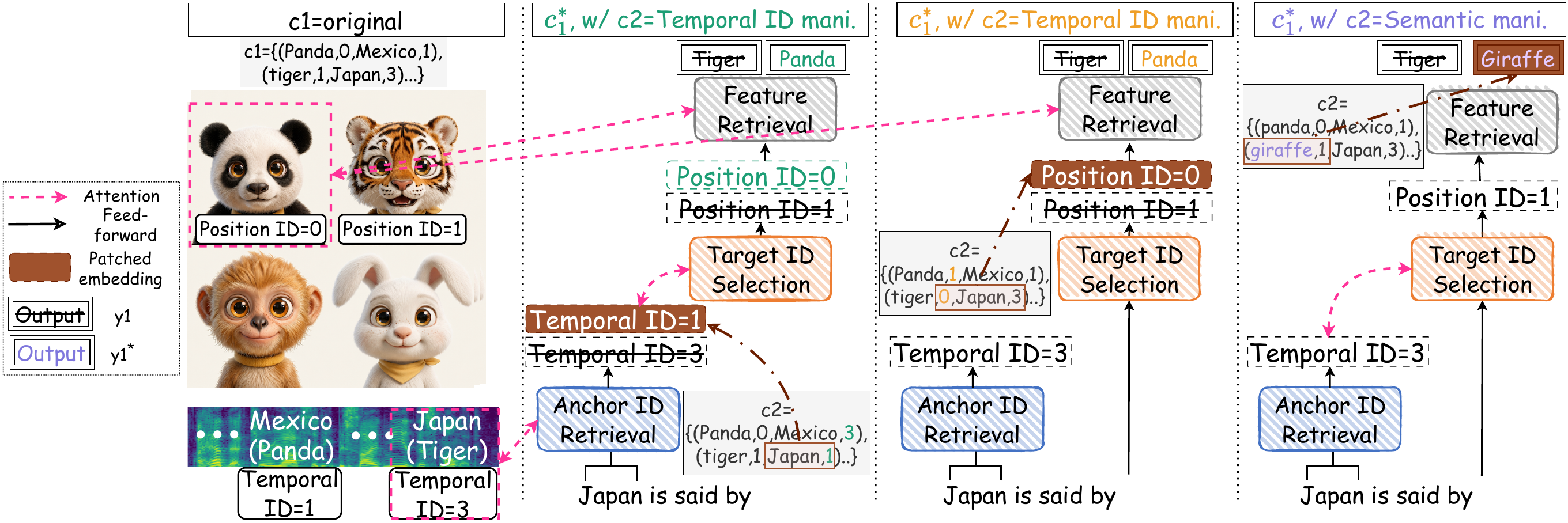}
    \caption{\textbf{Illustration of CMA for the AAVR task using three manipulated contexts.} We construct modified contexts $c_2$ via temporal ID, position ID, and semantic content manipulations, and illustrate the resulting patched run $c_1^*$ obtained by patching $c_2$ outputs into $c_1$ after the anchor ID retrieval, target ID selection, and feature retrieval stages, respectively.}
    \label{fig:aavr_temporal}
    \vspace{-0.4cm}
\end{figure}

\subsection{Causal Mediation Analysis}
\label{sec:cma}
To investigate whether symbolic IDs causally affect the final output, we conduct Causal Mediation Analysis (CMA)~\cite{pearl2022direct, wang2022interpretability, meng2022locating, yang2025emergent}. 
CMA estimates the causal effect of certain hidden states by copying selected states from one forward pass into the same locations of another forward pass and measuring how the original prediction changes. Specifically, it involves three forward passes: an original run under context $c_1$, producing the original output $y_1$; a modified run under context $c_2$, producing output $y_2$; and a patched run $c_1^*$, where selected embeddings from the $c_2$ run are copied (\textit{patch}) into the corresponding locations of the original $c_1$ run, with expected output $y_1^*$, where $y_1^* \neq y_1$.
We quantify the patching effect with the Causal Mediation Score (CM Score)~\cite{wang2022interpretability}:
\begin{equation}
    s = \left( M(c_{1}^*)[y_{1}^*] - M(c_{1}^*)[y_{1}] \right) - \left( M(c_{1})[y_{1}^*] - M(c_{1})[y_{1}] \right)
\end{equation}
where $M(c)[y]$ denotes model $M$'s logit for token $y$ under context $c$.
A higher score indicates that the patched embedding carries information that can steer the prediction toward $y_1^*$.

We apply three distinct manipulations to construct the modified context $c_2$ and define its corresponding expected answer $y_1^*$. 
For \textit{temporal ID manipulation}, we swap the temporal order of the utterance for the target event $k$ and another event $j$, resulting in $c_2 = \{\dots, (v_k, p_k, a_k, t_j), \dots, (v_j, p_j, a_j, t_k), \dots\}$. For \textit{position ID manipulation}, we swap the spatial positions of the animal for events $k$ and $j$, yielding $c_2 = \{\dots, (v_k, p_j, a_k, t_k), \dots, (v_j, p_k, a_j, t_j), \dots\}$. In both cases, the text prompt remains identical across $c_1$ and $c_2$, explicitly querying the target event $k$, yielding $y_1 = y_2 = v_k$ for AAVR, and $y_1 = y_2 = a_k$ for VAAR. In contrast, we set the expected patched answer $y_1^*$ to the target attribute of event $j$ ($v_j$ for AAVR, or $a_j$ for VAAR). Therefore, if this causal patching shifts the output to $y_1^*$, it indicates that these symbolic IDs are being actively utilized by the model. For \textit{semantic content manipulation}, we alter the target attribute by replacing it with a novel semantic attribute not present in the original video. For instance, we substitute the target visual attribute $v_k$ with an unseen $v^* \notin V$ in AAVR (yielding $c_2 = \{\dots, (v^*, p_k, a_k, t_k), \dots\}$), or the target acoustic attribute $a_k$ with an unseen $a^* \notin A$ in VAAR. In both cases, the expected answer is set to this novel attribute ($y_1^* = v^*$ or $a^*$).

For both AAVR and VAAR, we perform CMA using three variants of the modified context $c_2$, patching them at two token positions: the anchor attribute token and the last prompt token. We illustrate, as an example, the expected effect of each manipulation when applied at each stage in the AAVR task in \cref{app:illu_cma} of the supp. mat. 
At each position, we identify which manipulation yields the highest CM Score. 
We use attention outputs as the patched embeddings and apply patching over layer windows spanning one-quarter of the total depth.
We hypothesize the following behavior:

\newpara{Anchor ID retrieval.} When patching at the anchor attribute token, we expect manipulation of the anchor ID (temporal ID for AAVR, position ID for VAAR) to yield the highest CM scores in mid-to-late layers. As illustrated in \cref{fig:aavr_temporal}, when $c_2$ is constructed via temporal ID manipulation in AAVR, the anchor ID retrieval stage extracts the manipulated temporal ID $t_j$. Patching this state into $c_1$ then directs the model's target ID selection stage to choose the matching position ID $p_j$. Ultimately, this leads to the extraction of $v_j$ during feature retrieval stage, which is our expected answer $y_1^*$.

\newpara{Target ID selection.} Conversely, when patching at the last prompt token, we expect the manipulation of the target ID (position ID for AAVR, temporal ID for VAAR) to yield the highest CM scores in the late layers. As illustrated in \cref{fig:aavr_temporal}, when $c_2$ is constructed via position ID manipulation in AAVR, the target ID selection stage for $c_2$ extracts the manipulated position ID $p_j$. Patching this into $c_1$ triggers the extraction of $v_j$ during the feature retrieval stage, which is our expected answer $y_1^*$.

\newpara{Feature retrieval.} Finally, when patching at the deepest layers of the last prompt token, we expect semantic content manipulation to achieve the highest CM scores. As illustrated in \cref{fig:aavr_temporal}, in the AAVR task, the feature retrieval stage for $c_2$ directly extracts the novel attribute $v^*$. Once this state is patched into $c_1$, the model outputs $v^*$, which is our expected answer $y_1^*$.

\begin{figure}[t]
    \centering
    \scalebox{0.95}{%
    \begin{minipage}{\linewidth}
    \centering
    \begin{subfigure}[b]{0.495\textwidth}
        \centering
        \includegraphics[width=\linewidth]{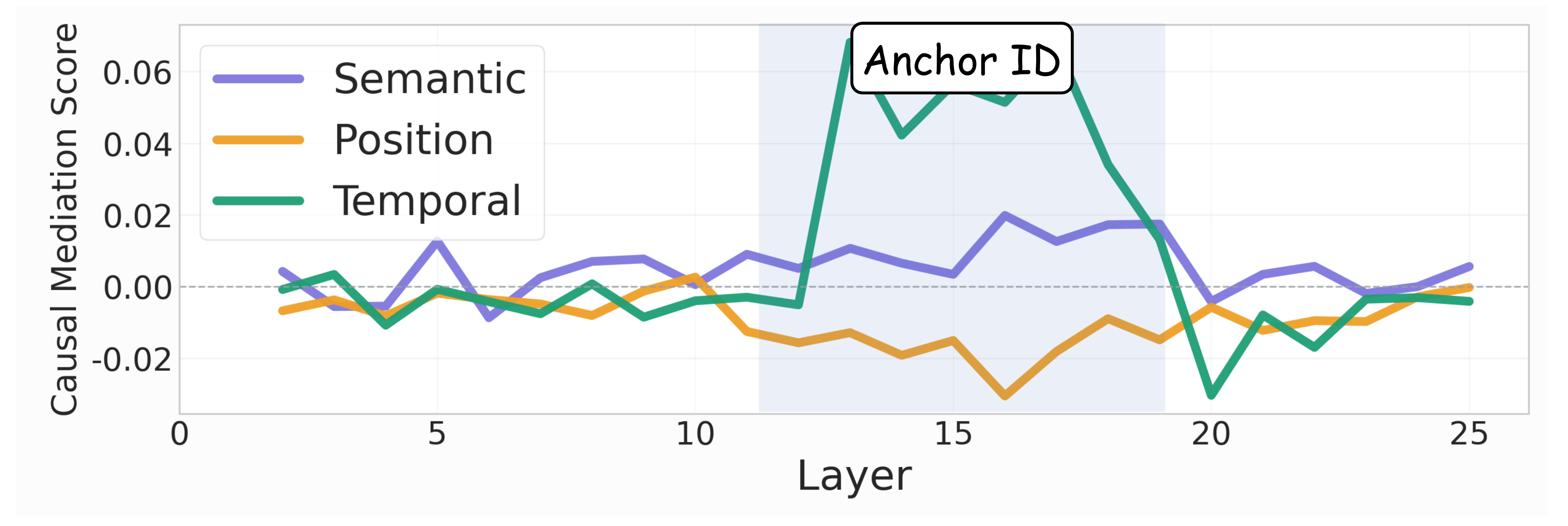}
        \vspace{-0.6cm}
        \caption{CMA at anchor attribute token in AAVR task}
        \label{fig:cma_1}
    \end{subfigure}
    \hfill
    \begin{subfigure}[b]{0.495\textwidth}
        \centering
        \includegraphics[width=\linewidth]{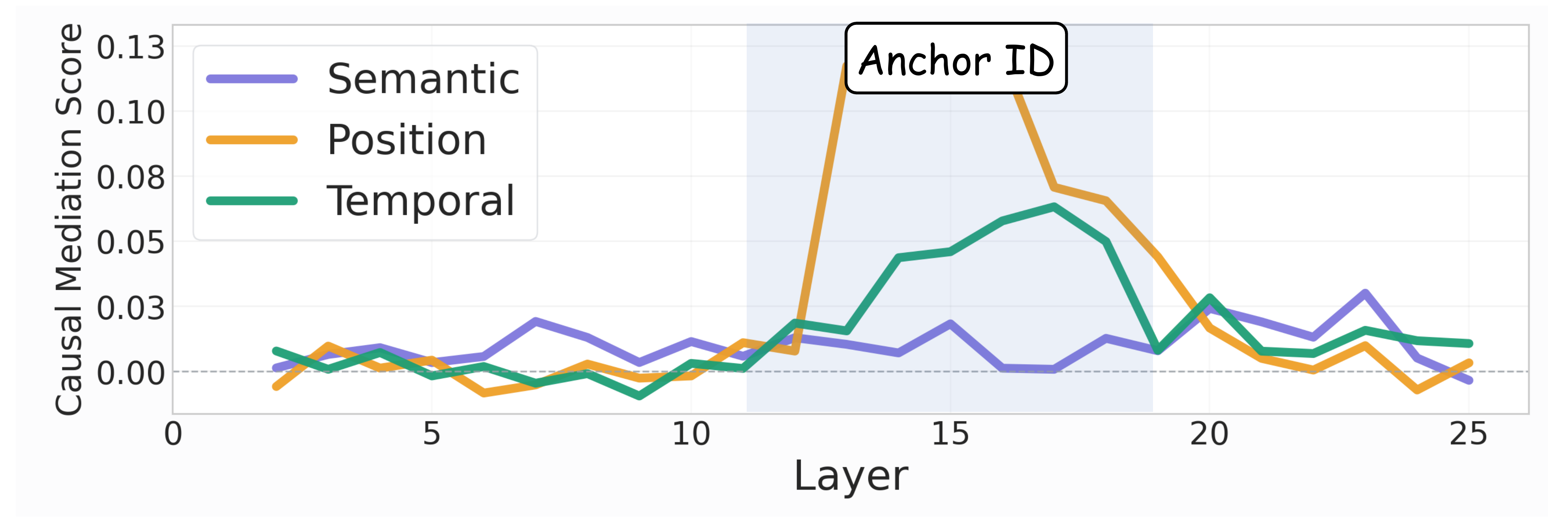}
        \vspace{-0.6cm}
        \caption{CMA at anchor attribute token in VAAR task}
        \label{fig:cma_2}
    \end{subfigure}
    \vspace{-1mm}
    \begin{subfigure}[b]{0.495\textwidth}
        \centering
        \includegraphics[width=\linewidth]{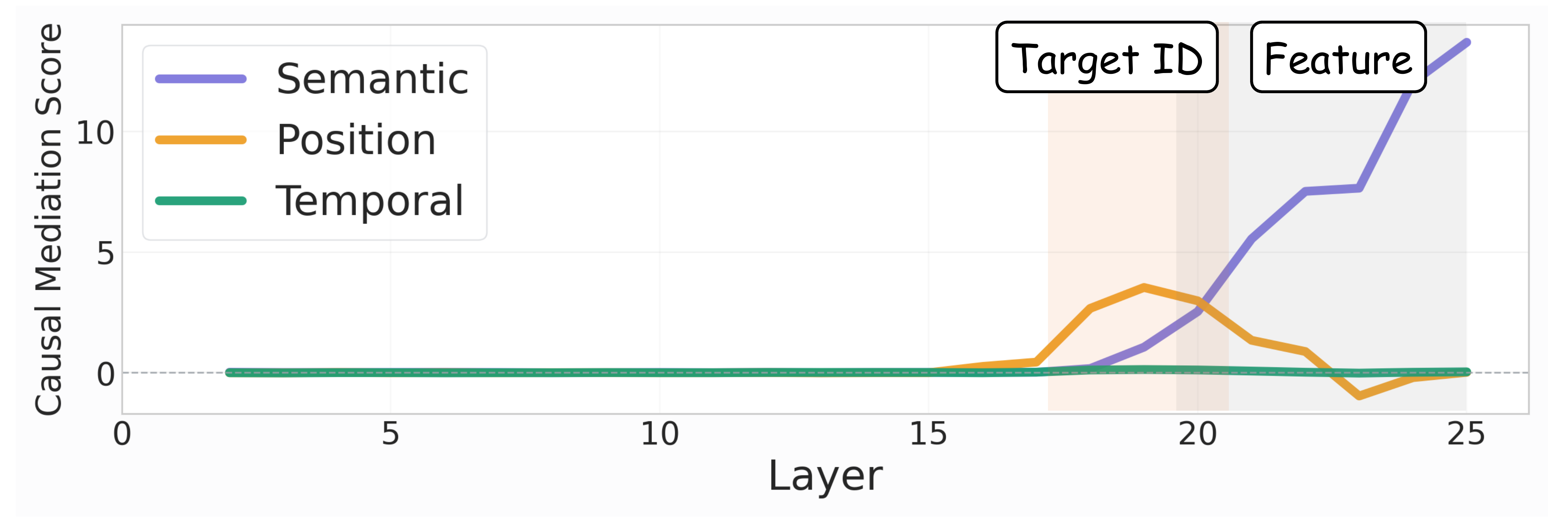}
        \vspace{-0.6cm}
        \caption{CMA at last token in AAVR task}
        \label{fig:cma_3}
    \end{subfigure}
    \hfill
    \begin{subfigure}[b]{0.495\textwidth}
        \centering
        \includegraphics[width=\linewidth]{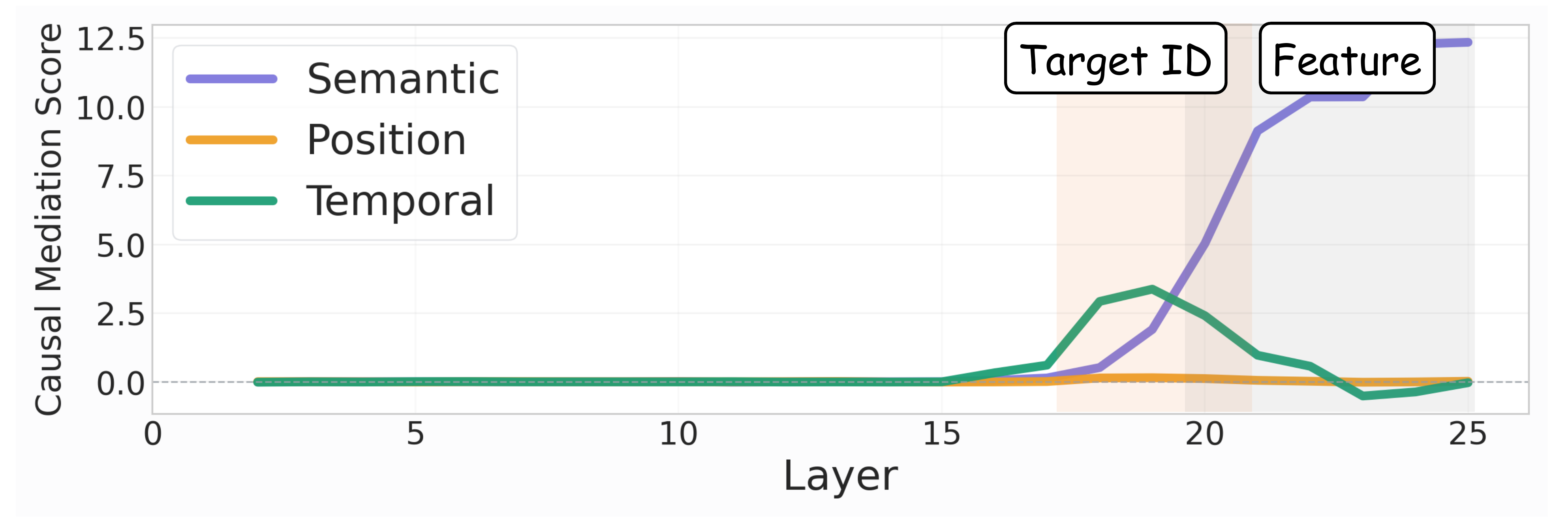}
        \vspace{-0.6cm}
        \caption{CMA at last token in VAAR task}
        \label{fig:cma_4}
    \end{subfigure}
    \end{minipage}%
    }
    \caption{\textbf{CMA results at anchor attribute and last prompt tokens for AAVR and VAAR tasks.} At the anchor token, anchor ID manipulation exhibit high CM scores in mid-to-late layers (anchor ID retrieval). At the last token, target ID manipulation exhibit high CM scores in late layers (target ID selection), while semantic content manipulation peaks in the deepest layers (feature retrieval).}
    \label{fig:cma}
\end{figure}

\begin{figure}[t]
    \centering
    \scalebox{0.95}{%
    \begin{minipage}{\linewidth}
    \centering
    \begin{subfigure}{0.495\textwidth}
        \centering
        \includegraphics[width=\linewidth]{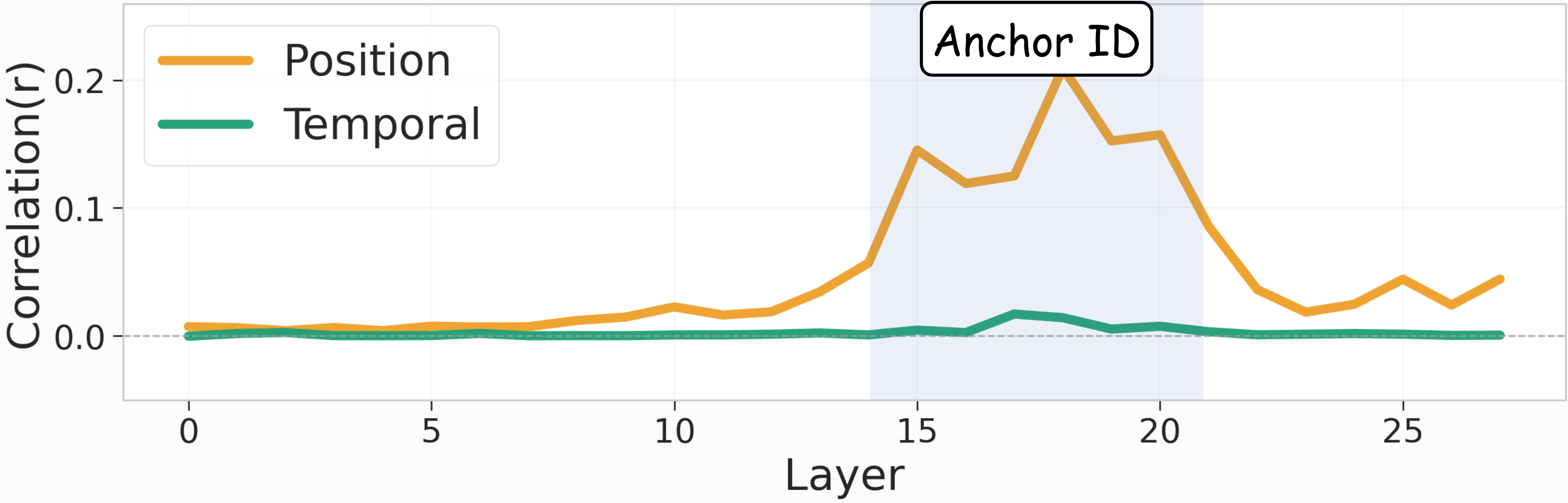}
        \vspace{-6mm}
        \caption{RSA at anchor attribute token in VAAR task}
        \label{fig:real_world_a}
    \end{subfigure}
    \hfill
    \begin{subfigure}{0.495\textwidth}
        \centering
        \includegraphics[width=\linewidth]{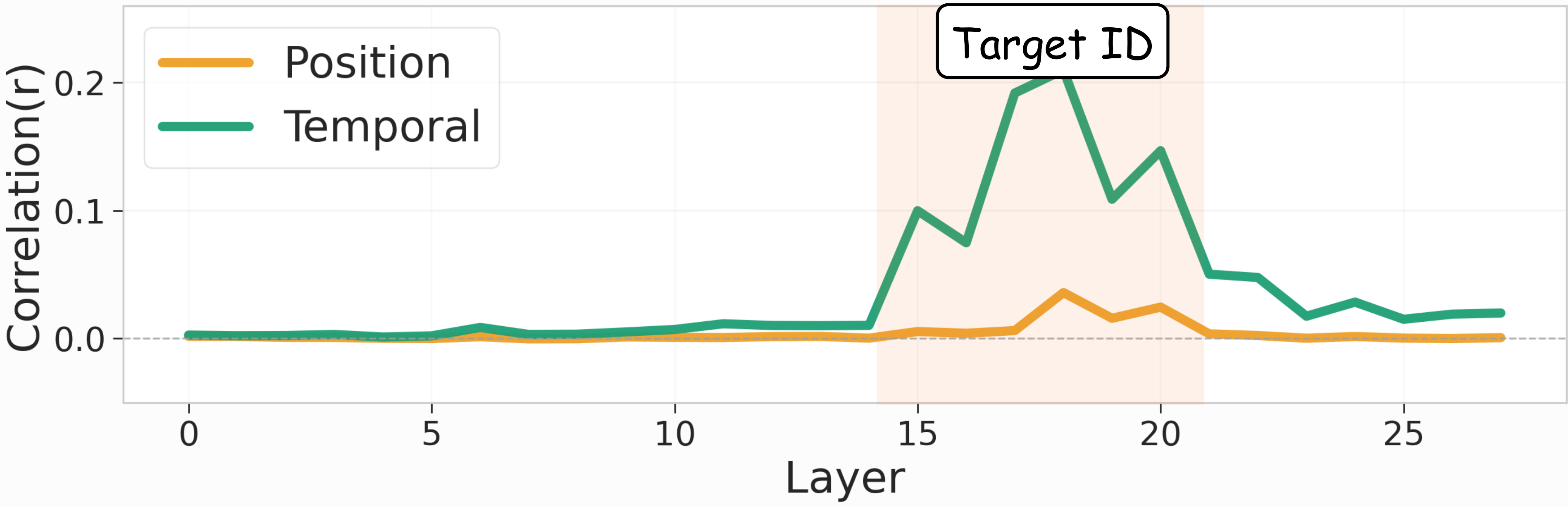}
        \vspace{-6mm}
        \caption{RSA at last token in VAAR task}
        \label{fig:real_world_b}
    \end{subfigure}
    \end{minipage}%
    }
    \caption{\textbf{RSA results in the VAAR task for a real-world dataset.} The symbolic trimodal binding mechanism generalizes to real-world settings. Position IDs exhibit strong correlations at the anchor token (anchor ID retrieval), whereas temporal IDs dominate at the last token (target ID selection).}
    \vspace{-6mm}
\end{figure}

\newpara{Results.}
\cref{fig:cma} presents the CMA results averaged over 960 samples for the AAVR and VAAR tasks. At the anchor attribute token, \cref{fig:cma_1,fig:cma_2} reveal that anchor ID manipulation yields significantly higher scores in the mid-to-late layers compared to other manipulations, consistent with the anchor ID retrieval stage. Moving to the last prompt token, \cref{fig:cma_3,fig:cma_4} show that target ID manipulation produces the strongest effect in the late layers, consistent with the target ID selection stage. In the deepest layers, semantic content manipulation reaches its peak, marking the feature retrieval stage.

\subsection{Real-World Generalization}
\label{sec:real}
To verify the generalization of symbolic mechanisms to real-world settings, we conduct RSA on 2000 clips from the SocialOmni~\citep{xie2026socialomni}. We focus on two-speaker videos to construct AAVR and VAAR tasks. As defining a semantic hypothesis space is infeasible in these unconstrained settings, we compare the embeddings against Temporal and Position IDs. \cref{fig:real_world_a,fig:real_world_b} illustrate the VAAR results at the anchor and last token, respectively. Although the inherent variability of real-world data attenuates the overall correlations, the results support our symbolic mechanism: positional information is pronounced in mid-to-late layers at the anchor token (anchor ID retrieval), while temporal information peaks in late layers at the last token (target ID selection). Similar trends are observed across other models and tasks (\cref{app:real} of the supp. mat.)

\section{Diagnosing and Mitigating Trimodal Binding Failures}
In this section, we first locate the source of trimodal binding failures in the unprimed setting (\cref{sec:failure}). Building on this diagnosis, we propose a simple audio-visual prompting method using an off-the-shelf Active Speaker Detection (ASD) module to compensate for the identified bottleneck (\cref{sec:av_prompt}).

\subsection{Locating the Failure Bottleneck in Trimodal Binding}
\label{sec:failure}
Unlike preceding sections that use contextual priming, we now shift to an unprimed setting, where such contextual cues are absent and failures often occur. We identify the locus of these failures through representational and causal intervention analyses.

\begin{figure}[t]
    \centering
    \scalebox{0.95}{%
    \begin{minipage}{\linewidth}
    \centering
    \begin{subfigure}{0.48\textwidth}
        \centering
        \includegraphics[width=\linewidth]{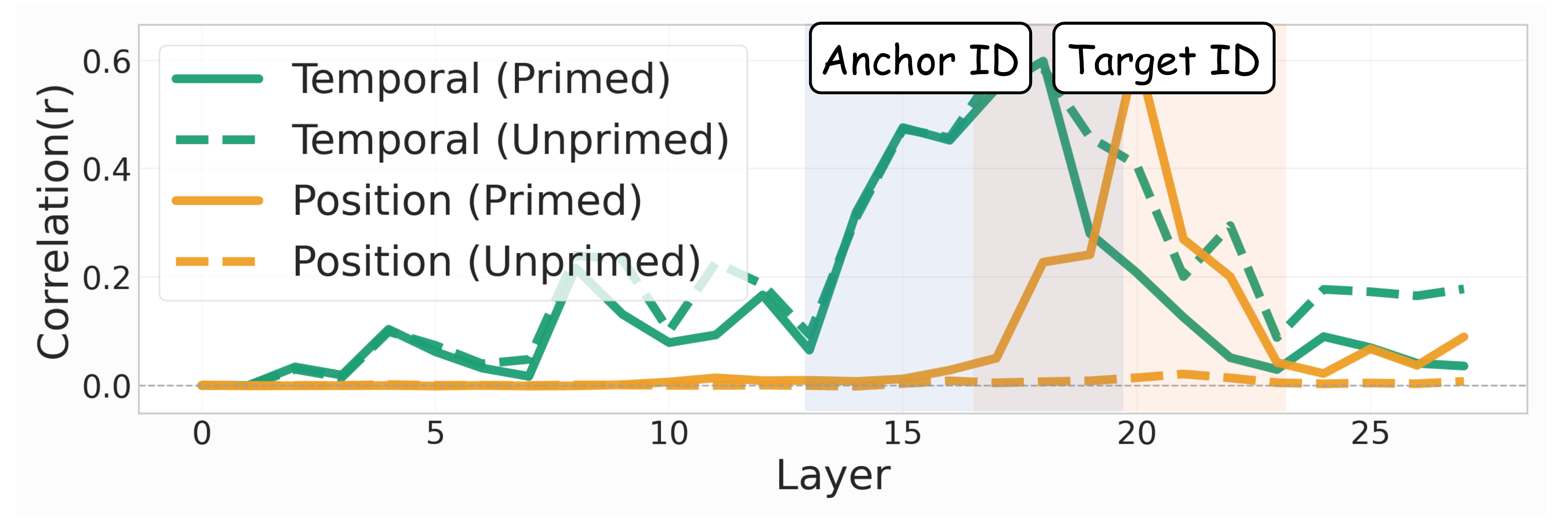}
        \vspace{-0.6cm}
        \caption{RSA results for primed and unprimed setting}
        \label{fig:failure_rep}
    \end{subfigure}
    \hfill
    \begin{subfigure}{0.48\textwidth}
        \centering
        \includegraphics[width=\linewidth]{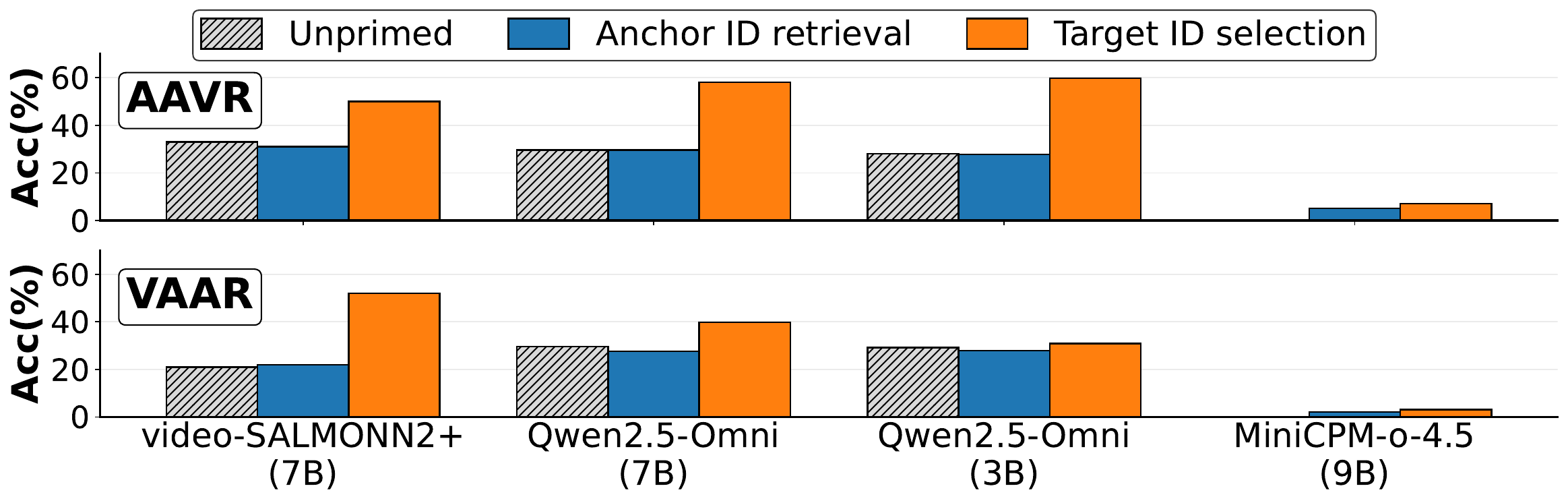}
        \vspace{-0.6cm}
        \caption{Causal intervention analysis}
        \label{fig:wo_hint_cma}
    \end{subfigure}
    \end{minipage}%
    }
    \vspace{-0.1cm}
    \caption{\textbf{Target ID selection stage identified as the locus of failure.} (a) The primed-unprimed representational gap peaks at the target ID selection stage in AAVR task. (b) Patching correct symbolic IDs at target ID selection stage significantly improves accuracy.}
    \label{fig:failure_rep_wo_hint}
    \vspace{-0.4cm}
\end{figure}

\newpara{Representational Analysis}
To localize the failure point, we repeat the RSA from \cref{sec:rsa} in the unprimed setting; the stage with the largest primed-unprimed deviation indicates the likely source of failure. As shown in \cref{fig:failure_rep} for AAVR task, the anchor ID retrieval stage (green solid/dashed lines) shows negligible divergence, whereas the target ID selection stage (yellow solid/dashed lines) exhibits a substantial gap. This suggests that performance degradation stems from a failure in establishing audio-visual correspondence, rather than an inability to parse or textually ground the visual scenes or audio streams. This pattern holds across models and tasks (\cref{app:failure_rep} of our supp.\ mat.).

\newpara{Causal Intervention Analysis}
\label{failure:cia}
To further localize the failure, we perform causal interventions by injecting correct symbolic IDs at each stage. We obtain correct symbolic representations from the primed setting and patch them into the unprimed setting. A substantial accuracy improvement after intervention would implicate the corresponding stage as the primary source of error.
We first identify the specific head-layer pairs for intervention. To minimize confounding factors, we refine our earlier layer-level analysis to individual attention heads, following~\cite{yang2025emergent, assouel2025visual}. By reapplying the CMA (\cref{sec:cma}) at this finer granularity, we select the top 10 pairs yielding the highest scores across a 100-sample subset for each binding stage. Then, to capture the correct symbolic ID, we compute a representative activation vector for each selected pair by averaging its outputs over 800 primed instances. Finally, we patch these cached representations into the unprimed setting to evaluate whether the intervention successfully restores model accuracy.
As shown in \cref{fig:wo_hint_cma}, intervening at the anchor ID retrieval stage has little effect, whereas intervention at the target ID selection stage improves accuracy. This suggests that target ID selection is the key bottleneck responsible for trimodal binding failures.

\subsection{Audio-Visual Prompting via Active Speaker Detection}
\label{sec:av_prompt}

Identifying the target ID selection stage as the critical bottleneck suggests that existing AVLLMs struggle primarily with audio-visual binding. To address this limitation, we propose a simple audio-visual prompting method using an off-the-shelf Active Speaker Detection (ASD) model.
As illustrated in \cref{fig:main_b}, we use a recent ASD model~\cite{nguyen2026laser} to overlay a red bounding box on the detected active speaker in each frame before feeding into AVLLM. To ensure the model correctly interprets this visual cue, the accompanying text prompt indicates that the red box denotes the active speaker. While this approach can work training-free, we further introduce a fine-tuning method to solidify the model's ability to ground these visual cues. To this end, we use videos with these audio-visual prompts to perform supervised fine-tuning (SFT) via Low-Rank Adaptation (LoRA).

\section{Experiments}

\subsection{Experimental Setup}
\begin{table}[t]
\centering
\footnotesize
\setlength{\tabcolsep}{4.5pt}
\renewcommand{\arraystretch}{1.05}

\newlength{\nscol}
\setlength{\nscol}{1.6cm}
\newlength{\wscol}
\setlength{\wscol}{1.6cm}
\caption{\textbf{Results of the proposed methods.} We evaluate three recent AVLLMs across four conversation-centric and three general benchmarks. For Qwen2.5-Omni and MiniCPM-o-4.5, training-free ASD improves  on conversation-centric tasks (For general benchmarks, which often lack speech or visible speakers, we do not apply ASD to draw bounding boxes, making ASD identical to the vanilla setting.). ASD-FT improves bounding box grounding capabilities, yielding consistent performance gains across all three AVLLMs on all seven benchmarks.} 
\resizebox{\textwidth}{!}{%
\begin{tabular}{l w{c}{\nscol} w{c}{\nscol} w{c}{\nscol} w{c}{\wscol} w{c}{\wscol}!{\vrule width 0.6pt} w{c}{\nscol} w{c}{\nscol} w{c}{\nscol}} 
\toprule
\multirow{3}{*}{\textbf{Model}} 
& \multicolumn{5}{c!{\vrule width 0.6pt}}{\textbf{Converse-Centric}} 
& \multicolumn{3}{c}{\textbf{General}} \\
\cmidrule(lr){2-6} \cmidrule(l){7-9}
& AVSpeaker\cite{nguyen2025see}
& DailyOmni\cite{zhou2025daily}
& SocialOmni\cite{xie2026socialomni}
& \multicolumn{2}{c!{\vrule width 0.6pt}}{DiaDemBench\cite{chen2026diadem}}
& OmniBench\cite{li2025omnibench}
& DAVE\cite{radevski2025dave}
& WorldSense\cite{hong2026worldsense} \\
\cmidrule(lr){2-2} \cmidrule(lr){3-3} \cmidrule(lr){4-4} \cmidrule(lr){5-6} \cmidrule(lr){7-7} \cmidrule(lr){8-8} \cmidrule(l){9-9}
& Acc(\%) $\uparrow$
& Acc(\%) $\uparrow$
& Acc(\%) $\uparrow$
& REF $\uparrow$ & ASR $\uparrow$
& Acc(\%) $\uparrow$
& Acc(\%) $\uparrow$
& Acc(\%) $\uparrow$ \\
\midrule
Qwen2.5-Omni\cite{xu2025qwen25omnitechnicalreport} {\scriptsize(7B)}
& 44.86 & 64.33 & 38.95
& 19.1{\scriptsize$\pm$0.4} & 28.1{\scriptsize$\pm$0.1}
& 50.79 & 30.75 & 50.06 \\
\rowcolor{gray!15}
\quad + \textbf{ASD}
& \underline{45.59} & \underline{64.66} & \underline{41.90}
& \textbf{22.5}{\scriptsize$\pm$0.2}& \textbf{33.0}{\scriptsize$\pm$0.2} 
& \cellcolor{white}50.79 & \cellcolor{white}30.75 & \cellcolor{white}50.06 \\
\rowcolor{gray!15}
\quad + \textbf{ASD-FT}
& \textbf{48.79} & \textbf{71.09} & \textbf{44.25}
& \underline{20.7}{\scriptsize$\pm$0.2} & \underline{32.3}{\scriptsize$\pm$0.1}
& \textbf{52.80} & \textbf{33.91} & \textbf{53.73}  \\
\midrule
MiniCPM-o-4.5\cite{MiniCPM} {\scriptsize(9B)}
& 49.41 & 61.99 & 46.05
& 6.3{\scriptsize$\pm$0.5} & 7.4{\scriptsize$\pm$0.7}
& 47.99 & 55.80 & 55.73 \\
\rowcolor{gray!15}
\quad + \textbf{ASD}
& \underline{50.34} & \underline{62.49} & \textbf{50.00}
& \underline{9.7}{\scriptsize$\pm$0.4} & \underline{10.8}{\scriptsize$\pm$0.1}
& \cellcolor{white}47.99 & \cellcolor{white}55.80 & \cellcolor{white}55.73 \\
\rowcolor{gray!15}
\quad + \textbf{ASD-FT}
& \textbf{51.77} & \textbf{63.41} & \underline{49.70}
& \textbf{24.8}{\scriptsize$\pm$0.2} & \textbf{30.9}{\scriptsize$\pm$0.0}
& \textbf{51.84} & \textbf{57.97} & \textbf{57.37} \\
\midrule
video-SALMONN2+\cite{tang2025video} {\scriptsize(7B)}
& \underline{44.74} & \underline{65.13} & 43.49
& \underline{14.0}{\scriptsize$\pm$0.3} & \underline{19.3}{\scriptsize$\pm$0.1}
& 37.04 & 35.28 & 45.04 \\
\quad + ASD
& 44.65 & 62.57 & \underline{43.89}
& 11.8{\scriptsize$\pm$0.3} & 19.0{\scriptsize$\pm$0.0}  
& 37.04 & 35.28 & 45.04 \\
\rowcolor{gray!15}
\quad + \textbf{ASD-FT}
& \textbf{49.10} & \textbf{69.13} & \textbf{49.62}
& \textbf{19.7}{\scriptsize$\pm$0.2} & \textbf{26.7}{\scriptsize$\pm$0.2}
& \textbf{44.92} & \textbf{39.65} & \textbf{46.38} \\ 
\bottomrule
\end{tabular}%
}
\vspace{-0.5cm}
\label{tab:main}
\end{table}
\newpara{Training datasets.}
For the LoRA fine-tuning dataset, we construct a specialized corpus of 400 synthetic videos by adapting the toy dataset framework from our preceding analysis. Specifically, we replace the original animal speakers with human speakers to bridge the domain gap, ensuring the active speaker is highlighted with a red bounding box. Each video is paired with two types of tasks: a video captioning task describing the appearances and utterances of all speakers, and multiple-choice questions evaluating AAVR and VAAR tasks. See \cref{app:training_data} of the supp.\ mat. for more details.

\newpara{Implementation details.} 
We adopt Qwen2.5-Omni (7B), video-SALMONN2+ (7B), and MiniCPM-o-4.5. All fine-tuning is conducted on a single NVIDIA RTX A6000 GPU. We apply LoRA with a rank of 16, employing a batch size of 1 and 8 gradient accumulation steps, with all models trained for fewer than 300 optimization steps. See \cref{app:training_detail} of the supp.\ mat. for more details.

\newpara{Evaluation datasets and protocol.}
We evaluate our approach on four conversation-centric benchmarks encompassing diverse task formats. For multiple-choice question answering, we adopt AVSpeaker~\cite{nguyen2025see},  DailyOmni~\cite{zhou2025daily}, and perceptual task subset of SocialOmni~\cite{xie2026socialomni}. For multi-speaker dialogue captioning, we use DiaDemBench~\cite{chen2026diadem}. Performance is evaluated on two metrics-ASR for transcription accuracy and REF for speaker attribution consistency-with scoring conducted using gemini-2.5-Pro and gemini-2.5-Flash. To assess generalization beyond conversation-centric scenarios, we further evaluate on three broader audio-visual benchmarks: DAVE~\cite{radevski2025dave} for egocentric videos, OmniBench~\cite{li2025omnibench} for image-audio pairs, and WorldSense~\cite{hong2026worldsense} for diverse audio categories including music and environmental sounds.

\subsection{Experimental Results}

\paragraph{Quantitative results.}
\begin{table}[t]
\centering
\scriptsize
\renewcommand{\arraystretch}{1.0}
\begin{minipage}[t]{0.45\linewidth}
\centering
\setlength{\tabcolsep}{10pt}
\caption{\textbf{Comparison of training-free methods.} Our ASD surpasses both decoding- and prompting-based methods.}
\label{tab:decoding_comparison}
\vspace{1mm}
\resizebox{\linewidth}{!}{
\begin{tabular}{llc}
\toprule
\textbf{Type} & \textbf{Method} & \textbf{AVSpeaker} \\
\midrule
Vanilla & --- & 44.86 \\
\midrule
\multirow{3}{*}{Decoding} & AVCD \cite{jung2025avcd} & 42.82 \\
 & FMD \cite{jung2025fork} & \underline{45.27} \\
 & OutRo \cite{yoo2026nature} & 45.24 \\
\midrule
\multirow{3}{*}{Prompting} & NumPro \cite{wu2025number} & 45.08 \\
 & VISER \cite{izadi2025visual} & 44.55 \\
 & \cellcolor{gray!15}\textbf{ASD(Ours)} & \cellcolor{gray!15}\textbf{45.59} \\
\bottomrule
\end{tabular}}
\end{minipage}%
\hfill
\begin{minipage}[t]{0.5\linewidth}
\centering
\setlength{\tabcolsep}{6pt}
\caption{\textbf{Ablation on the prompting style of training-free ASD.} \up and \dn indicate performance gain and drop compared to vanilla.}
\label{tab:ablation1}
\vspace{1mm}
\resizebox{\linewidth}{!}{
\begin{tabular}{cccc|c}
\toprule
\textbf{BBox} & \textbf{Text} & \textbf{Color} & \textbf{Thickness} & \textbf{AVSpeaker} \\
\midrule
\multicolumn{4}{c|}{\textit{Vanilla}} & 44.86 \\
\midrule
Inactive & \cmark & \textcolor{red}{Red}           & Medium & 44.77\,\dn \\
Random   & \cmark & \textcolor{red}{Red}           & Medium & 44.52\,\dn \\
Active   & \xmark & \textcolor{red}{Red}           & Medium & 44.80\,\dn \\
\midrule
Active   & \cmark & \textcolor{ForestGreen}{Green} & Medium & 45.95\,\up \\
Active   & \cmark & \textcolor{red}{Red}           & Thick  & 45.70\,\up \\
Active   & \cmark & \textcolor{red}{Red}           & Thin   & 45.58\,\up \\
\bottomrule
\end{tabular}}
\end{minipage}
\end{table}

\cref{tab:main} presents the quantitative results. The training-free ASD approach demonstrates clear effectiveness across three speech-centric datasets for Qwen2.5-Omni and MiniCPM-o-4.5. This suggests that the ASD intervention effectively compensates for the models' inherent deficiencies in audio-visual binding. However, because this approach relies on a pre-existing ability to comprehend and ground bounding boxes, models for which this ability is less reliable in a zero-shot setting, such as video-SALMONN2+, benefit from explicit fine-tuning. After fine-tuning on ASD-prompted videos (ASD-FT), all three models improve across the three conversation-centric datasets. Interestingly, the ASD-FT models also show consistent gains on general datasets. These results suggest that ASD-FT strengthens audio-visual alignment beyond simple bounding-box comprehension, resolving the binding bottlenecks identified in our analysis. See \cref{app:diss,app:quali} of
our supp. mat. for more discussions and qualitative results.

\newpara{Comparison with other training-free methods.} We compare our training-free ASD against existing training-free decoding methods for AVLLMs-AVCD~\cite{jung2025avcd}, FMD~\cite{jung2025fork}, and OutRo~\cite{yoo2026nature}-and visual prompting methods-NumPro~\cite{wu2025number} and VISER~\cite{izadi2025visual} on Qwen2.5-Omni~(7B) on the AVSpeaker~\cite{nguyen2025see}. \cref{tab:decoding_comparison} shows that ASD outperforms all baselines.

\newpara{Ablation on prompting styles.}
To examine the effect of prompting styles in training-free ASD, we conduct an ablation study using Qwen2.5-Omni on the AVSpeaker (\cref{tab:ablation1}). First, applying the bounding box to the inactive speaker or a random region degrades performance, showing that ASD's gains stem from accurate active-speaker localization rather than generic visual saliency. Second, removing the explanatory text prompt that defines the bounding box degrades performance, confirming that explicit instructions are necessary for visual grounding. Third, the gains are robust to bounding box style variations, such as color or thickness.

\begin{wraptable}{r}{0.48\linewidth}
\vspace{-4.5mm}
\centering
\setlength{\tabcolsep}{5pt}
\renewcommand{\arraystretch}{1.0}
\caption{\textbf{Ablation on finetuning methods.} ASD-FT yields greater gains than baseline FT.}
\label{tab:ablation2}
\vspace{-2mm}
\scriptsize
\resizebox{\linewidth}{!}{
\begin{tabular}{lccccc}
\toprule
& \multicolumn{2}{c}{\textbf{Converse-Centric}} & \multicolumn{3}{c}{\textbf{General}} \\
\cmidrule(lr){2-3} \cmidrule(lr){4-6}
\textbf{Method}
 & \makecell{\textbf{AV-}\\\textbf{Speaker}}
 & \makecell{\textbf{Daily-}\\\textbf{Omni}}
 & \makecell{\textbf{Omni-}\\\textbf{Bench}}
 & \textbf{DAVE}
 & \makecell{\textbf{World-}\\\textbf{Sense}} \\
\midrule
Vanilla & 44.86 & 64.33 & 50.79 & 30.75 & 50.06 \\
\quad + FT & \underline{46.89} & \underline{66.67} & \underline{51.05} & \textbf{35.15} & \underline{52.05} \\
\rowcolor{gray!15} \quad + \textbf{ASD-FT} & \textbf{48.79} & \textbf{71.09} & \textbf{52.80} & \underline{33.91} & \textbf{53.73} \\
\bottomrule
\end{tabular}}
\vspace{-16pt}
\end{wraptable}
\newpara{Ablation on finetuning methods.}
We compare ASD-FT with a baseline fine-tuned on videos without bounding boxes (FT) using Qwen2.5-Omni (\cref{tab:ablation2}). While FT improves performance, ASD-FT yields larger gains 4 out of 5 datasets. This suggests that the ASD-prompted setup directs the model's learning capacity toward aligning audio with the correct visual region, rather than merely improving general task performance.

\begin{figure}[t]
    \centering
    \scalebox{0.95}{%
    \begin{minipage}{\linewidth}
    \centering
    \begin{subfigure}{0.495\linewidth}
        \centering
        \includegraphics[width=\linewidth]{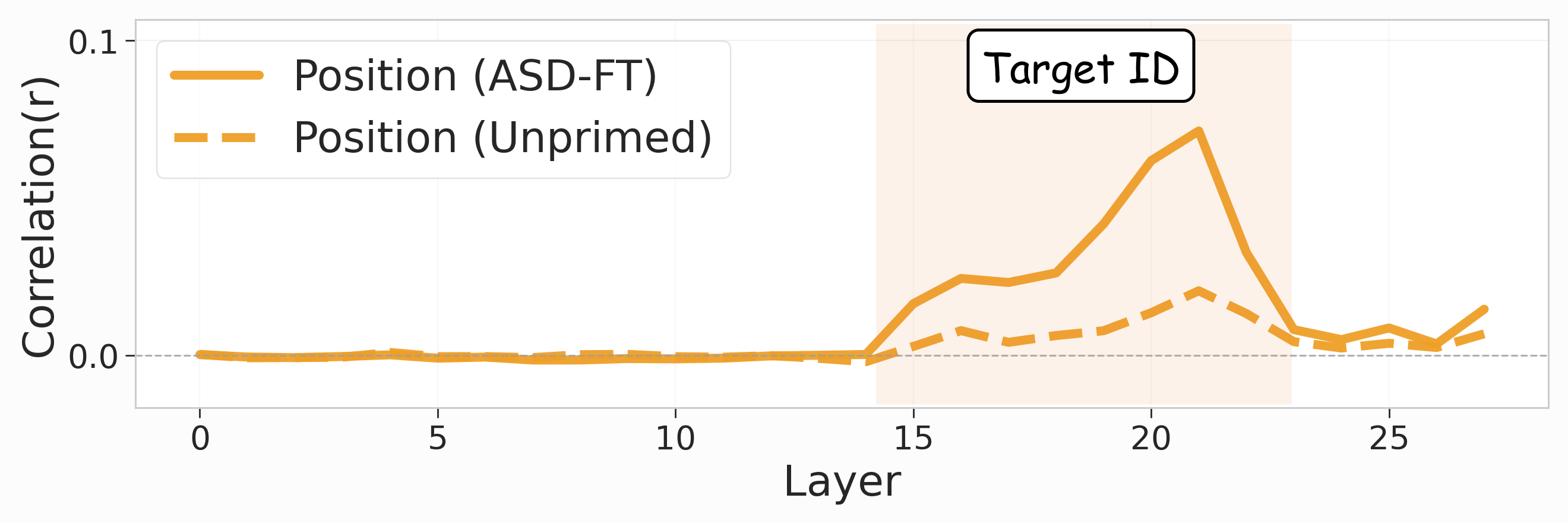}
        \vspace{-0.4cm}
        \caption{AAVR task}
        \label{fig:after_lora_aavr}
    \end{subfigure}
    \hfill
    \begin{subfigure}{0.495\linewidth}
        \centering
        \includegraphics[width=\linewidth]{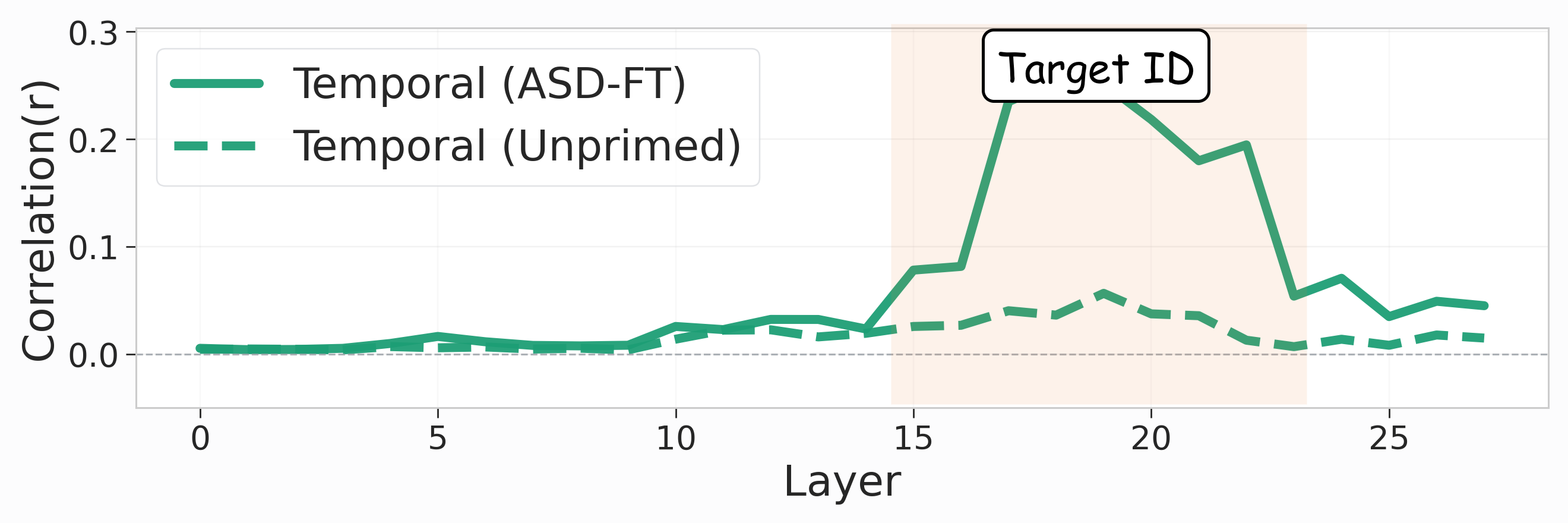}
        \vspace{-0.4cm}
        \caption{VAAR task}
        \label{fig:after_lora_vaar}
    \end{subfigure}
    \end{minipage}}
    \vspace{-0.2cm}
    \caption{\textbf{Target ID selection after ASD-FT in the unprimed setting.} Compared to the vanilla model (dashed), ASD-FT (solid) shows stronger correlation with the target ID.}
    \vspace{-0.4cm}
    \label{fig:after_lora}
\end{figure}
\newpara{Revisiting the target ID selection stage.}
We revisit the RSA in \cref{sec:failure} to assess whether ASD-FT resolves the target ID selection bottleneck. \cref{fig:after_lora} compares the vanilla model (dashed lines) and the ASD-FT model (solid lines) on unprimed AAVR and VAAR tasks. Notably, the ASD-FT model exhibits higher correlation with the target ID in the late layers (target ID selection stage). This suggests that ASD-FT mitigate the target ID selection errors observed in the vanilla model.

\section{Conclusion}

In this paper, motivated by the observation that AVLLMs struggle to resolve ``who says what'' in multi-speaker settings, we investigate the underlying trimodal binding mechanisms. Our investigation reveals that AVLLMs utilize modality-specific symbolic IDs and process information through a distinct three-stage mechanism. Additionally, we identify faulty audio-visual binding, rather than text-audio or text-visual grounding, as the primary bottleneck for overall binding failures. To address this, we propose a simple yet effective audio-visual prompting method utilizing an off-the-shelf Active Speaker Detection model (ASD). We envision that these findings will provide foundational insights for the architectural design and training strategies of next-generation AVLLMs.
 
\section*{Acknowledgement}
This work was supported by Institute of Information \& communications Technology Planning \& Evaluation (IITP) grant funded by the Korea government (MSIT) (RS-2025-
02215122, Development and Demonstration of Lightweight AI Model for Smart Homes).

\bibliographystyle{plainnat}
\bibliography{shorstrings,main} 

\newpage

\appendix
\clearpage

\phantomsection
\section*{Appendix}
\addcontentsline{toc}{section}{Appendix}
\label{app:main}

\vspace{0.5em}
\noindent{\Large\bfseries Appendix Contents}

\vspace{1.2em}

\noindent\textbf{\hyperref[app:symbolic_binding]{A. Symbolic Binding Mechanisms}}\par
\vspace{0.3em}
\noindent\hspace{1.5em}\hyperref[app:hint_just]{A.1 \quad Reasons to Introduce Priming}\par
\noindent\hspace{1.5em}\hyperref[app:rep]{A.2 \quad Representational Analysis and Causal Mediation Analysis on Different Models}\par
\noindent\hspace{1.5em}\hyperref[app:gen_more_spk]{A.3 \quad Generalization of Binding Mechanism Beyond the Controlled Four-Speaker Setting}\par
\noindent\hspace{1.5em}\hyperref[app:failure_rep]{A.4 \quad Representational Analysis in Primed vs. Unprimed Settings}\par
\noindent\hspace{1.5em}\hyperref[app:real]{A.5 \quad Real-World Generalization Analysis on Different Models}\par

\vspace{1.0em}

\noindent\textbf{\hyperref[app:audio_visual_prompting]{B. Audio-Visual Prompting}}\par
\vspace{0.3em}
\noindent\hspace{1.5em}\hyperref[app:training_Free_prompting]{B.1 \quad Details of Training-Free Prompting}\par
\noindent\hspace{1.5em}\hyperref[app:details_of_ft]{B.2 \quad Details of Finetuning}\par
\noindent\hspace{1.5em}\hyperref[app:diss]{B.3 \quad Discussion on the Experimental Results}\par

\vspace{1.0em}

\noindent\textbf{\hyperref[app:illu]{C. Illustrations}}\par
\vspace{0.3em}
\noindent\hspace{1.5em}\hyperref[app:vaar_illu]{C.1 \quad Illustration of the Symbolic Mechanism in VAAR}\par
\noindent\hspace{1.5em}\hyperref[app:illu_cma]{C.2 \quad Illustration of Causal Mediation Analysis for AAVR}\par

\vspace{1.0em}

\noindent\textbf{\hyperref[app:quali]{D. Qualitative Results}}\par

\vspace{1.0em}

\noindent\textbf{\hyperref[app:limit]{E. Limitations}}\par

\vspace{1.0em}

\noindent\textbf{\hyperref[app:compu]{F. Computational Resource}}\par

\vspace{1.0em}

\noindent\textbf{\hyperref[app:social]{G. Social Impact}}\par

\clearpage
\section{Symbolic Binding Mechanisms}
\label{app:symbolic_binding}
\subsection{Reasons to Introduce Priming}
\label{app:hint_just}

Our mechanistic analysis (\cref{sec:symbolic_mech}) conditions on contextually primed inputs. This appendix first clarifies what information the prime provides, and then explains why such priming is methodologically necessary. We further address three potential concerns: whether the prime leaks the answer, whether the primed model merely emits residual audio-visual content, and whether priming induces a circuit unrelated to the unprimed setting.

\newpara{Primed setting.}
In the primed setting, the model receives a prompt of the following form:
\begin{quote}
\textit{``There are four animals in the clip, each speaking one word. Continue the sentence with the single most likely next word. Output exactly one word. Do not add any explanation, extra words, or punctuation. Sentence: Germany is said by the panda, canada is said by the monkey, japan is said by the tiger, and egypt is said by the \_\_\_''}
\end{quote}
The order of the three demonstration speakers in the prime is randomized across samples, so that no fixed positional cue is associated with the queried speaker. The prime serves two roles. First, it provides structural scaffolding by specifying the task format: what type of token to emit and in what grammatical slot. Second, it helps the model form the correct binding for the target speaker by demonstrating the form of the audio-visual binding computation using three speakers.

\newpara{Priming is necessary for causal analysis.}
Mechanistic causal analyses are typically conducted on inputs for which the model produces the correct output~\cite{meng2022locating, geva2023dissecting, fierro2025multilingual}. This is necessary because patching effects are meaningful only when there is a reliable target behavior to explain. When model performance is near chance, causal effects are easily dominated by noise rather than by the computation of interest. In the unprimed setting, the AVLLMs we study do not provide such a success regime (\cref{tab:unprimed-acc}): all four models achieve less than $30\%$ accuracy. In contrast, priming raises accuracy to near-ceiling levels across all models, enabling us to measure causal effects on trajectories where the model reliably performs the intended audio-visual binding computation.

\begin{table}[h]
\centering
\caption{Accuracy across the four AVLLMs in three conditions: unprimed, primed, and primed prompt only. All values are averaged over $800$ samples per model.}
\label{tab:unprimed-acc}
\begin{tabular}{lccc}
\toprule
Model & Unprimed & Primed & Primed prompt only \\
\midrule
video-SALMONN2+ (7B) & $27.0\%$ & $99.2\%$ & $1.00\%$ \\
Qwen2.5-Omni (7B)    & $29.6\%$ & $100.0\%$ & $12.50\%$ \\
Qwen2.5-Omni (3B)    & $28.7\%$ & $99.94\%$ & $12.50\%$ \\
MiniCPM-o-4.5 (9B)   & $0.0\%$  & $99.31\%$ & $0.06\%$ \\
\bottomrule
\end{tabular}
\end{table}

\newpara{Priming does not leak the answer.}
Although the prime demonstrates three speaker--word bindings, it does not specify the queried speaker. The correct answer therefore still depends on the audio-visual content of the clip. Moreover, the task is formulated as open-ended next-token generation rather than multiple-choice QA: the model generates from its full output vocabulary, and the candidate set is never enumerated in the prompt. Thus, text-only shortcuts based on selecting from an explicitly provided candidate list are not available.

We verify this empirically by removing the audio and video while keeping the primed prompt unchanged. As shown in \cref{tab:unprimed-acc}, accuracy collapses to near-floor levels in the prompt-only condition: $1.0\%$ for video-SALMONN2+, $12.5\%$ for Qwen2.5-Omni(7B), and $0.06\%$ for MiniCPM-o-4.5. The relatively higher prompt-only accuracy of Qwen2.5-Omni(7B) does not indicate successful binding from text alone. Rather, its outputs are largely driven by lexical priors over common country and animal names; some of these high-prior tokens, such as \textit{japan} and \textit{panda}, overlap with labels used in our dataset. Thus, prompt-only performance reflects coincidental lexical bias rather than recovery of the correct audio-visual binding.

\newpara{Primed predictions are not residual content emission.}
Another concern is that, in the primed setting, the model may not form a binding with the anchor at all; instead, it may simply emit residual audio-visual content that remains salient after processing the prime. The CMA results in \cref{sec:cma} argue against this interpretation. When we patch the symbolic ID at the anchor attribute token, the output logit shifts toward the target attribute associated with the manipulated ID. If the model were merely emitting residual content independently of the anchor, changing the symbolic representation patched at the anchor position should not systematically change the predicted target. The observed direction of the logit shift therefore indicates that primed predictions are produced through an active cross-modal binding computation.

\newpara{Priming does not induce a separate circuit.}
Finally, one might worry that priming induces a circuit unrelated to the one used in the unprimed setting. Our causal intervention analysis in \cref{failure:cia} suggests otherwise. We identify the top-$10$ attention heads associated with target-ID selection in the primed regime and patch their averaged activations into unprimed forward passes at the same locations. This intervention substantially improves unprimed accuracy: from $27.0\%$ to $51.0\%$ for video-SALMONN2+(7B), from $29.6\%$ to $49.0\%$ for Qwen2.5-Omni(7B), and from $28.7\%$ to $45.4\%$ for Qwen2.5-Omni(3B). If the unprimed model did not engage this circuit, injecting activations from these primed heads should have little effect.

Together, these results show that priming is an analytical device for obtaining reliable successful trajectories, not a source of a separate mechanism. The primed and unprimed regimes engage the same symbolic binding circuit: priming makes this circuit measurable, while unprimed failures arise when the same circuit produces incorrect symbolic IDs.

\subsection{Representational Analysis and Causal Mediation Analysis on Different Models}
\label{app:rep}

\subsubsection{Qwen2.5-Omni(7B)}
\begin{figure}[t]
    \centering
    \begin{subfigure}[b]{0.47\textwidth}
        \centering
        \includegraphics[width=\textwidth]{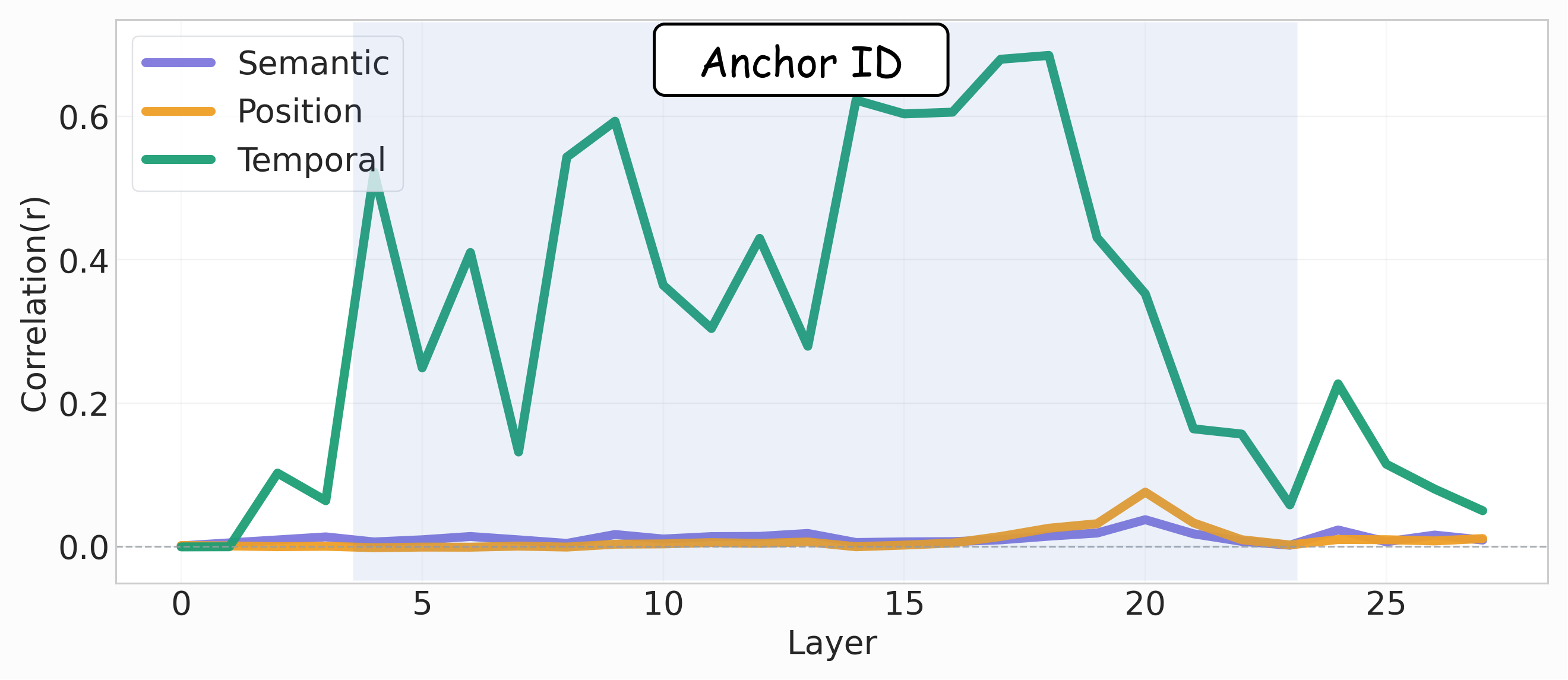} 
        \caption{RSA at anchor attribute token in AAVR task}
    \end{subfigure}
    \hfill 
    \begin{subfigure}[b]{0.47\textwidth}
        \centering
        \includegraphics[width=\textwidth]{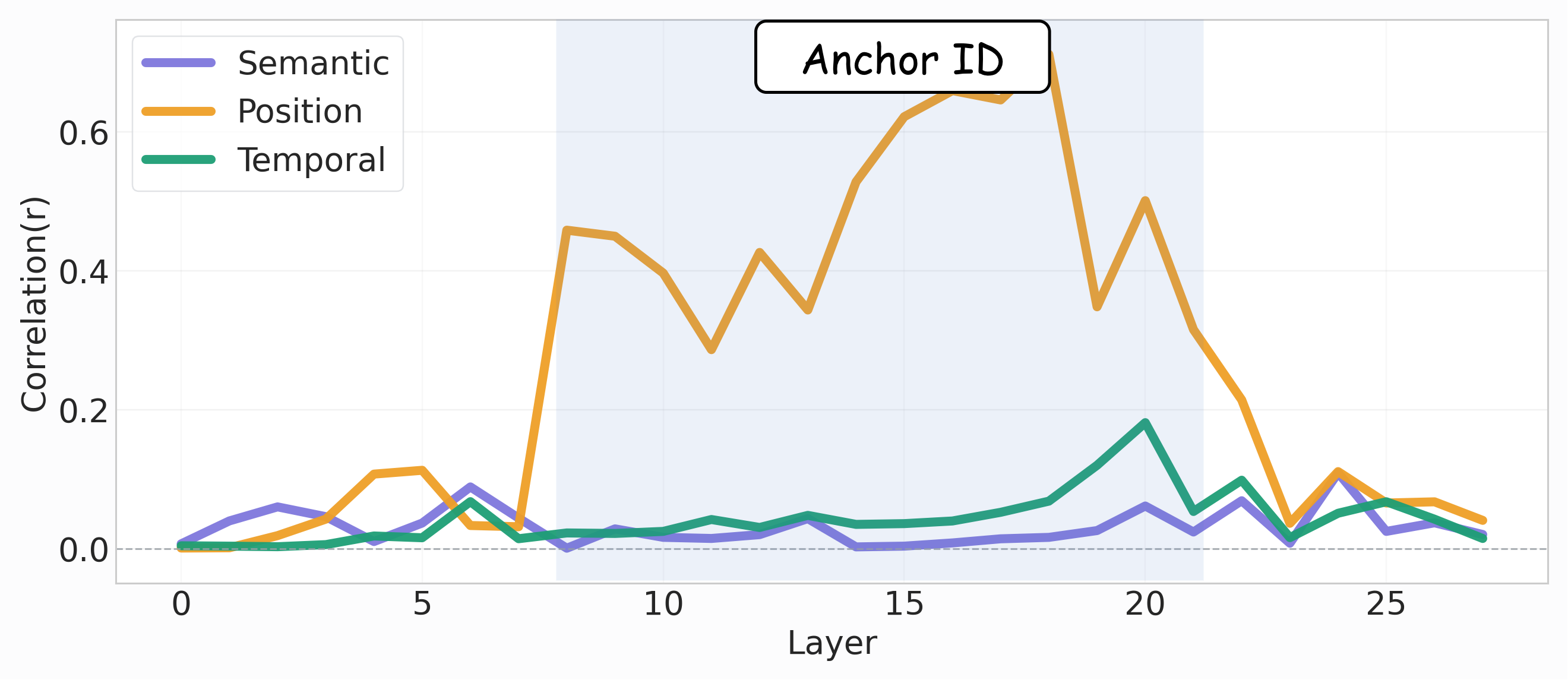} 
         \caption{RSA at anchor attribute token in VAAR task}
    \end{subfigure}
    \vspace{0.1cm} 
    \begin{subfigure}[b]{0.47\textwidth}
        \centering
        \includegraphics[width=\textwidth]{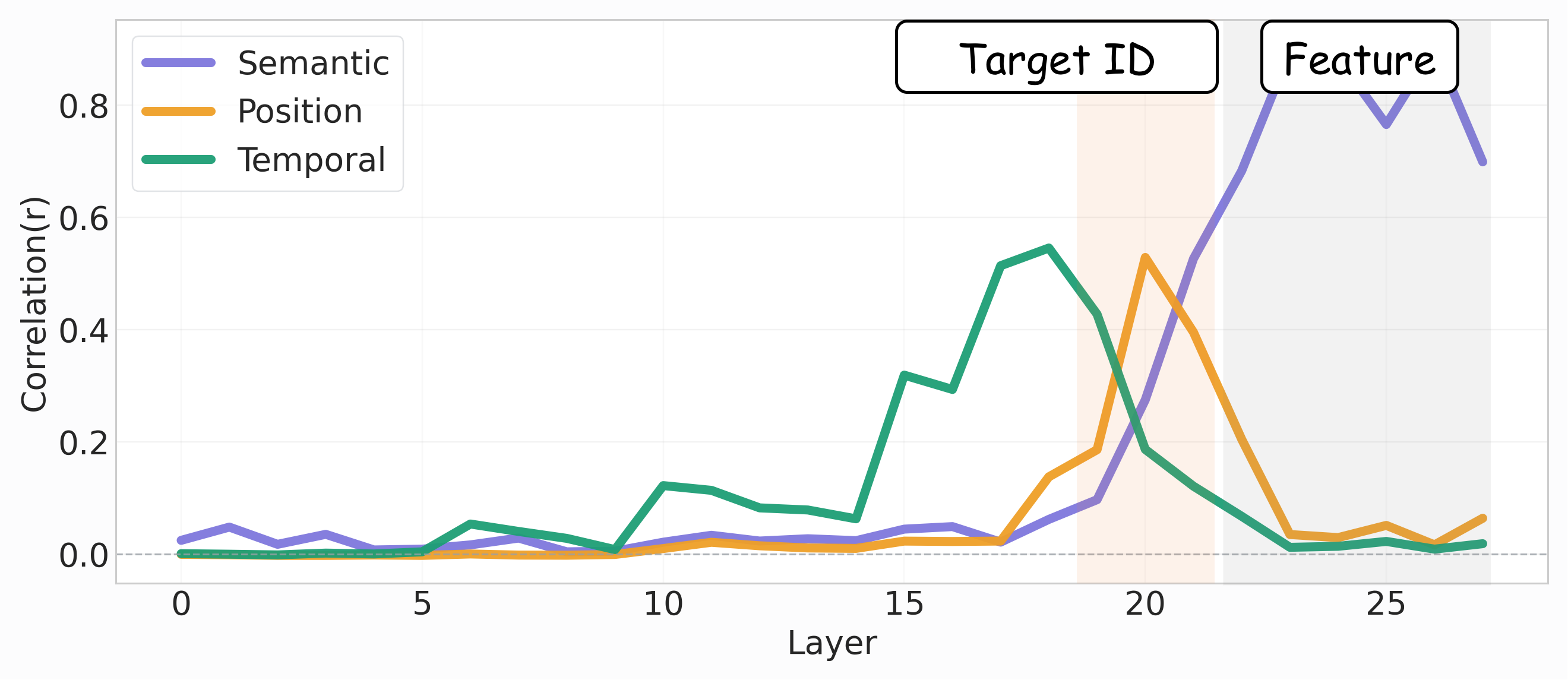} 
       \caption{RSA at last token in AAVR task}
    \end{subfigure}
    \hfill
    \begin{subfigure}[b]{0.47\textwidth}
        \centering
        \includegraphics[width=\textwidth]{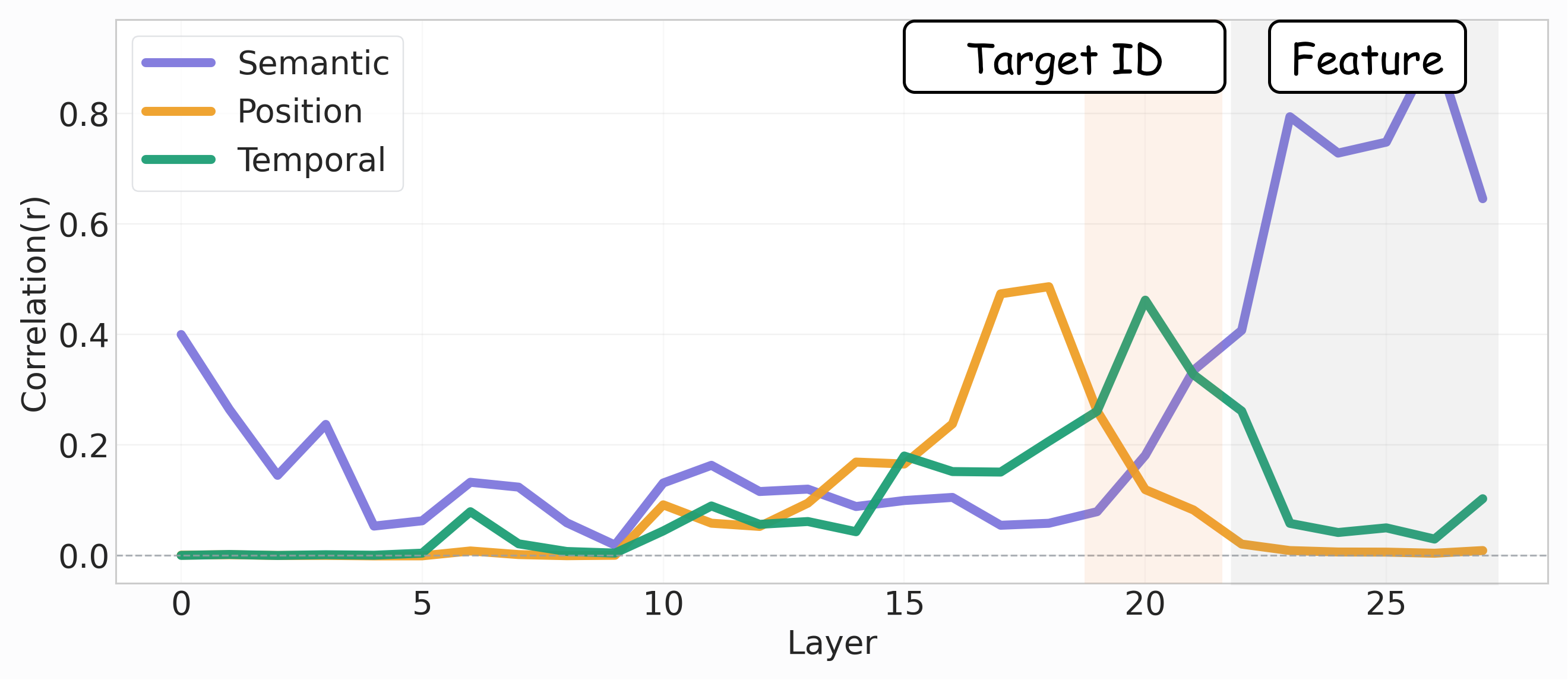} 
       \caption{RSA at last token in VAAR task}
    \end{subfigure}

 \begin{subfigure}[b]{0.47\textwidth}
        \centering
        \includegraphics[width=\textwidth]{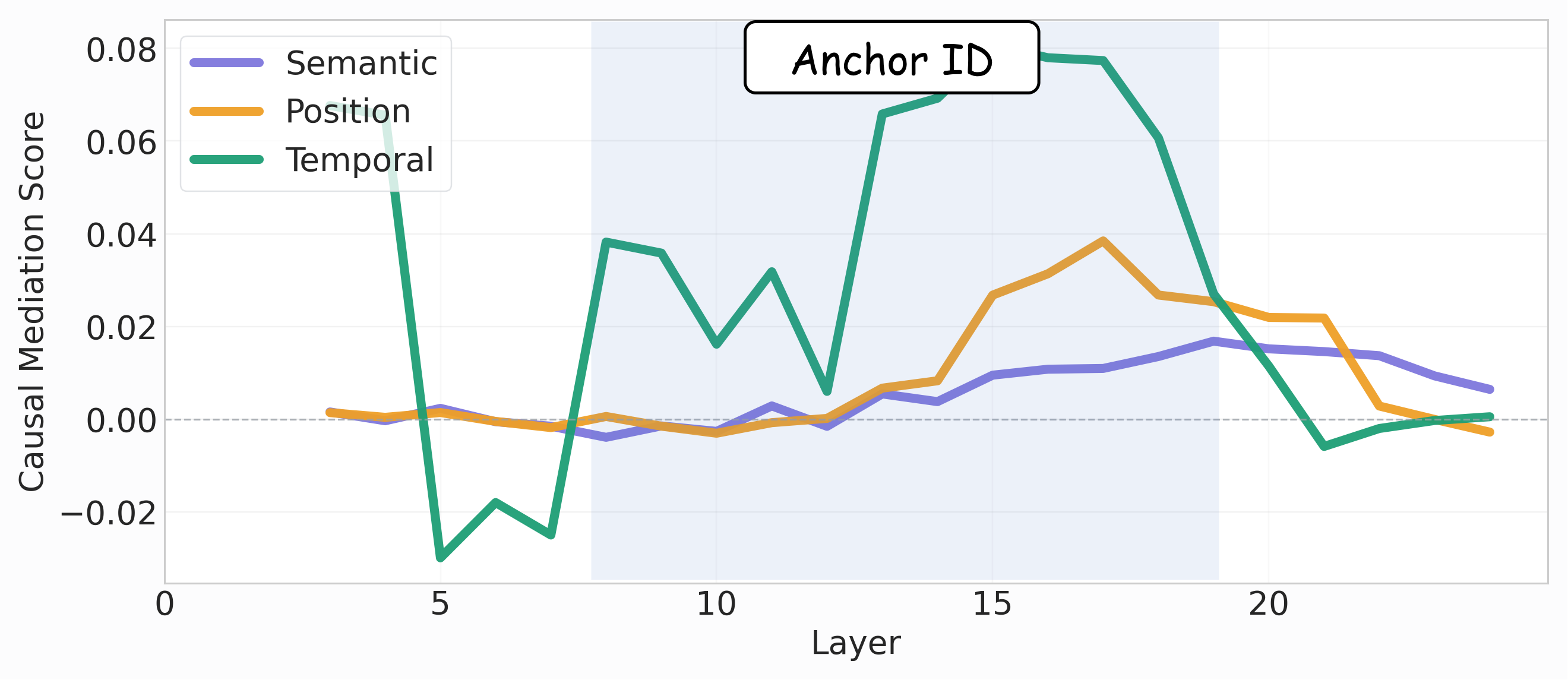} 
        \caption{CMA at anchor attribute token in AAVR task}
    \end{subfigure}
    \hfill 
    \begin{subfigure}[b]{0.47\textwidth}
        \centering
        \includegraphics[width=\textwidth]{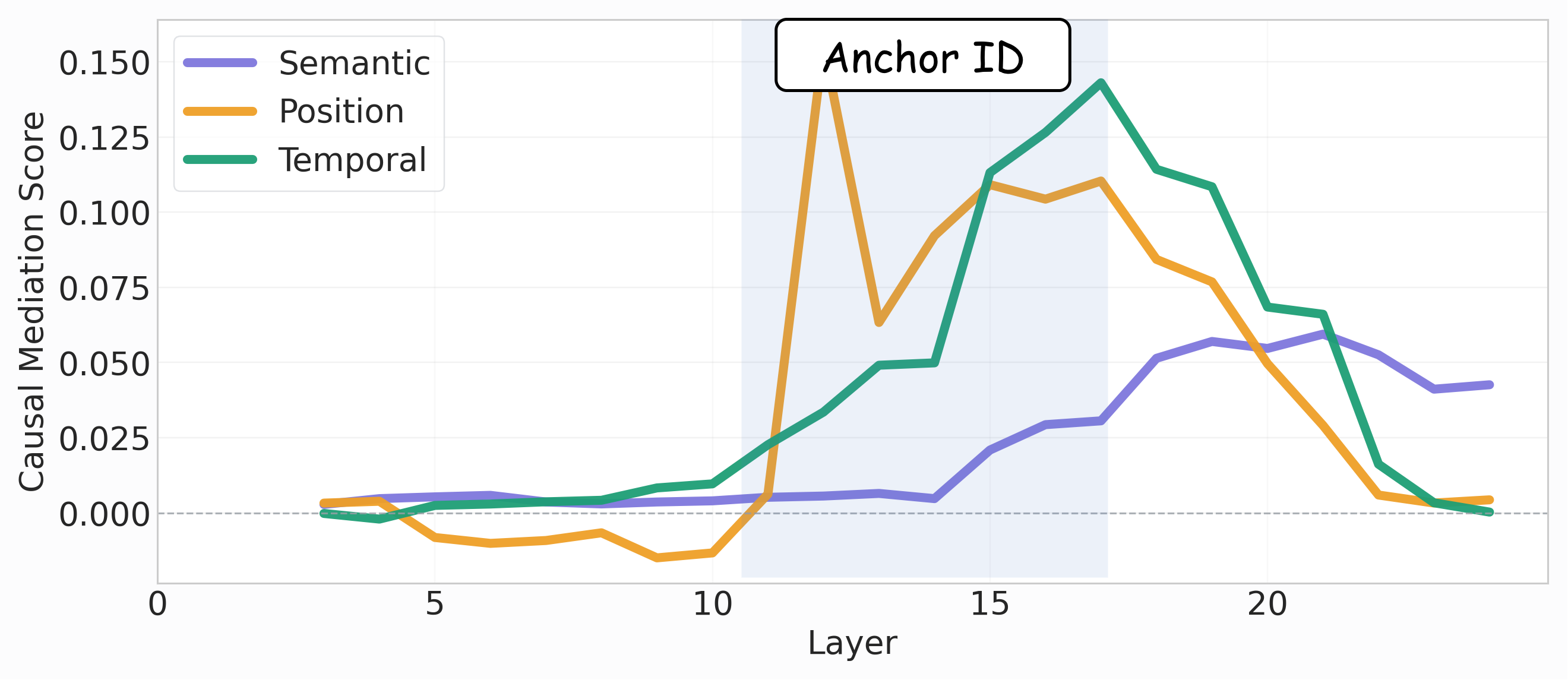} 
        \caption{CMA at anchor attribute token in VAAR task}
    \end{subfigure}
    \vspace{0.1cm} 
    \begin{subfigure}[b]{0.47\textwidth}
        \centering
        \includegraphics[width=\textwidth]{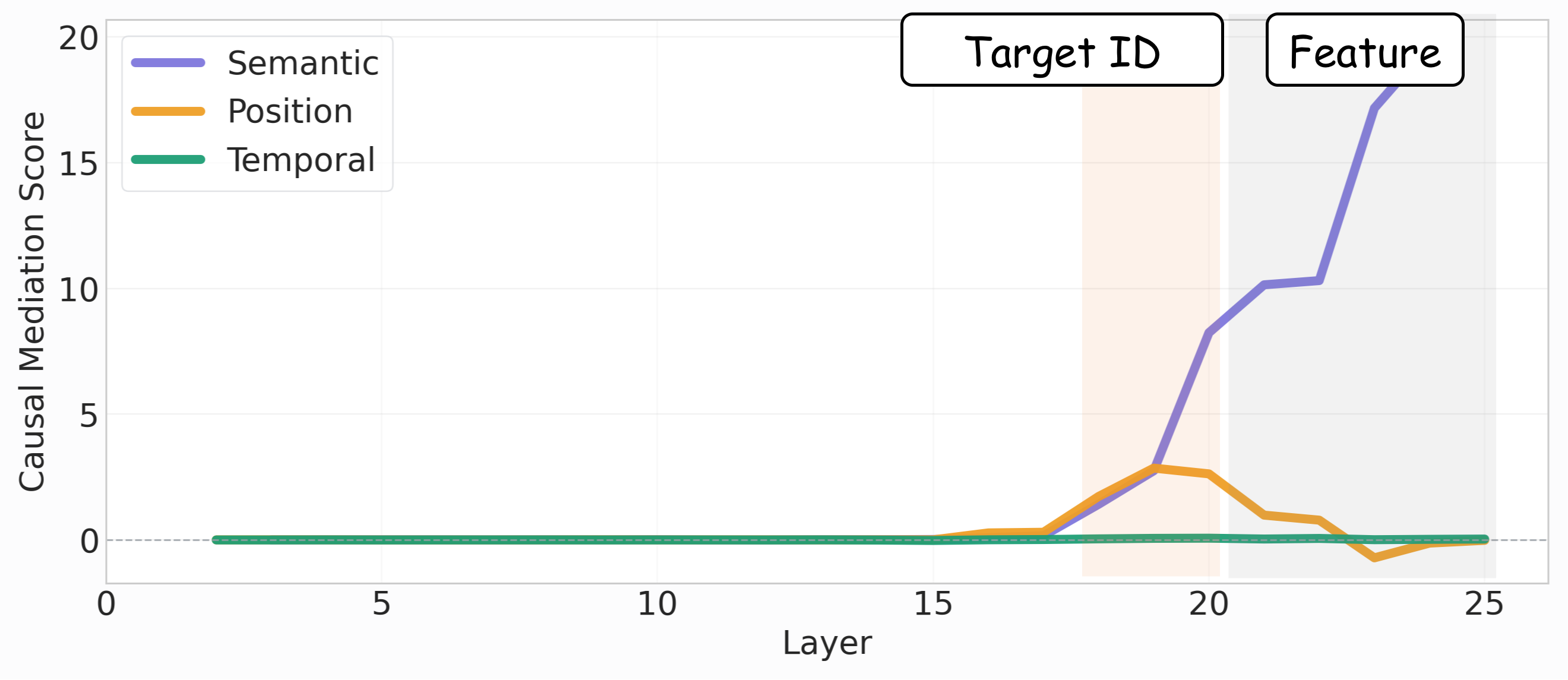} 
       \caption{CMA at last token in AAVR task}
    \end{subfigure}
    \hfill
    \begin{subfigure}[b]{0.47\textwidth}
        \centering
        \includegraphics[width=\textwidth]{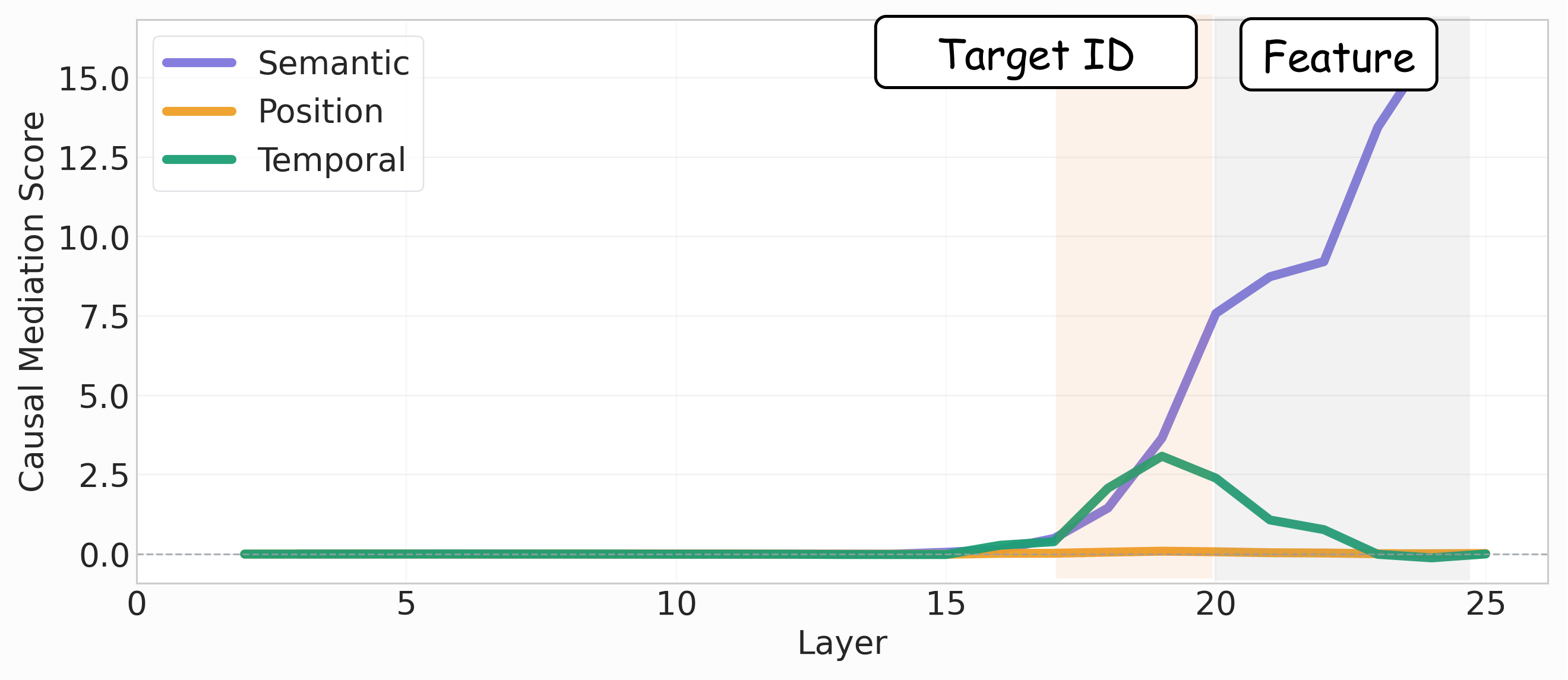} 
       \caption{CMA at last token in VAAR task}
    \end{subfigure}
    \caption{RSA and CMA results for Qwen2.5-Omni(7B)}
    \label{app_fig:qwen7}
\end{figure}

\cref{app_fig:qwen7} illustrates the representation and causal mediation analysis results for Qwen2.5-Omni-7B. These results corroborate our proposed mechanism. A minor distinction from video-SALMONN2+ is observed in the VAAR case, where target ID selection appears to initiate at the layer of prompt tokens. However, this is merely a model-specific variation and remains completely consistent with our proposed three-stage framework (anchor ID retrieval, target ID selection, and feature retrieval). Crucially, our analysis does not posit that Stage 1 occurs exclusively at the prompt token while Stages 2 and 3 occur exclusively at the last token. Rather, it demonstrates that this sequential, three-stage flow manifests layer-wise across the prompt and last tokens collectively.

\subsubsection{Qwen2.5-Omni(3B)}
\begin{figure}[t]
    \centering
    \begin{subfigure}[b]{0.47\textwidth}
        \centering
        \includegraphics[width=\textwidth]{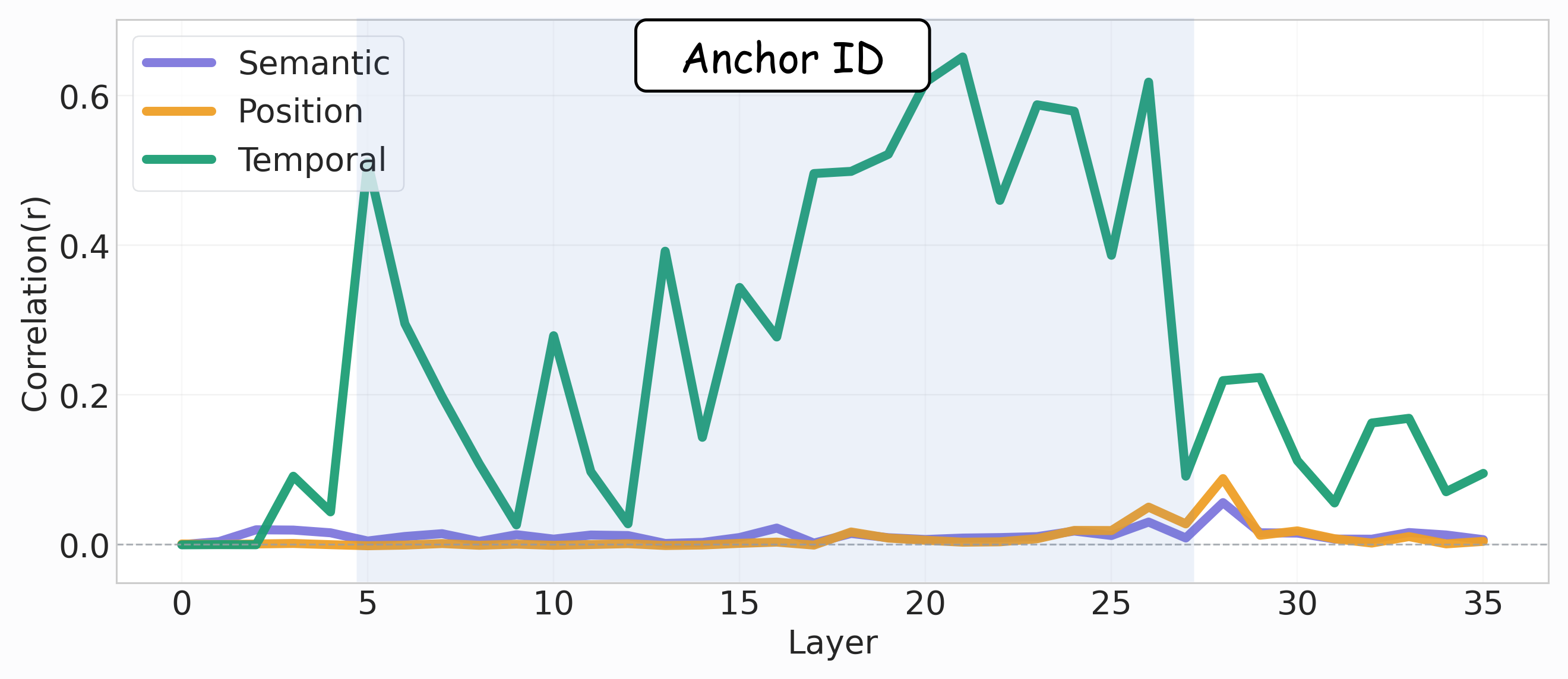} 
        \caption{RSA at anchor attribute token in AAVR task}
    \end{subfigure}
    \hfill 
    \begin{subfigure}[b]{0.47\textwidth}
        \centering
        \includegraphics[width=\textwidth]{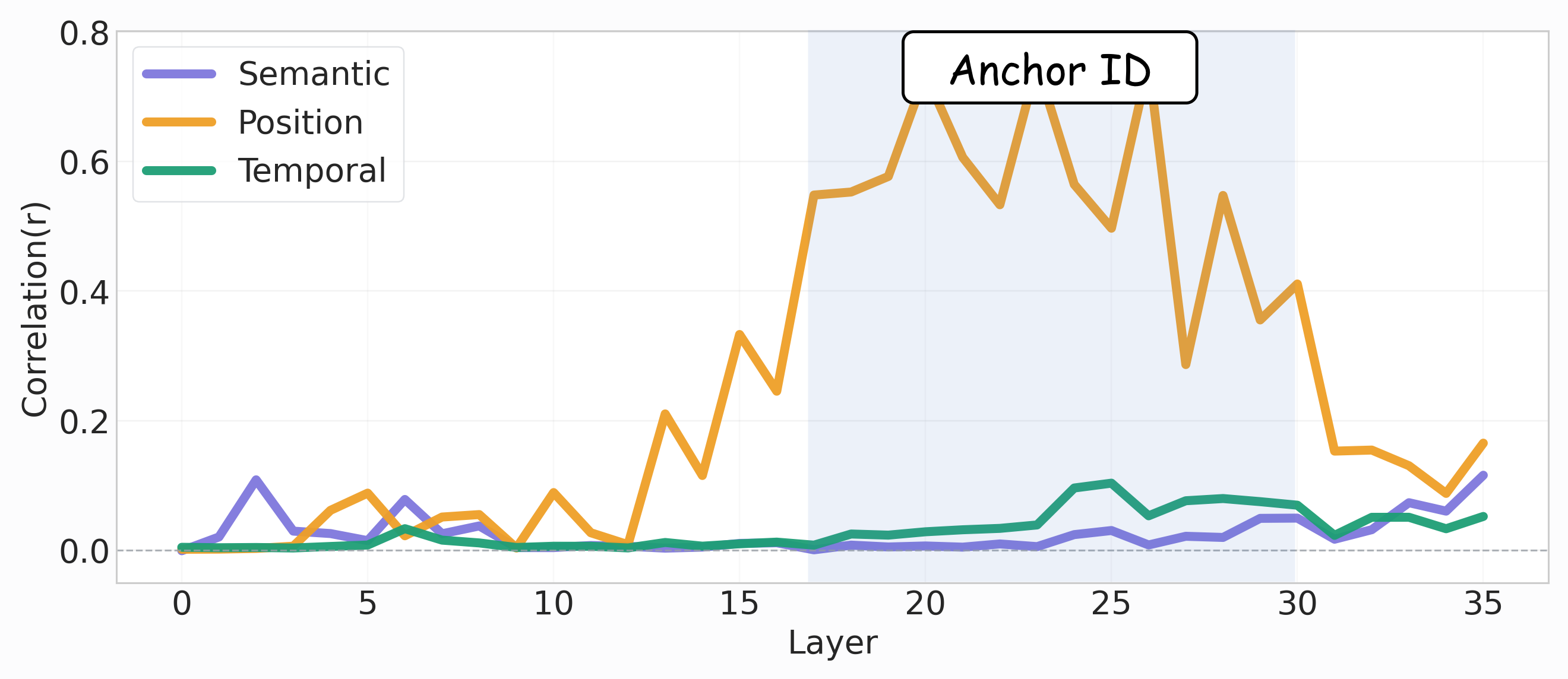} 
        \caption{RSA at anchor attribute token in VAAR task}
    \end{subfigure}
    \vspace{0.1cm} 
    \begin{subfigure}[b]{0.47\textwidth}
        \centering
        \includegraphics[width=\textwidth]{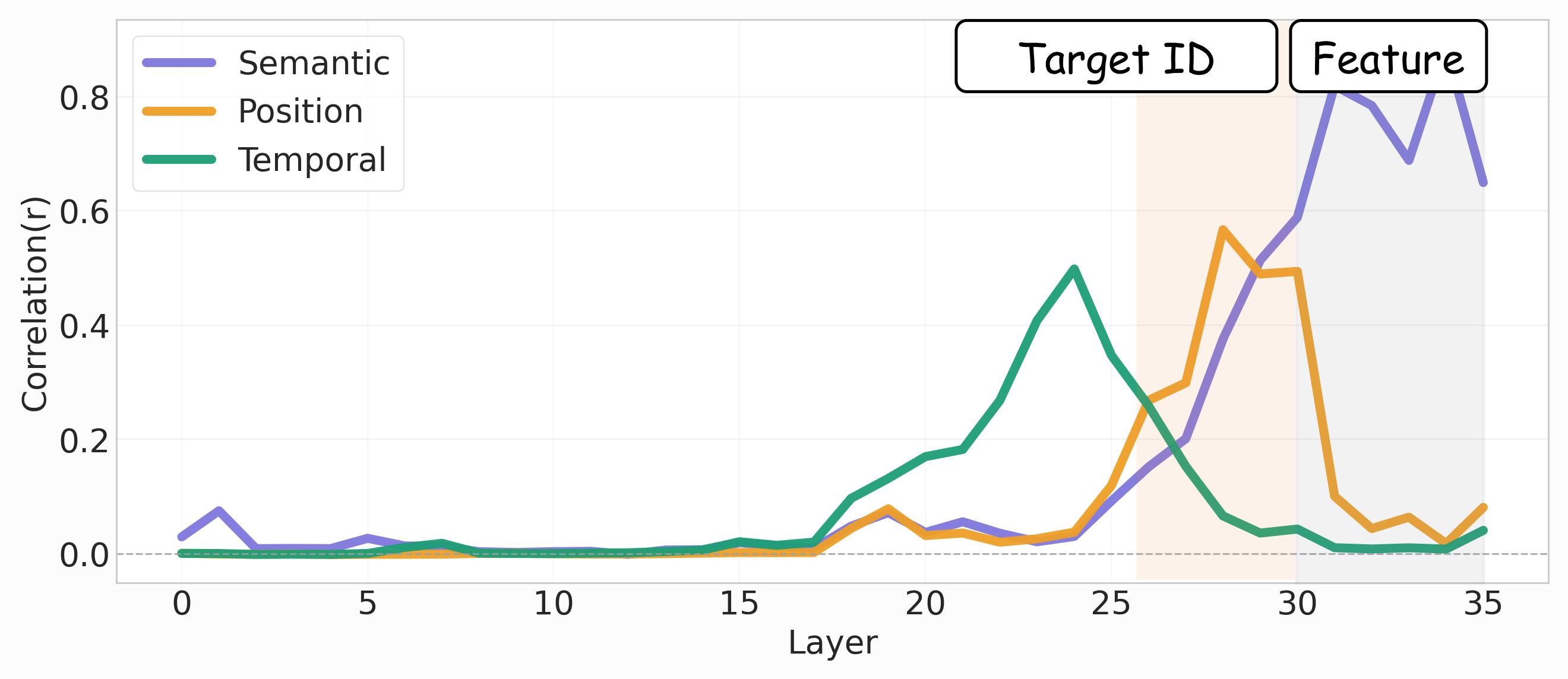} 
       \caption{RSA at last token in AAVR task}
    \end{subfigure}
    \hfill
    \begin{subfigure}[b]{0.47\textwidth}
        \centering
        \includegraphics[width=\textwidth]{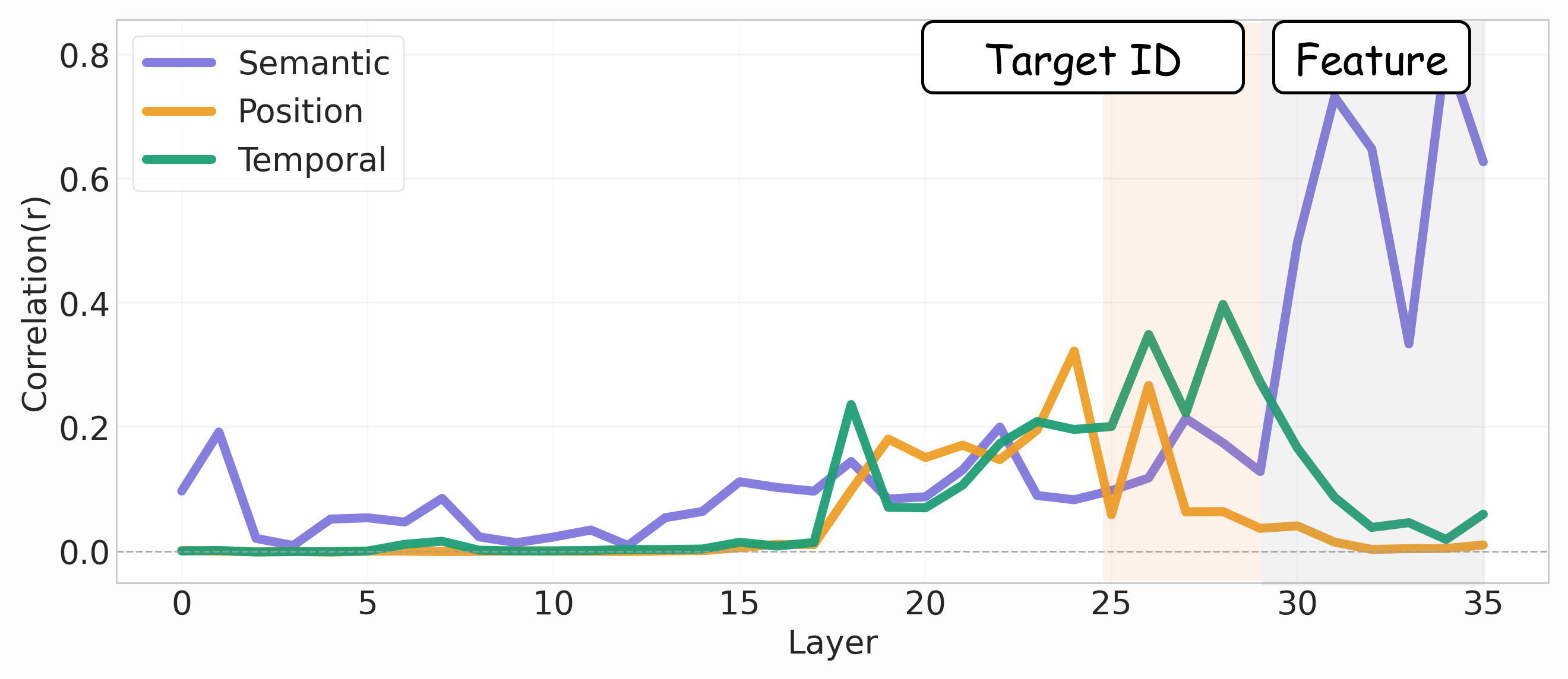}
       \caption{RSA at last token in VAAR task}
    \end{subfigure}

 \begin{subfigure}[b]{0.47\textwidth}
        \centering
        \includegraphics[width=\textwidth]{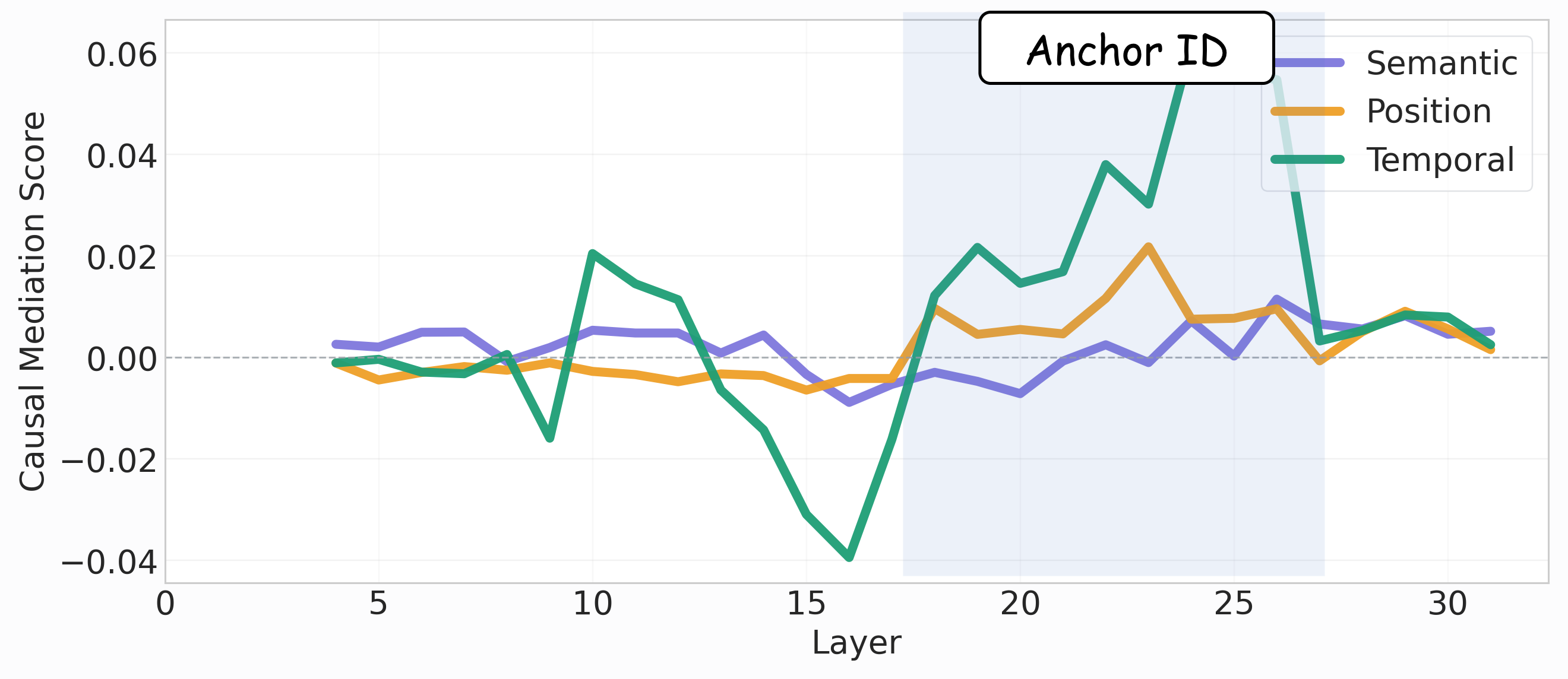} 
       \caption{CMA at anchor attribute token in AAVR task}
    \end{subfigure}
    \hfill 
    \begin{subfigure}[b]{0.47\textwidth}
        \centering
        \includegraphics[width=\textwidth]{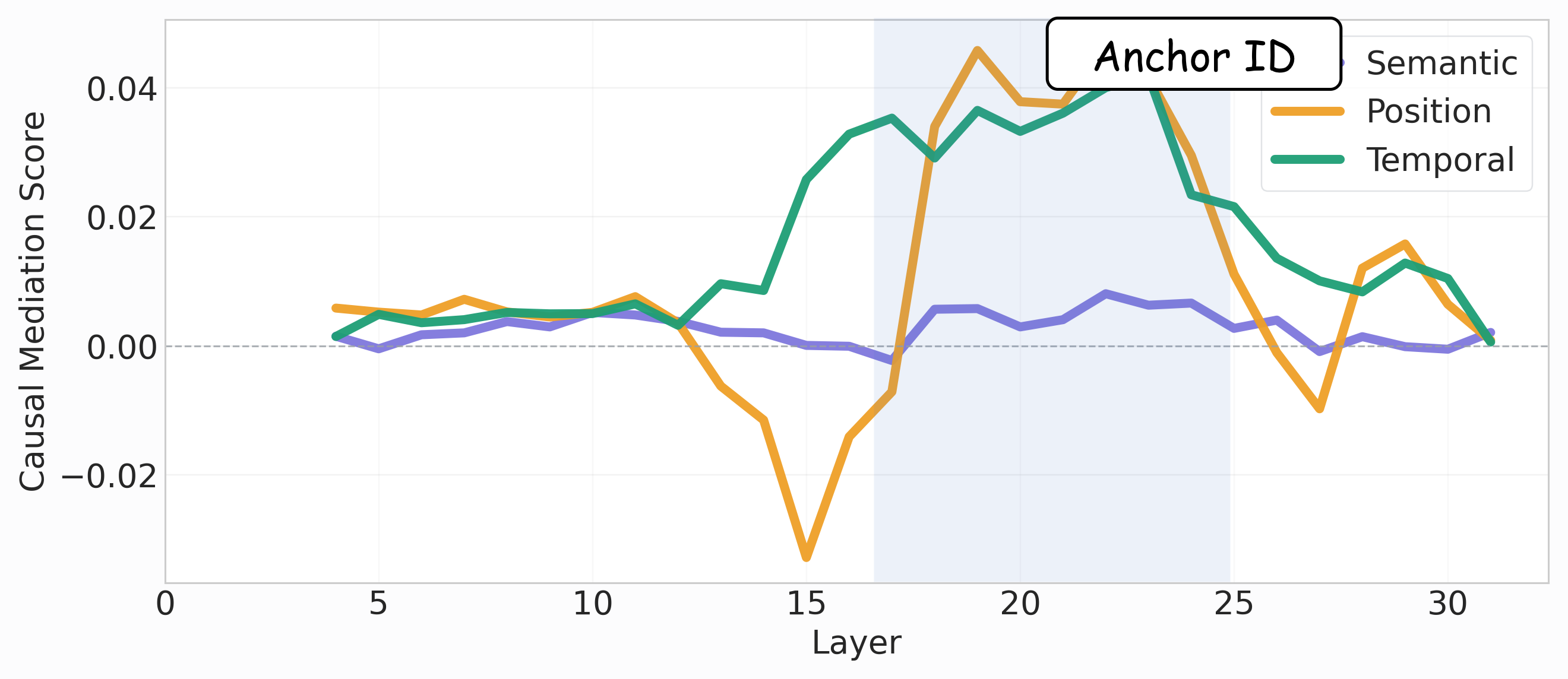}
         \caption{CMA at anchor attribute token in VAAR task}
    \end{subfigure}
    \vspace{0.1cm} 
    \begin{subfigure}[b]{0.47\textwidth}
        \centering
        \includegraphics[width=\textwidth]{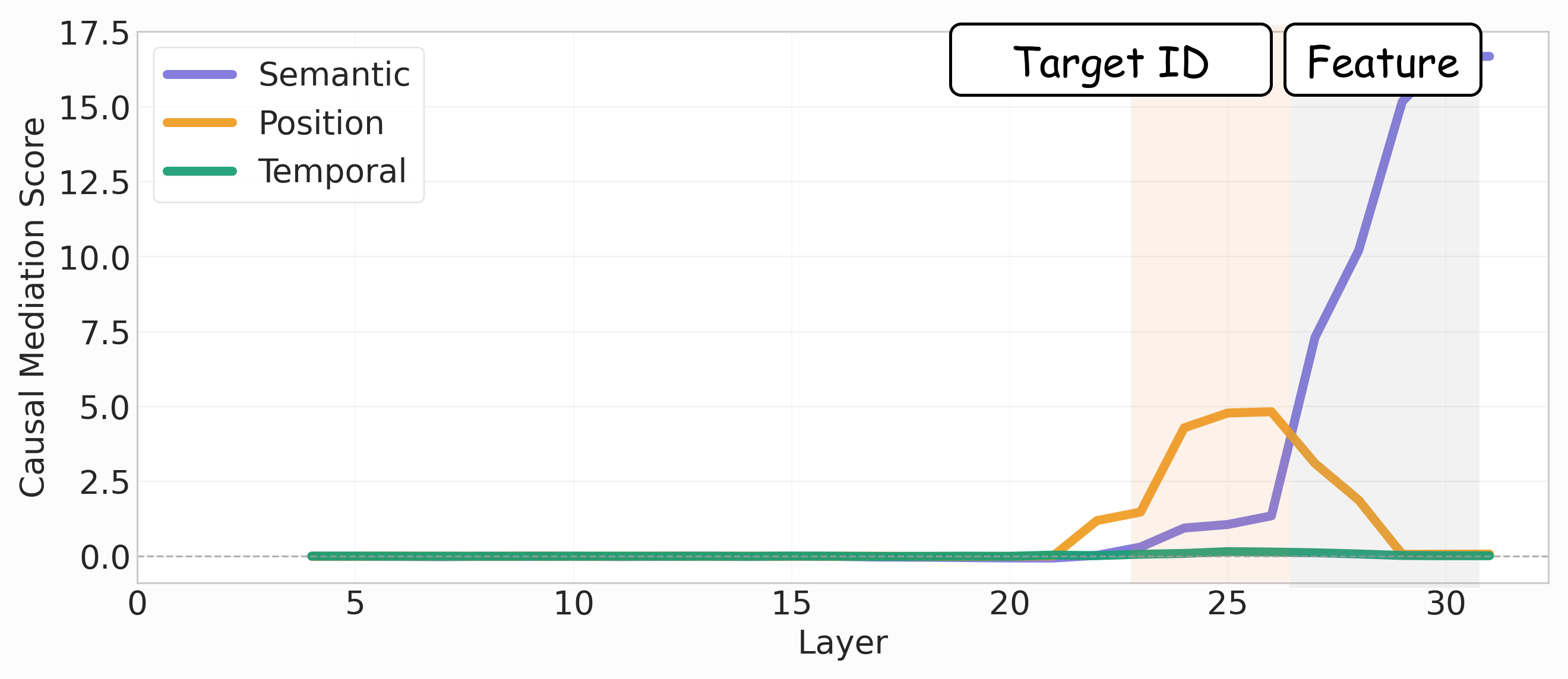}
      \caption{CMA at last token in AAVR task}
    \end{subfigure}
    \hfill
    \begin{subfigure}[b]{0.47\textwidth}
        \centering
        \includegraphics[width=\textwidth]{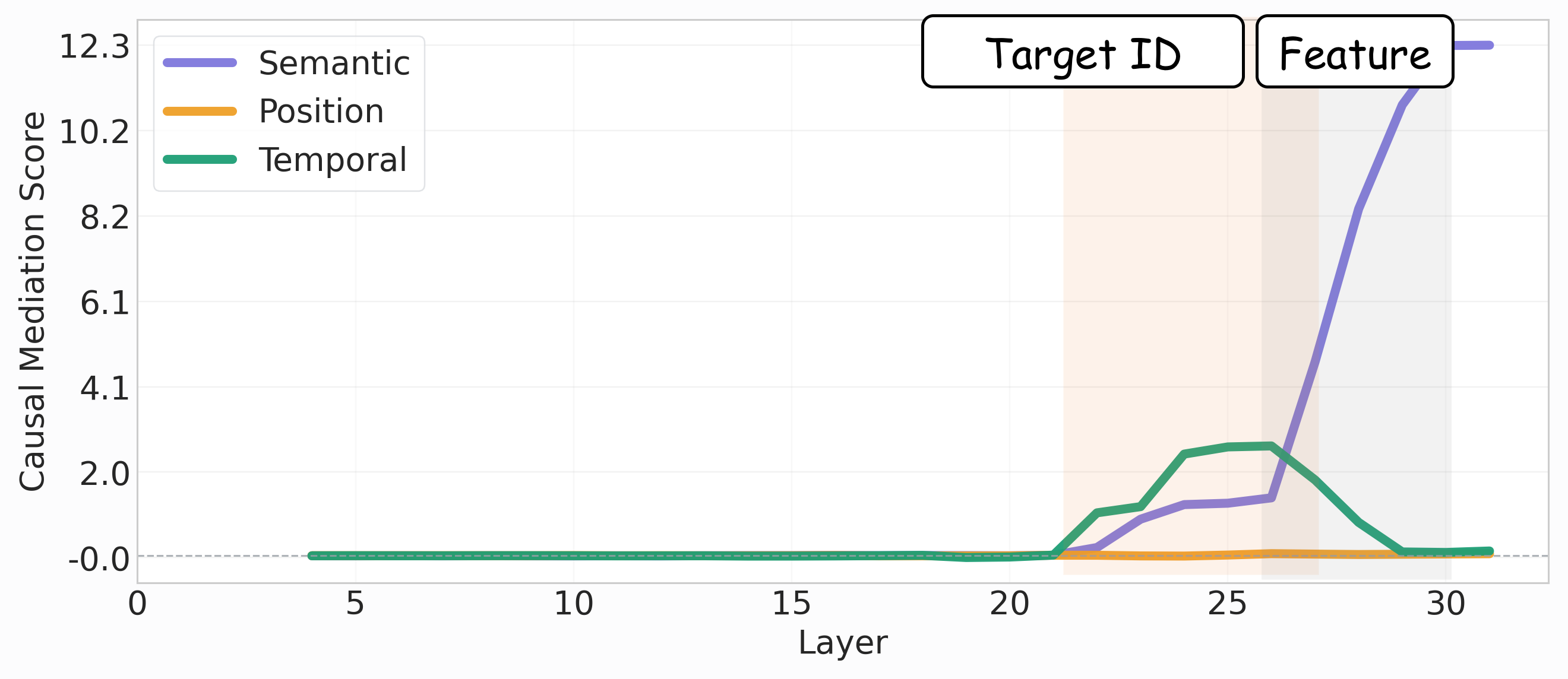}
       \caption{CMA at last token in VAAR task}
    \end{subfigure}
    \caption{RSA and CMA results for Qwen2.5-Omni(3B)}
    \label{app_fig:qwen3}
    
\end{figure}

\cref{app_fig:qwen3} presents the RSA and CMA results for Qwen2.5-Omni(3B), which exhibit trends consistent with our proposed three-stage mechanism. At the anchor attribute token, the temporal ID dominates in the AAVR task while the position ID dominates in the VAAR task across the mid-to-late layers, indicating anchor ID retrieval. At the last token, this pattern inverts in the late layers-position information becomes prominent in AAVR and temporal information in VAAR-consistent with target ID selection, before semantic information takes over in the deepest layers, marking feature retrieval. The CMA results corroborate these findings: anchor ID manipulation yields the largest causal effect at the anchor token in mid-to-late layers, target ID manipulation peaks at the last token in late layers, and semantic content manipulation dominates in the deepest layers. 

\subsubsection{MiniCPM-o-4.5}
\begin{figure}[t]
    \centering
    \begin{subfigure}[b]{0.47\textwidth}
        \centering
        \includegraphics[width=\textwidth]{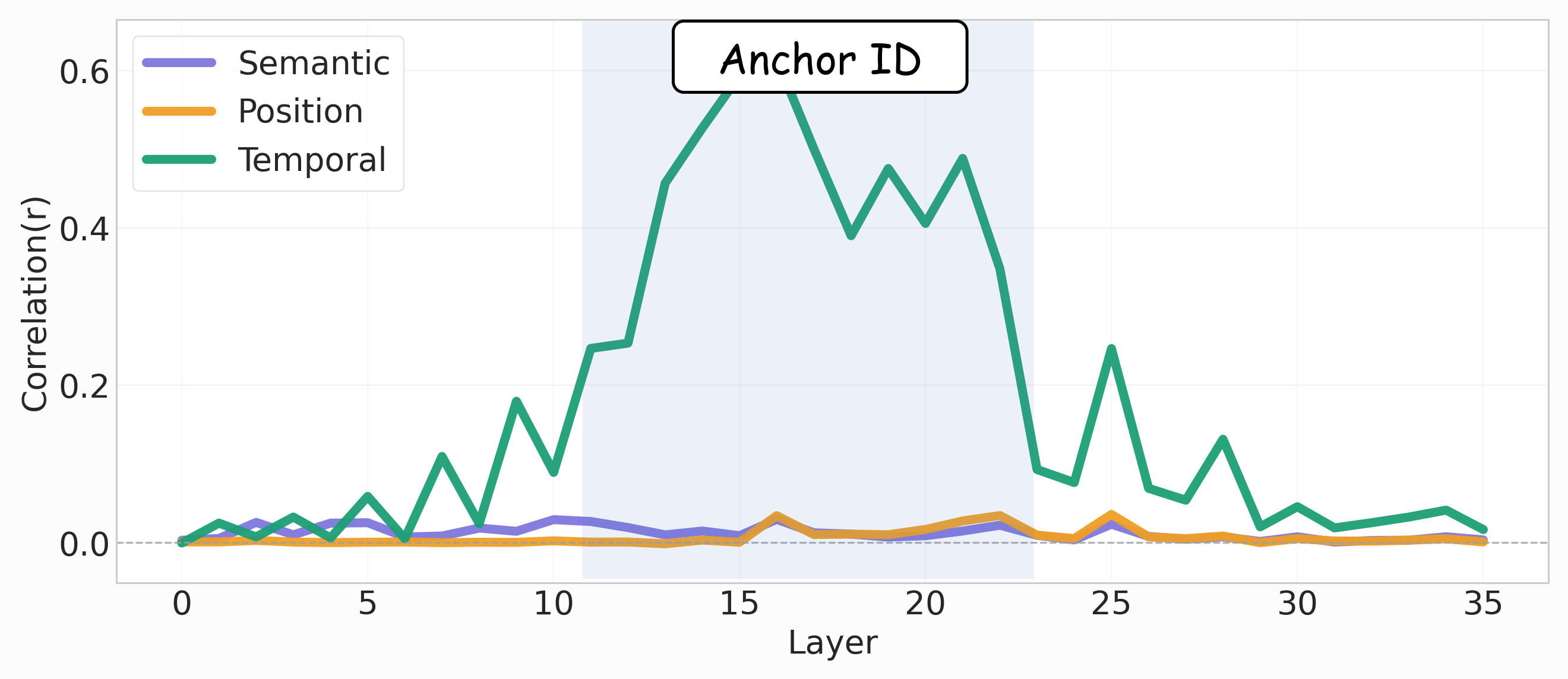} 
        \caption{Prompt tokens in AAVR case}
    \end{subfigure}
    \hfill 
    \begin{subfigure}[b]{0.47\textwidth}
        \centering
        \includegraphics[width=\textwidth]{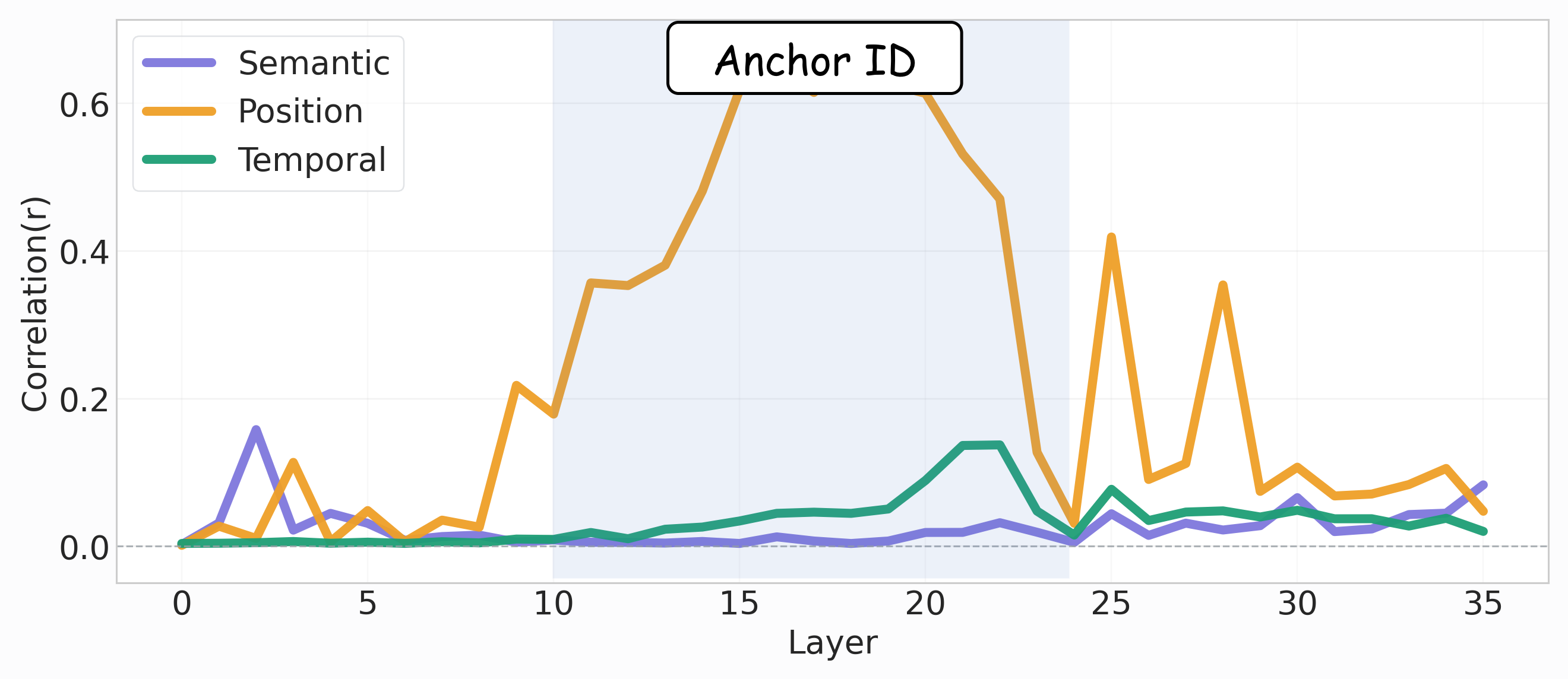}
        \caption{Prompt tokens in VAAR case}
    \end{subfigure}
    \vspace{0.1cm} 
    \begin{subfigure}[b]{0.47\textwidth}
        \centering
        \includegraphics[width=\textwidth]{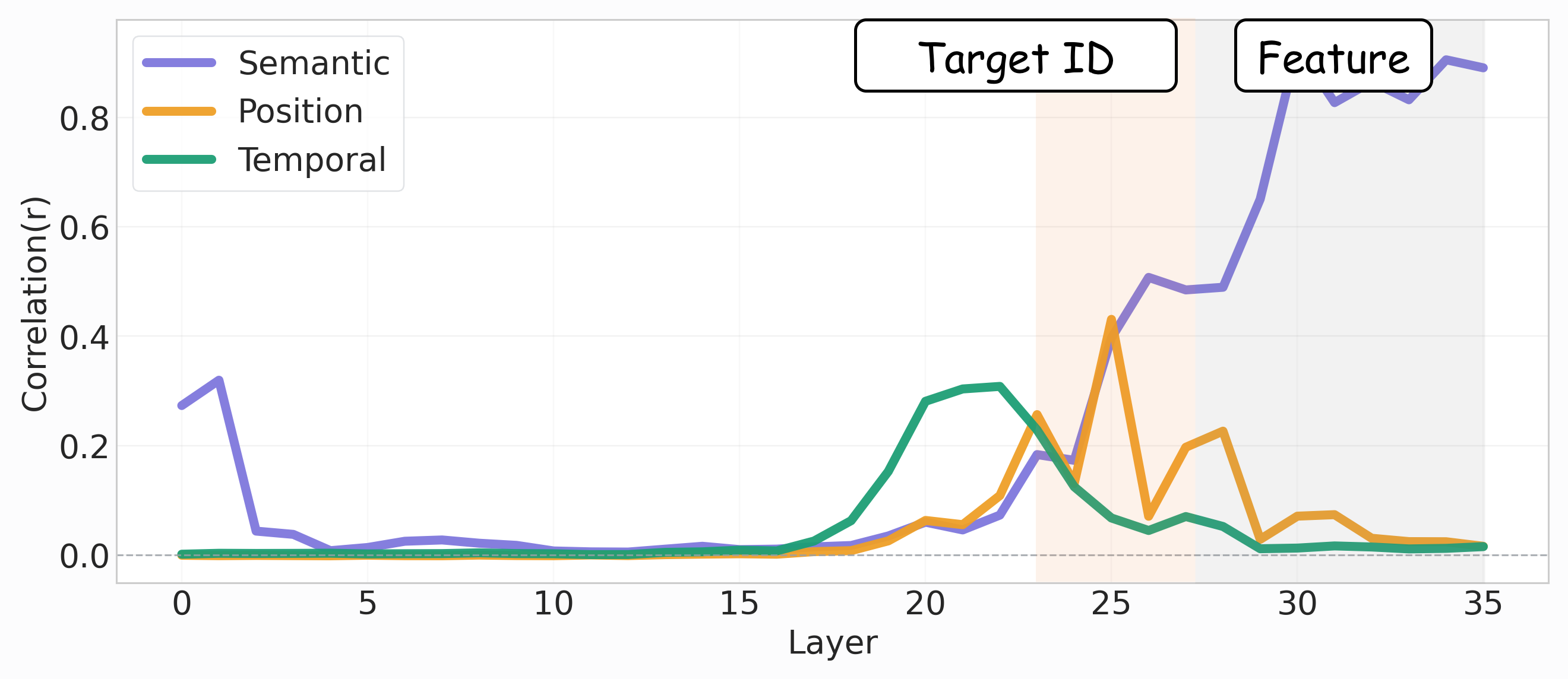}
       \caption{Last tokens in AAVR case}
    \end{subfigure}
    \hfill
    \begin{subfigure}[b]{0.47\textwidth}
        \centering
        \includegraphics[width=\textwidth]{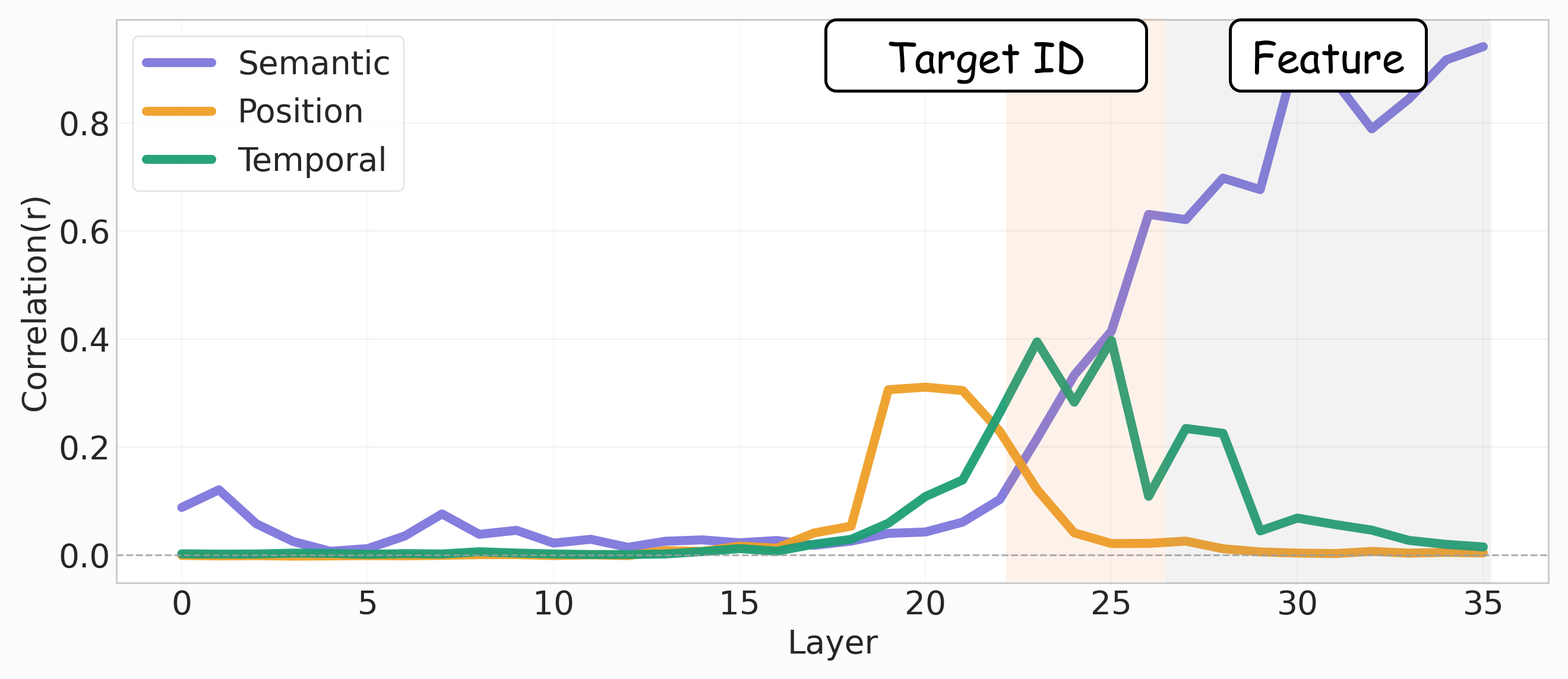} 
       \caption{Last tokens in VAAR case}
    \end{subfigure}

 \begin{subfigure}[b]{0.47\textwidth}
        \centering
        \includegraphics[width=\textwidth]{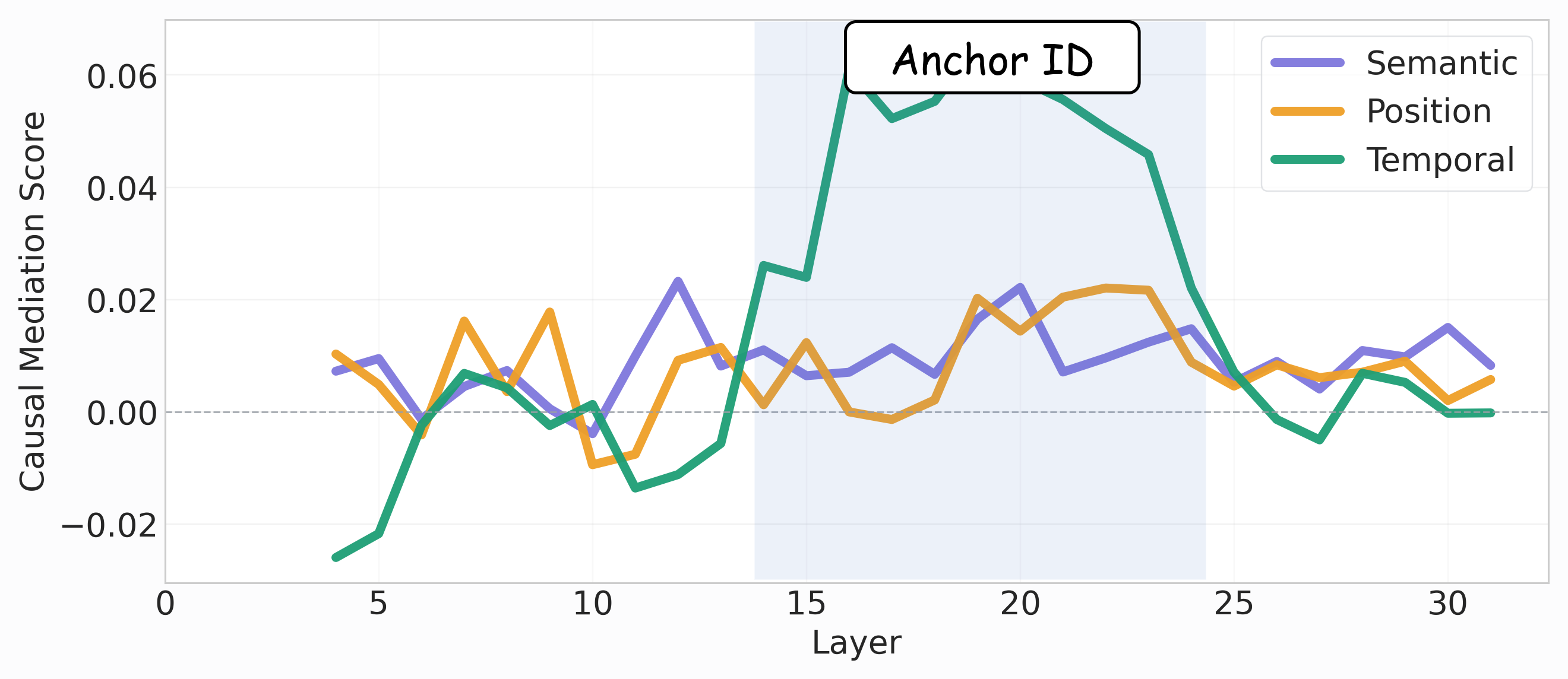} 
        \caption{Prompt tokens in AAVR case}
    \end{subfigure}
    \hfill 
    \begin{subfigure}[b]{0.47\textwidth}
        \centering
        \includegraphics[width=\textwidth]{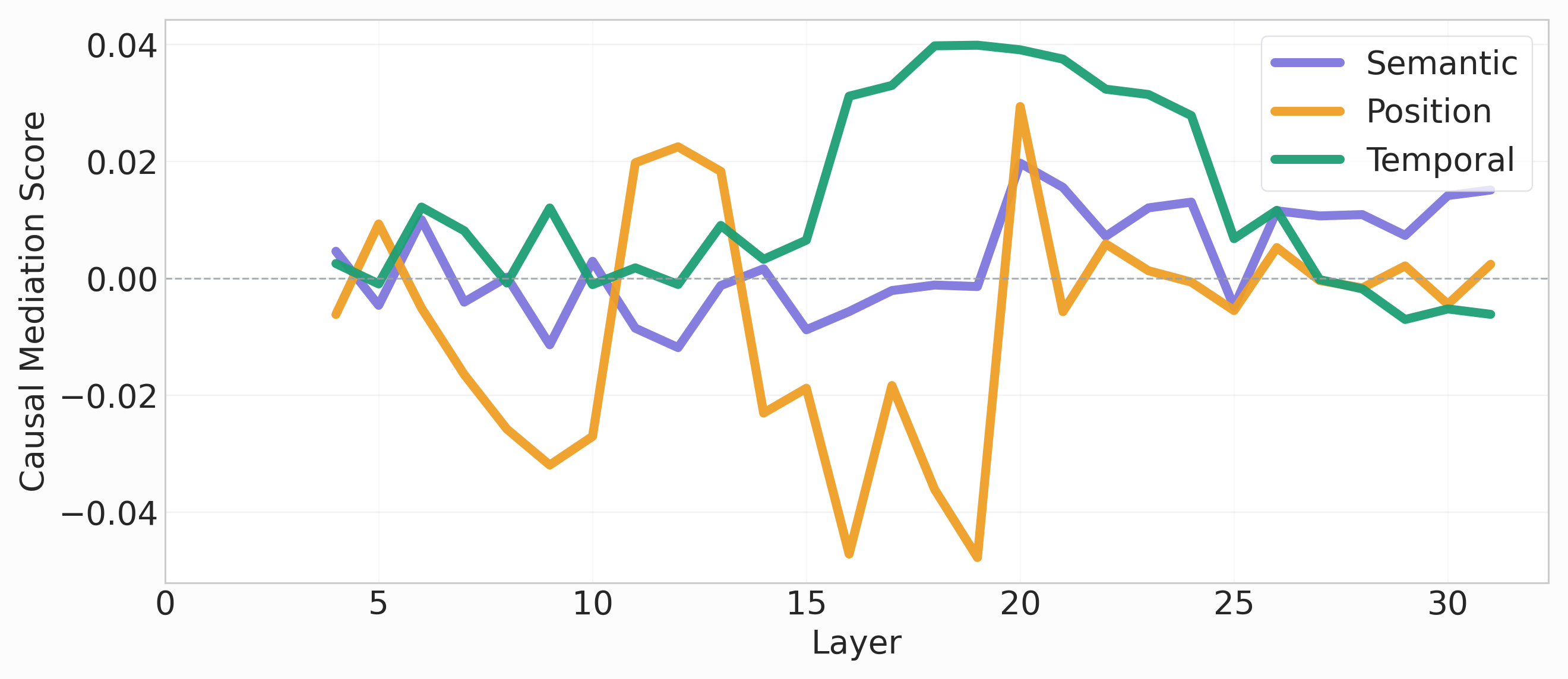} 
        \caption{Prompt tokens in VAAR case}
    \end{subfigure}
    \vspace{0.1cm} 
    \begin{subfigure}[b]{0.47\textwidth}
        \centering
        \includegraphics[width=\textwidth]{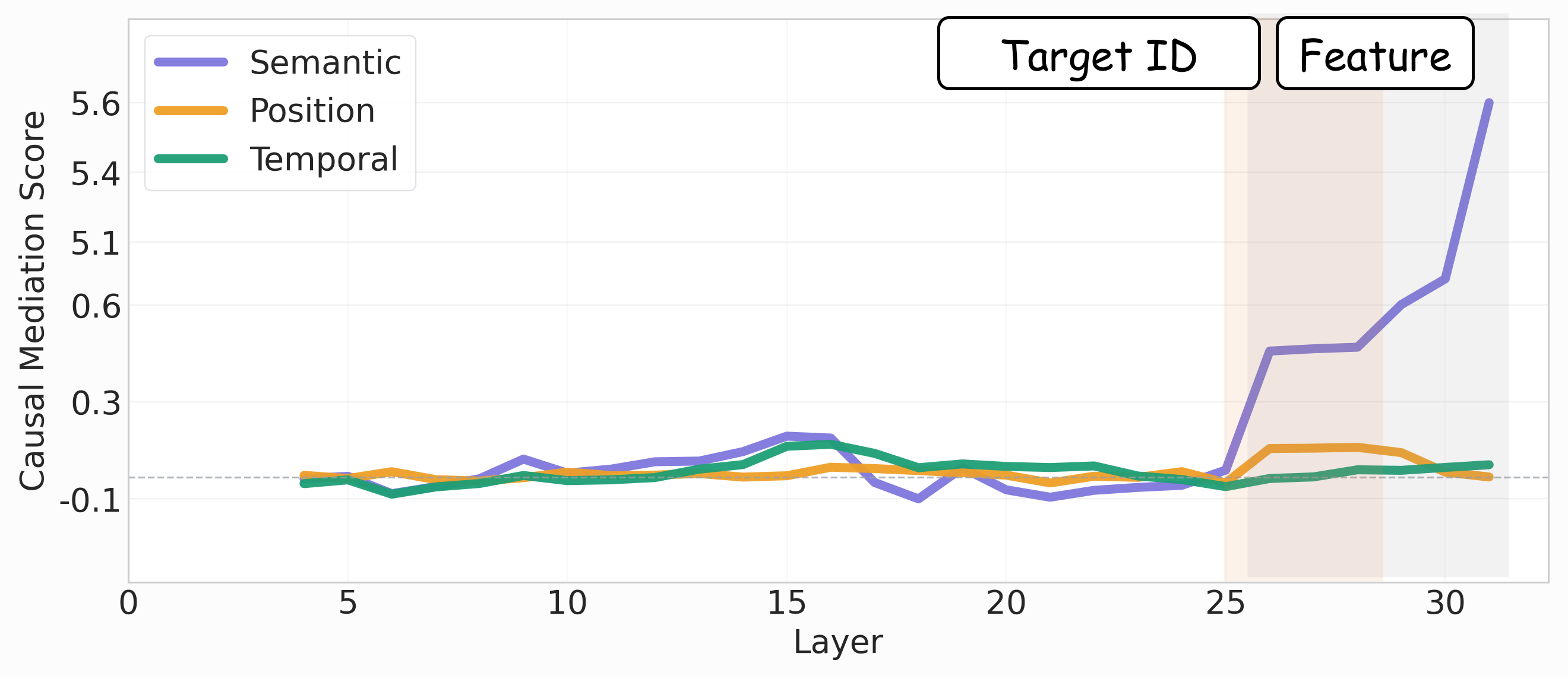} 
       \caption{Last tokens in AAVR case}
    \end{subfigure}
    \hfill
    \begin{subfigure}[b]{0.47\textwidth}
        \centering
        \includegraphics[width=\textwidth]{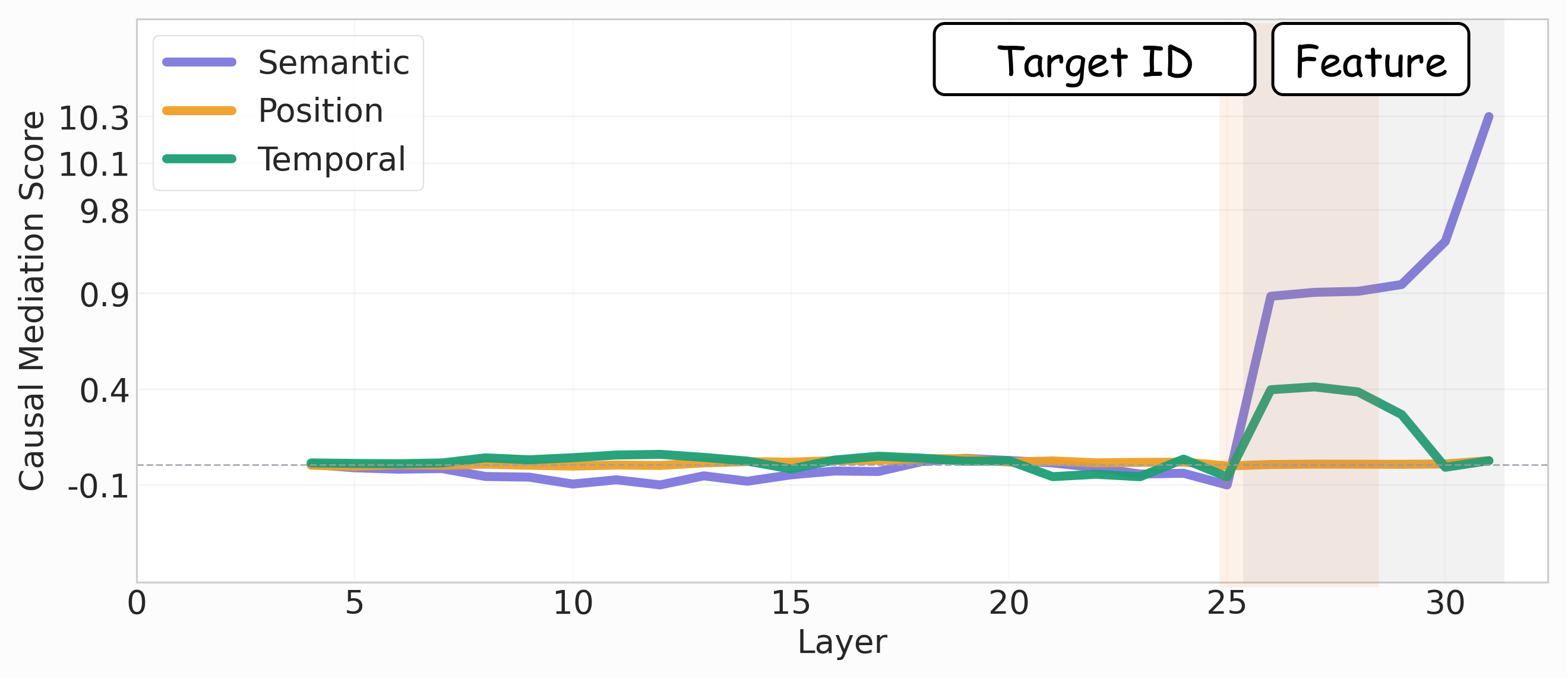} 
       \caption{Last tokens in VAAR case}
    \end{subfigure}
    \caption{RSA and CMA results for MiniCPM-o-4.5(9B)}
    \label{app_fig:minicpm}
\end{figure}

\cref{app_fig:minicpm} reports the RSA and CMA results for MiniCPM-o-4.5 (9B), which again align with our proposed mechanism. At the prompt token, the modality-specific anchor ID (temporal for AAVR, position for VAAR) emerges as the dominant signal in the mid-to-late layers, reflecting anchor ID retrieval. At the last token, the target ID takes over in the late layers, followed by semantic content in the deepest layers, consistent with the target ID selection and feature retrieval stages. The CMA results largely echo these representational patterns, with only minor deviations at certain stages. Overall, MiniCPM-o-4.5 suggests that the symbolic trimodal binding mechanism generalizes across model families and parameter scales.

\subsection{Generalization of Binding Mechanism Beyond the Controlled Four-Speaker Setting}
\label{app:gen_more_spk} We further extend the synthetic analysis to five- and six-speaker settings.

\smallskip\noindent\textbf{Beyond the $2\times2$ grid, with acoustic noise.}
To demonstrate that our trimodal binding mechanism generalizes beyond the $2\times2$ grid, we applied Sec.~3.3 RSA to five- and six-speaker settings, including five-speaker variants with MUSAN noise~\cite{snyder2015musan} at 20 and 10\,dB SNR to approximate realistic noisy conditions. In the five-speaker setting, three speakers are arranged in the top row and two in the bottom row; in the six-speaker setting, three speakers are arranged in each row. For each hypothesized representational space---semantic content, Position IDs, and Temporal IDs---we report the maximum correlation and its corresponding layer. Table~\ref{tab:multispk} reports video-SALMONN2+ (7B) results and aligns with the proposed mechanism. At the anchor token in the middle layers, the modality-specific anchor ID (Temporal ID for AAVR; Position ID for VAAR) dominates (anchor ID retrieval). At the last prompt token, the complementary target ID (Position ID for AAVR; Temporal ID for VAAR) becomes dominant in later layers (target ID selection)---followed by semantic information in the deepest layers (feature retrieval).

\begin{table}[h]
\centering
\scriptsize
\setlength{\tabcolsep}{3.5pt}
\renewcommand{\arraystretch}{1.05}
\caption{RSA in five-/six-speaker settings (video-SALMONN2+, 7B). Each cell reports the peak correlation and its layer. \textbf{Bold}: highest among the three spaces; \underline{underline}: second highest.}
\label{tab:multispk}
\begin{tabular}{llccc|ccc}
\toprule
\multirow{2}{*}{\textbf{Data}} & \multirow{2}{*}{\textbf{Token}}
& \multicolumn{3}{c|}{\textbf{AAVR}} & \multicolumn{3}{c}{\textbf{VAAR}} \\
\cmidrule(lr){3-5}\cmidrule(l){6-8}
& & Semantic & Position & Temporal & Semantic & Position & Temporal \\
\midrule
5spk & Anchor & 0.03@L20 & \underline{0.07@L20} & \textbf{0.52@L18} & 0.15@L20 & \textbf{0.66@L18} & \underline{0.17@L20} \\
5spk & Last   & \textbf{0.83@L23} & \underline{0.47@L20} & 0.27@L18 & \textbf{0.81@L23} & 0.32@L18 & \underline{0.34@L20} \\
\addlinespace[2pt]
6spk & Anchor & 0.03@L20 & \underline{0.06@L20} & \textbf{0.45@L18} & 0.09@L20 & \textbf{0.57@L18} & \underline{0.10@L20} \\
6spk & Last   & \textbf{0.65@L24} & \underline{0.34@L20} & 0.22@L18 & \textbf{0.74@L23} & \underline{0.32@L18} & 0.26@L20 \\
\addlinespace[2pt]
5spk, 20dB & Anchor & 0.04@L20 & \underline{0.08@L20} & \textbf{0.52@L18} & 0.15@L20 & \textbf{0.66@L18} & \underline{0.18@L20} \\
5spk, 20dB & Last   & \textbf{0.82@L23} & \underline{0.47@L20} & 0.24@L18 & \textbf{0.79@L23} & 0.32@L18 & \underline{0.34@L20} \\
\addlinespace[2pt]
5spk, 10dB & Anchor & 0.04@L20 & \underline{0.08@L20} & \textbf{0.52@L18} & 0.15@L20 & \textbf{0.66@L18} & \underline{0.18@L20} \\
5spk, 10dB & Last   & \textbf{0.82@L23} & \underline{0.47@L20} & 0.23@L18 & \textbf{0.78@L23} & 0.31@L18 & \underline{0.33@L20} \\
\bottomrule
\end{tabular}
\end{table}

These results show that the trimodal binding mechanism is not merely an artifact of the specific $2\times2$ grid structure, but generalizes across broader synthetic and real-world settings.

\subsection{Representation Analysis on Different Models on Primed vs. Unprimed Settings}
\label{app:failure_rep}

\begin{figure}[t]
    \centering
    \begin{subfigure}{0.5\linewidth}
        \centering
        \includegraphics[width=\linewidth]{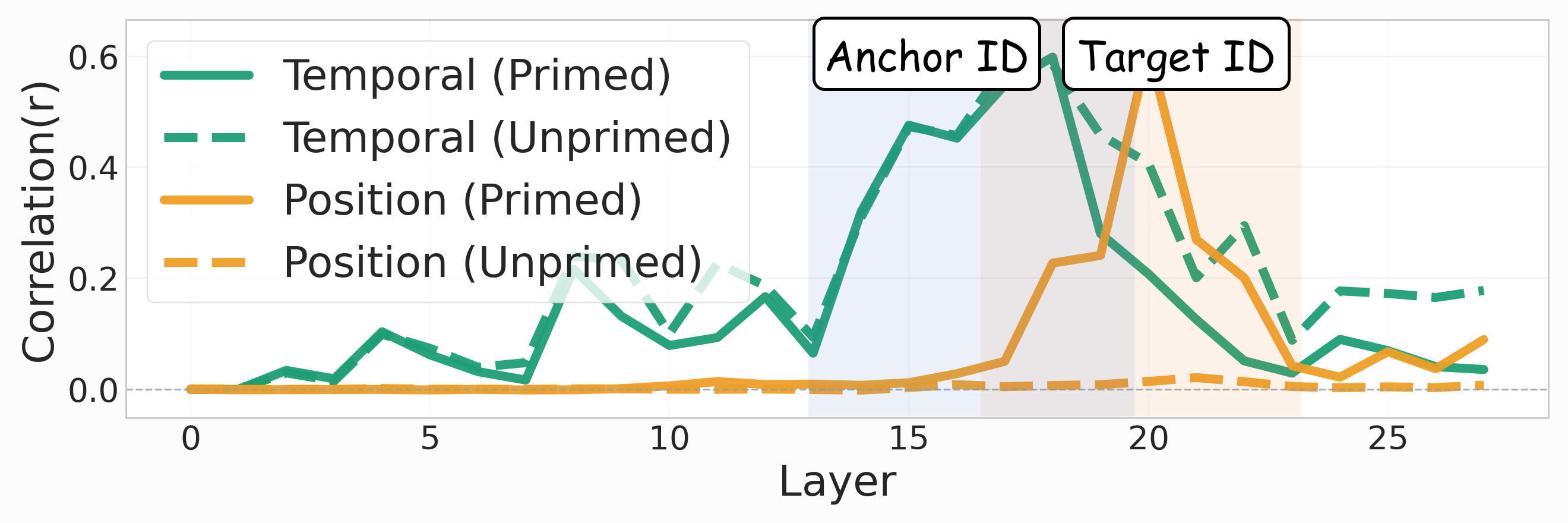}
        \caption{RSA results for primed and unprimed setting in AAVR task}
    \end{subfigure}
    \hfill
    \begin{subfigure}{0.48\linewidth}
        \centering
        \includegraphics[width=\linewidth]{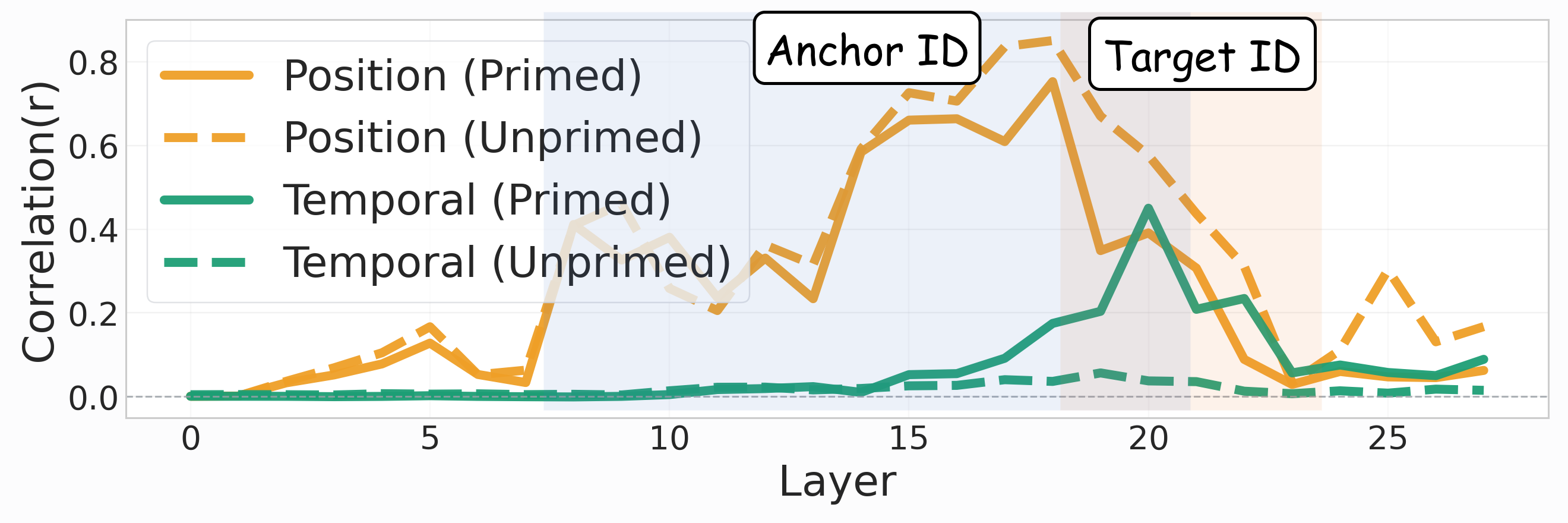}
        \caption{RSA results for primed and unprimed setting in VAAR task}
    \end{subfigure}
    \caption{RSA results primed vs unprimed settings for video-SALMONN2+(7B)}
    \label{app_fig:salmon_wo_hint}
\end{figure}

\begin{figure}[t]
    \centering
    \begin{subfigure}{0.5\linewidth}
        \centering
        \includegraphics[width=\linewidth]{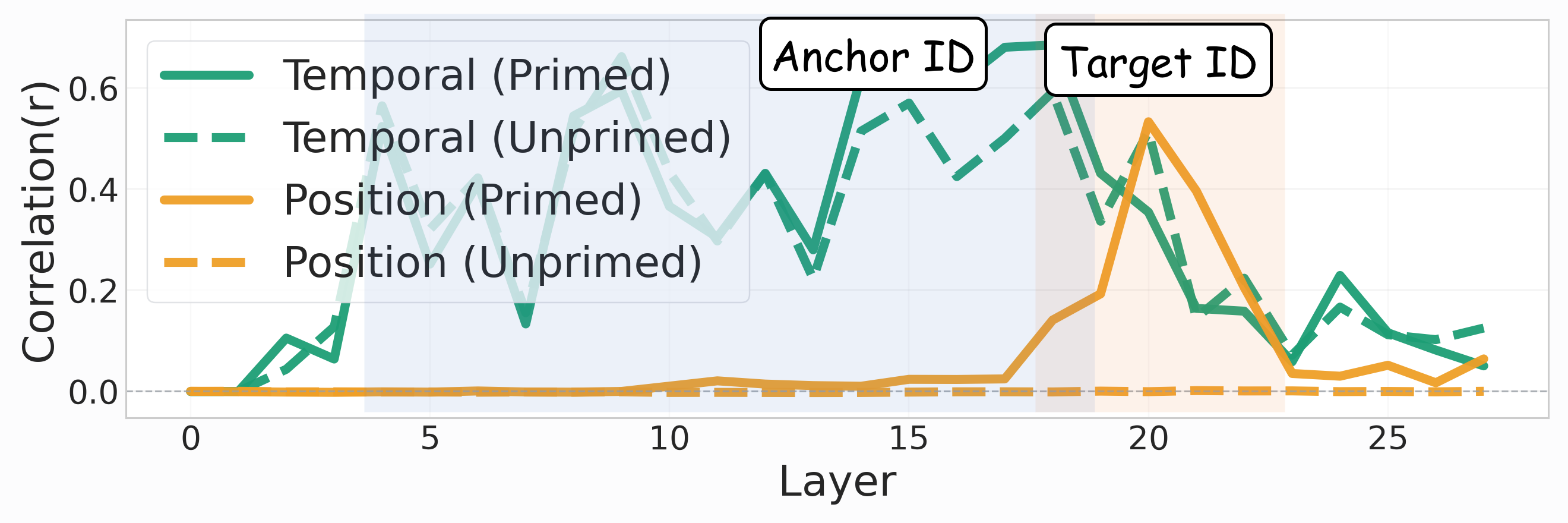}
        \caption{RSA results for primed and unprimed setting in AAVR task}
    \end{subfigure}
    \hfill
    \begin{subfigure}{0.48\linewidth}
        \centering
        \includegraphics[width=\linewidth]{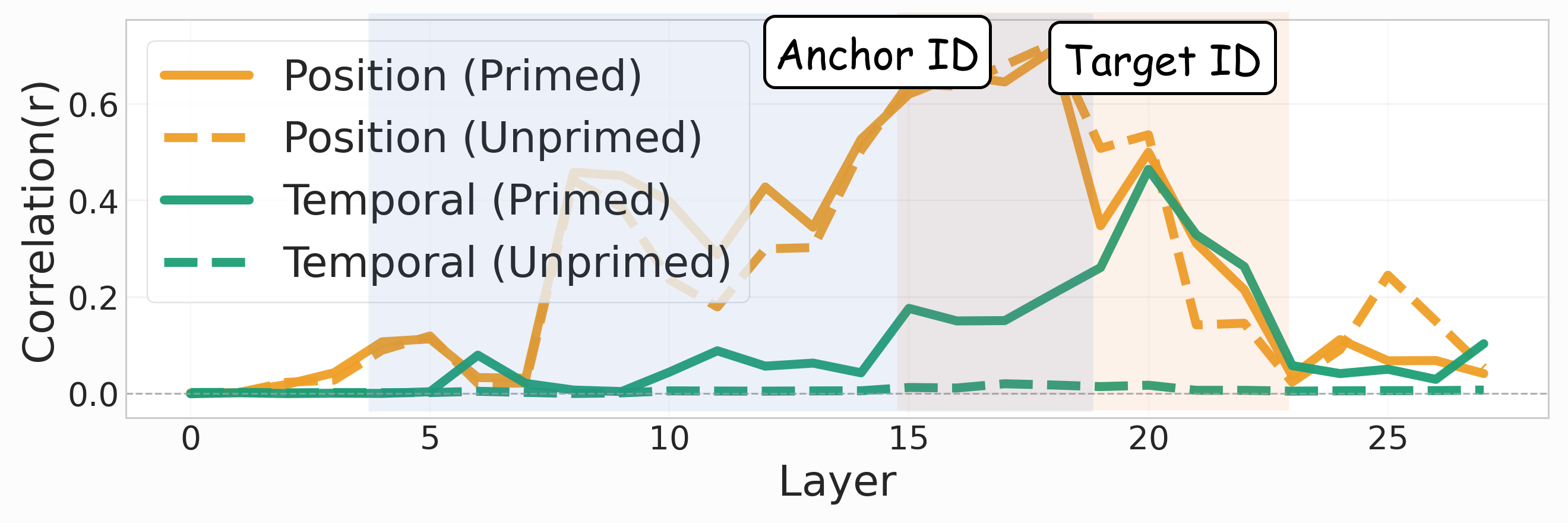}
        \caption{RSA results for primed and unprimed setting in VAAR task}
    \end{subfigure}
    \caption{RSA results primed vs unprimed settings for Qwen2.5-Omni(7B)}
    \label{app_fig:qwen7b_wo_hint}
\end{figure}

\begin{figure}[t]
    \centering
    \begin{subfigure}{0.5\linewidth}
        \centering
        \includegraphics[width=\linewidth]{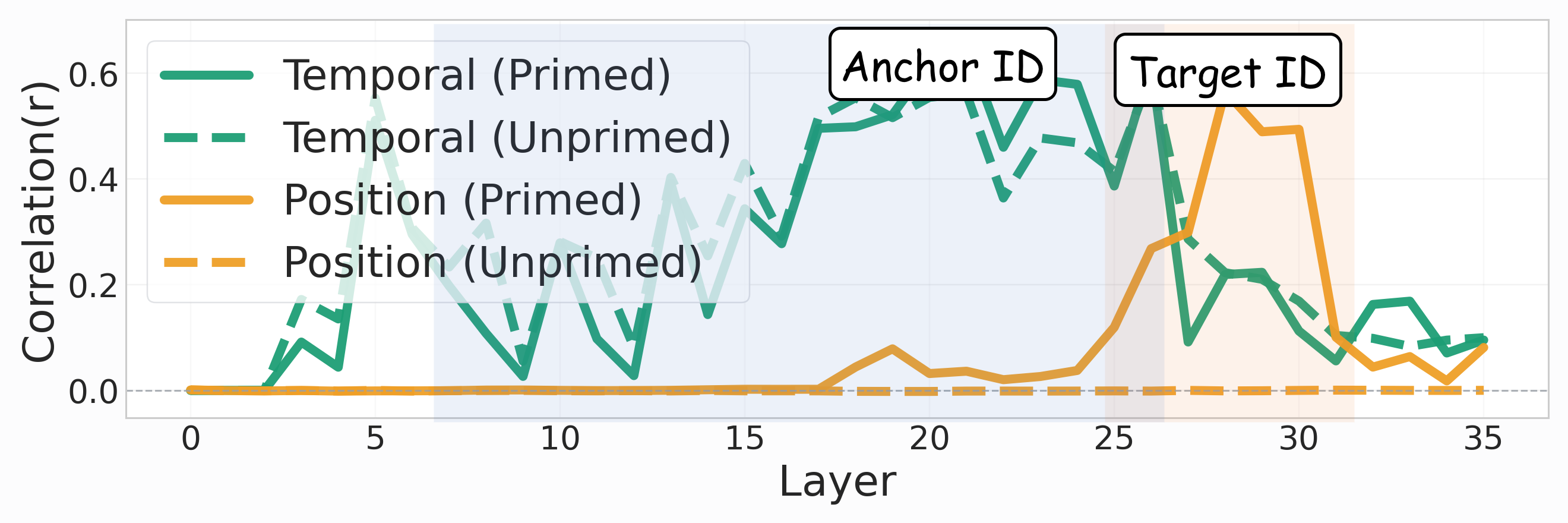}
        \caption{RSA results for primed and unprimed setting in AAVR task}
    \end{subfigure}
    \hfill
    \begin{subfigure}{0.48\linewidth}
        \centering
        \includegraphics[width=\linewidth]{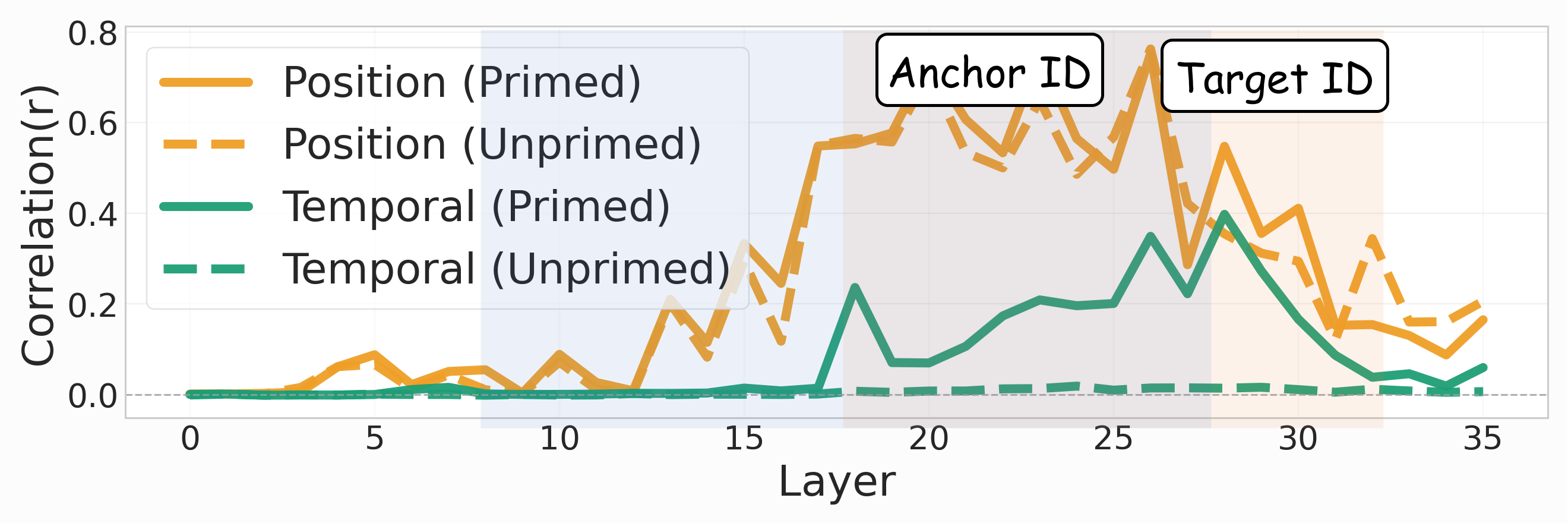}
        \caption{RSA results for primed and unprimed setting in VAAR task}
    \end{subfigure}
    \caption{RSA results primed vs unprimed settings for Qwen2.5-Omni(3B)}
    \label{app_fig:qwen3b_wo_hint}
\end{figure}

\begin{figure}[t]
    \centering
    \begin{subfigure}{0.5\linewidth}
        \centering
        \includegraphics[width=\linewidth]{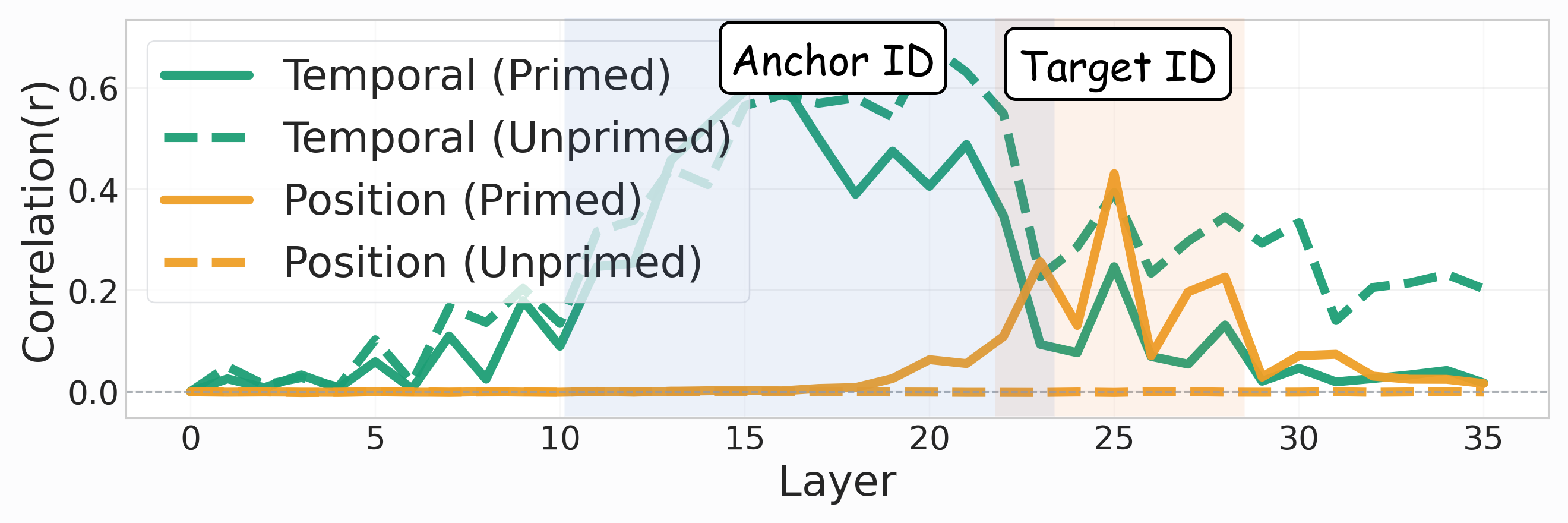}
        \caption{RSA results for primed and unprimed setting in AAVR task}
    \end{subfigure}
    \hfill
    \begin{subfigure}{0.48\linewidth}
        \centering
        \includegraphics[width=\linewidth]{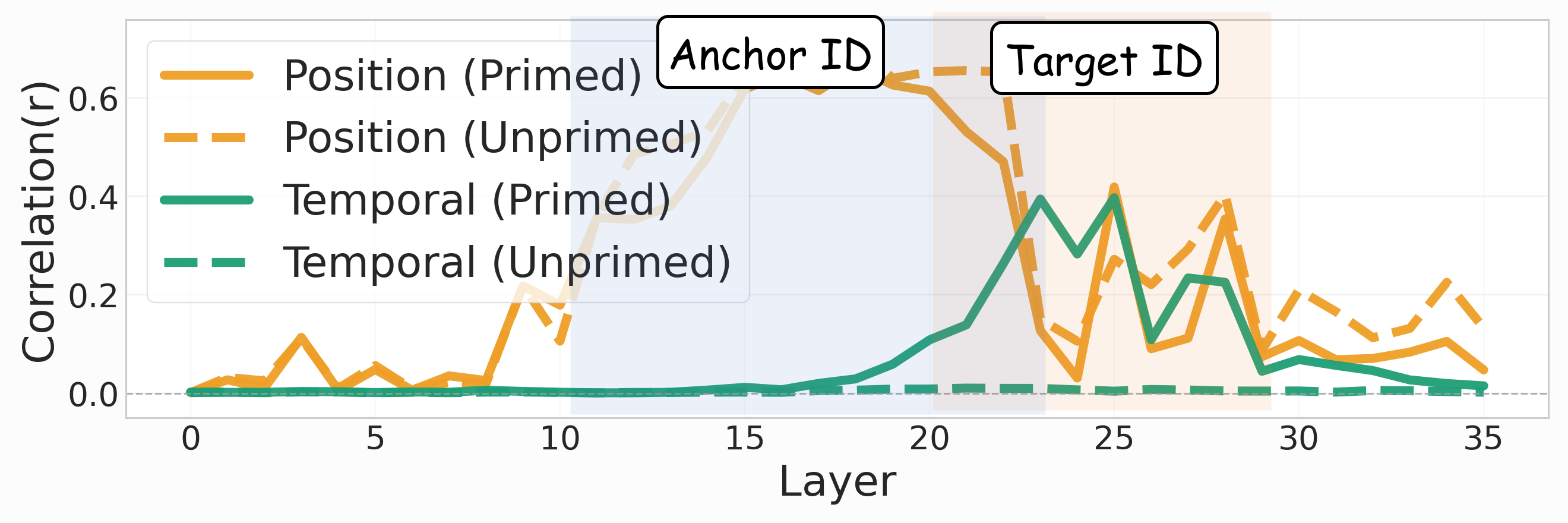}
        \caption{RSA results for primed and unprimed setting in VAAR task}
    \end{subfigure}
    \caption{RSA results primed vs unprimed settings for MiniCPM-o-4.5}
    \label{app_fig:minicpm_wo_hint}
\end{figure}

\cref{app_fig:salmon_wo_hint,app_fig:qwen7b_wo_hint,app_fig:qwen3b_wo_hint,app_fig:minicpm_wo_hint} present the representation analysis results comparing the unprimed setting and primed conditions for both the AAVR and VAAR tasks across four models: video-SALMONN2+, Qwen2.5-Omni(7B), Qwen2.5-Omni(3B), and MiniCPM-o-4.5. Across all cases, a common trend emerges: during the anchor ID retrieval stage, there is no observable difference between the unprimed and primed settings. This indicates that this initial stage functions correctly even in the absence of explicit priming. However, a significant divergence occurs during the subsequent target ID selection stage, where the unprimed condition clearly struggles to perform effectively compared to the primed condition.

\subsection{Real World Generalization Analysis on Different Models}
\label{app:real}
 We conduct RSA on approximately 2,000 SocialOmni~\cite{xie2026socialomni} two-speaker clips. We extract segments in which two speakers are positioned left and right and each produces one utterance, then annotate speaker position and utterance order to construct RSA-compatible datasets. Although this forms a $2\times1$ layout, speaker locations vary freely across clips rather than occupying fixed grid positions. As in Section~3.5, we report Position ID and Temporal ID correlations. 

For our study, we utilize a subset restricted to videos containing exactly two individuals. We enrich this data by extracting explicit visual attributes. Specifically, we sample a single frame from each video and prompt GPT-4o-mini to generate detailed descriptions of both speakers' appearances and attire (e.g., ``a man wearing a white shirt and a hat'').

Furthermore, we generate timestamped audio transcriptions using Whisper-large-v3~\cite{whisper}. By aligning these transcriptions with the original active speaker bounding box annotations, we accurately attribute each spoken utterance to either the ``left'' or ``right'' speaker. Finally, we formulate our AAVR and VAAR tasks using the first two utterances from each speaker.

\begin{figure}[t]
    \centering
    \begin{subfigure}[b]{0.47\textwidth}
        \centering
        \includegraphics[width=\textwidth]{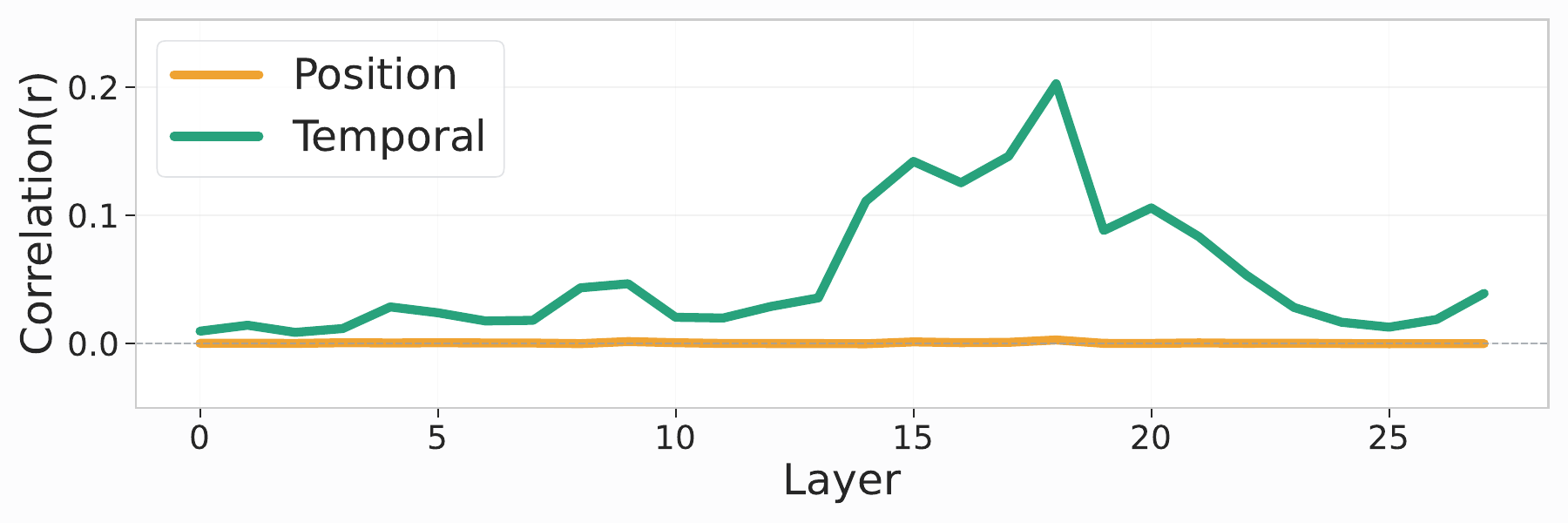} 
        \caption{RSA at anchor attribute token in AAVR task}
    \end{subfigure}
    \hfill 
    \begin{subfigure}[b]{0.47\textwidth}
        \centering
        \includegraphics[width=\textwidth]{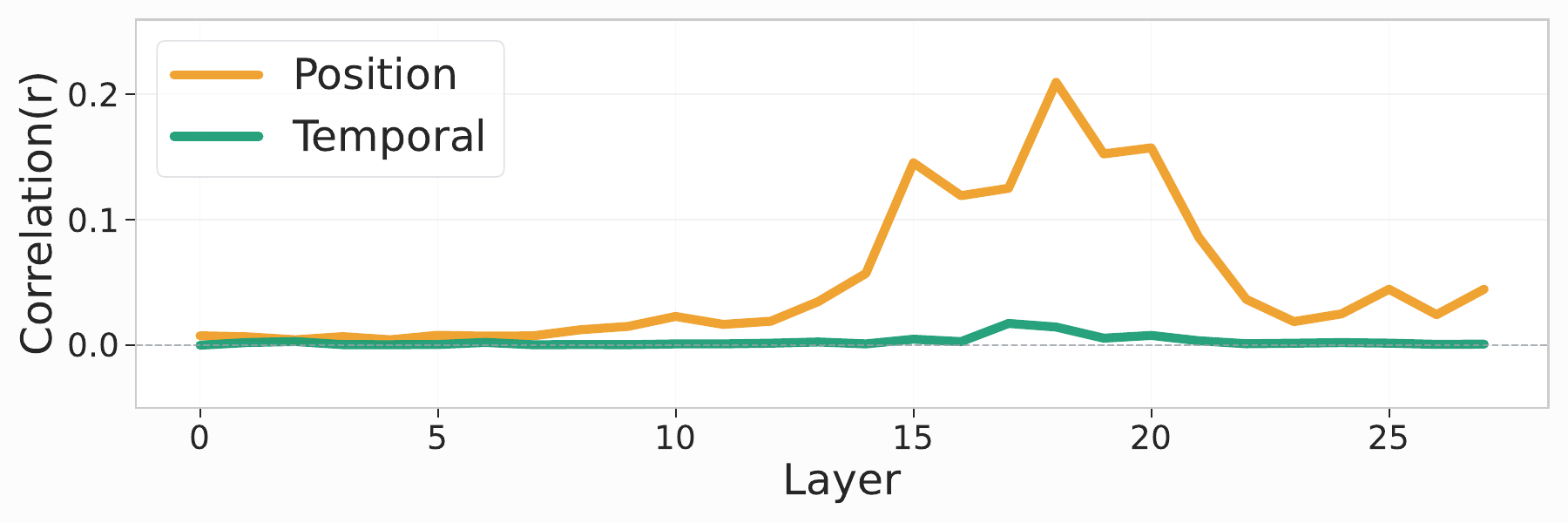} 
        \caption{RSA at anchor attribute token in VAAR task}
    \end{subfigure}
    \vspace{0.1cm} 
    \begin{subfigure}[b]{0.47\textwidth}
        \centering
        \includegraphics[width=\textwidth]{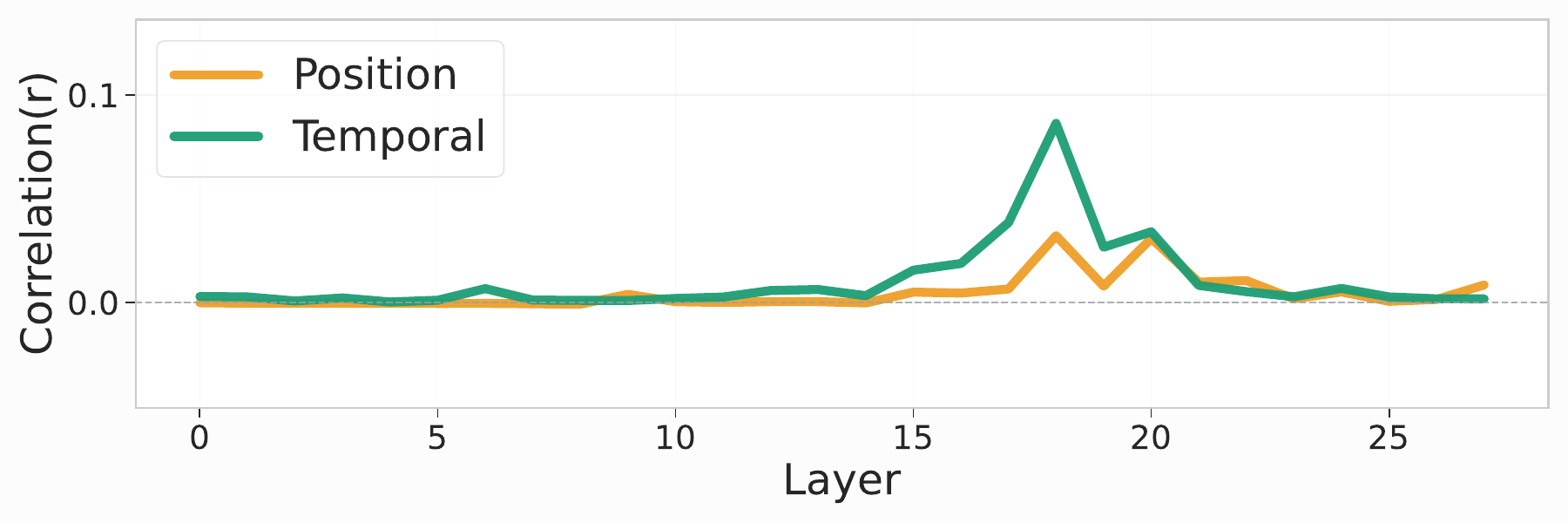} 
       \caption{RSA at last token in AAVR task}
    \end{subfigure}
    \hfill
    \begin{subfigure}[b]{0.47\textwidth}
        \centering
        \includegraphics[width=\textwidth]{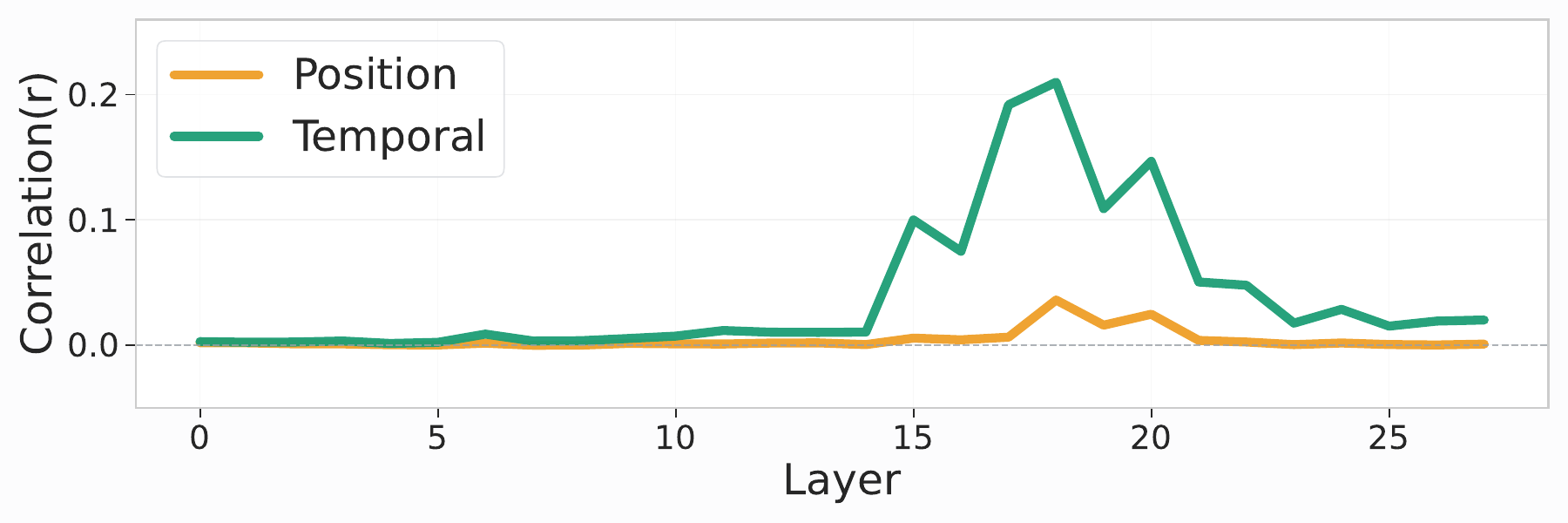} 
        \caption{RSA at last token in VAAR task}
    \end{subfigure}
    \caption{RSA results for video-SALMONN2+(7B) on real-world data.}
    \label{app_fig:plus_real}
\end{figure}

\begin{figure}[t]
    \centering
    \begin{subfigure}[b]{0.47\textwidth}
        \centering
        \includegraphics[width=\textwidth]{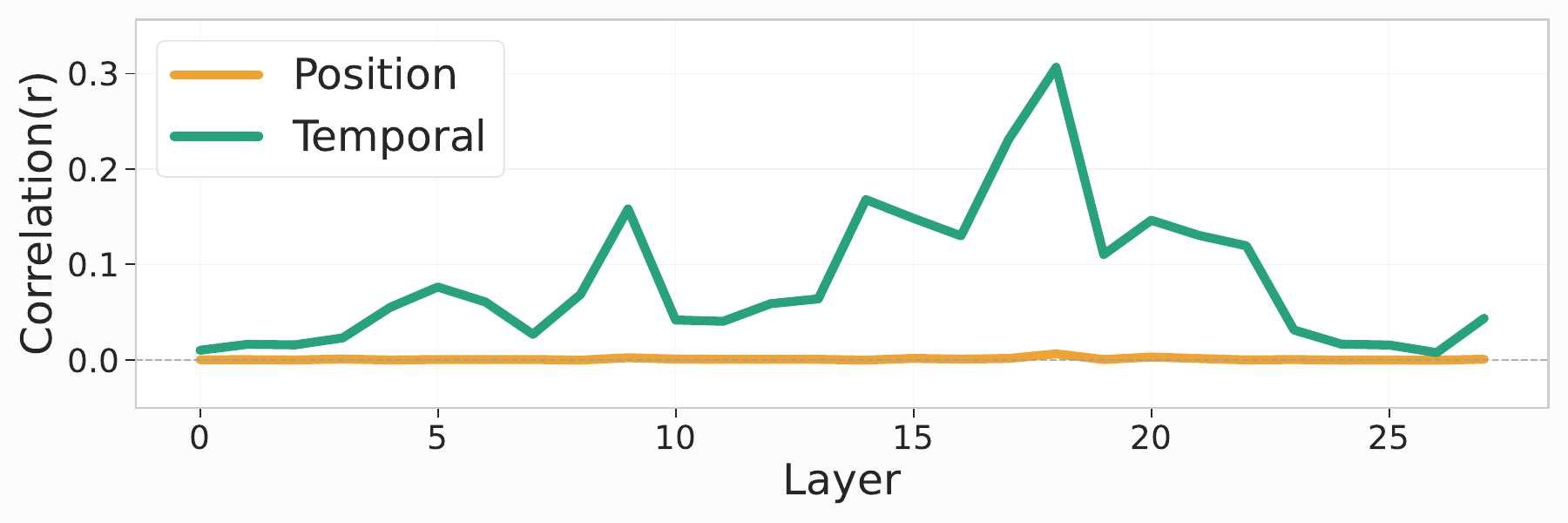} 
        \caption{RSA at anchor attribute token in AAVR task}
    \end{subfigure}
    \hfill 
    \begin{subfigure}[b]{0.47\textwidth}
        \centering
        \includegraphics[width=\textwidth]{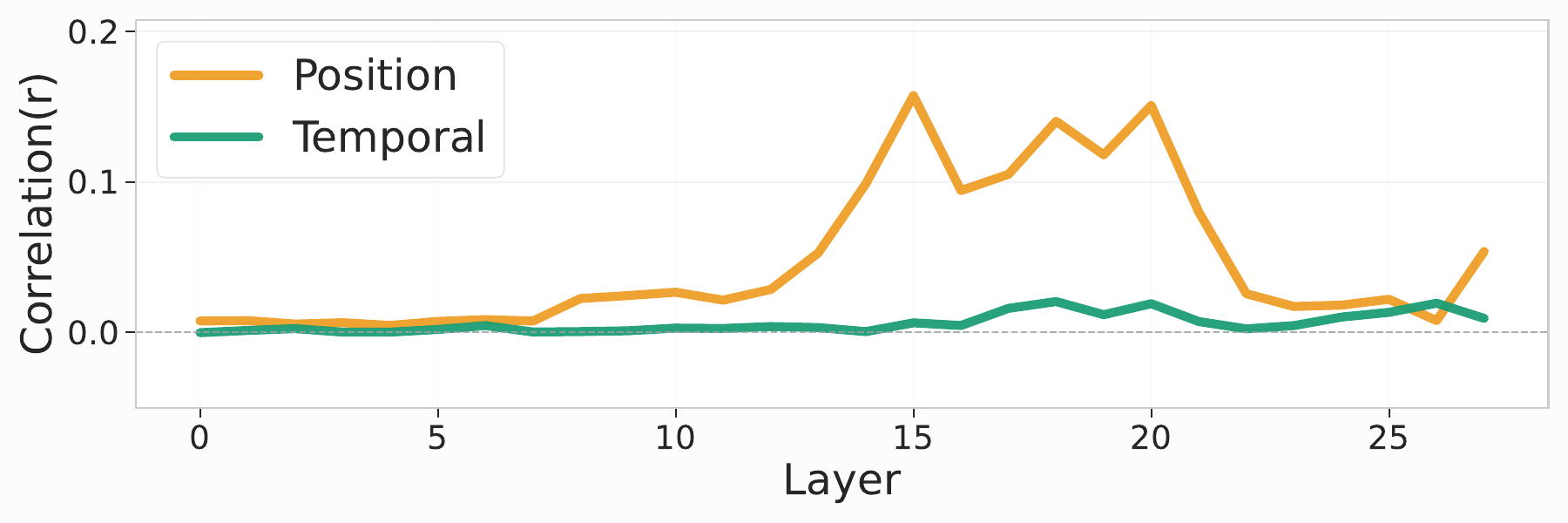} 
        \caption{RSA at anchor attribute token in VAAR task}
    \end{subfigure}
    \vspace{0.1cm} 
    \begin{subfigure}[b]{0.47\textwidth}
        \centering
        \includegraphics[width=\textwidth]{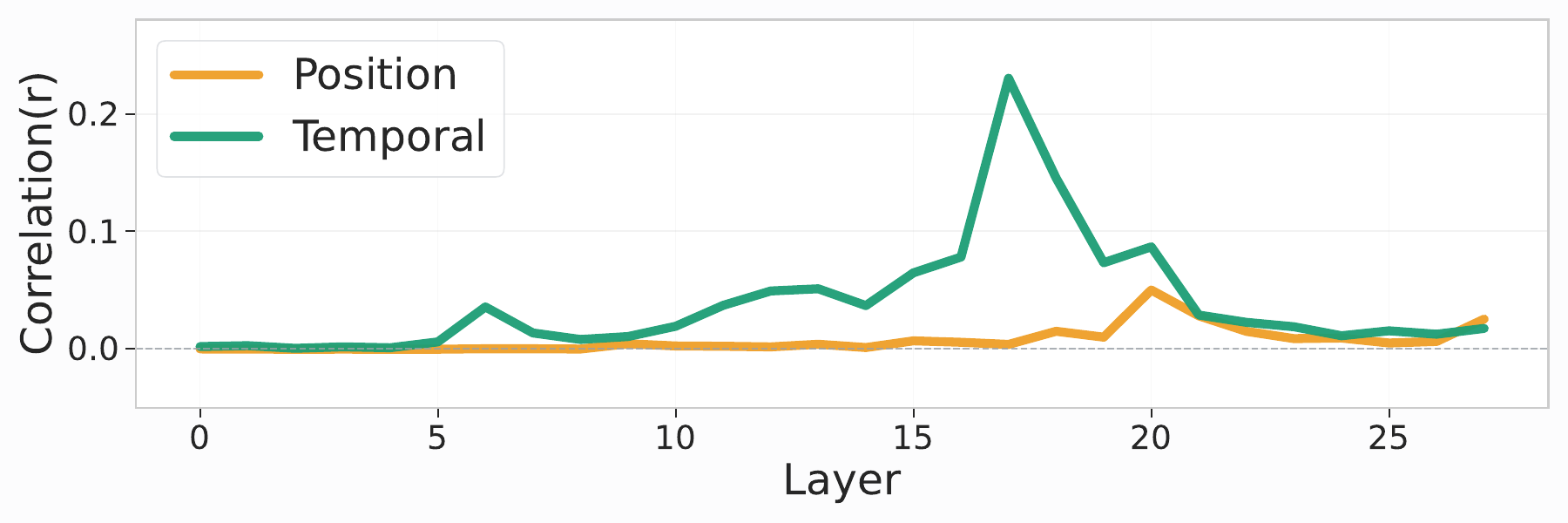} 
       \caption{RSA at last token in AAVR task}
    \end{subfigure}
    \hfill
    \begin{subfigure}[b]{0.47\textwidth}
        \centering
        \includegraphics[width=\textwidth]{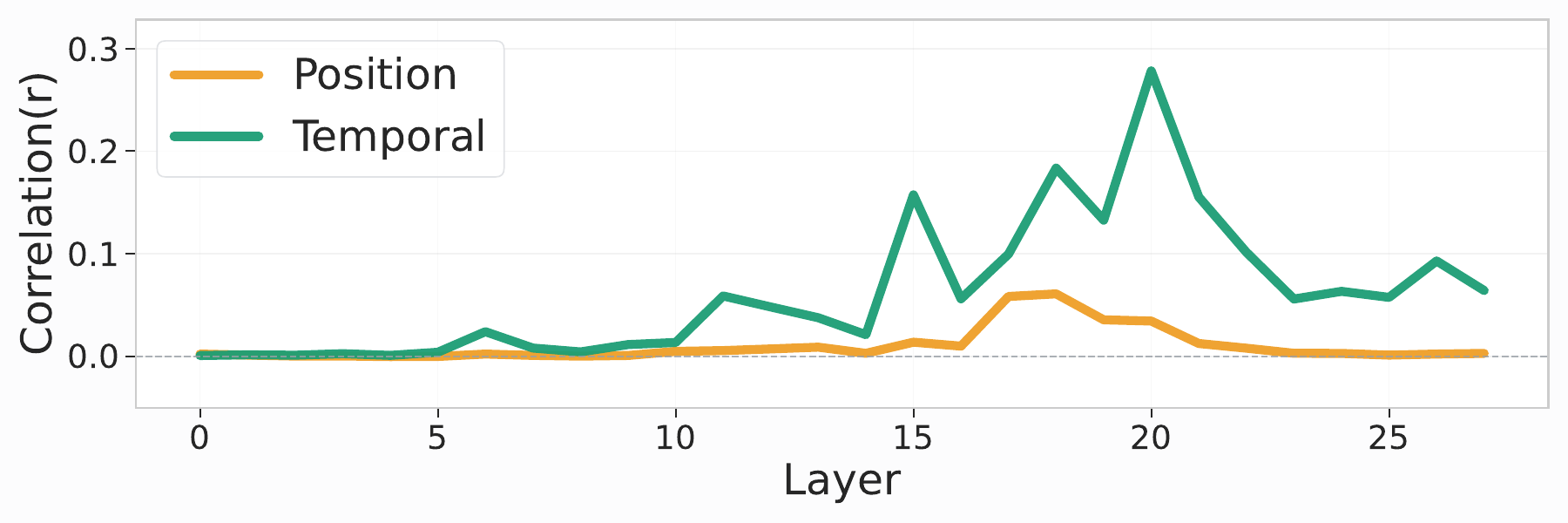} 
        \caption{RSA at last token in VAAR task}
    \end{subfigure}
    \caption{RSA results for Qwen2.5-Omni(7B) on real-world data.}
    \label{app_fig:qwen7b_real}
\end{figure}

\begin{figure}[t]
    \centering
    \begin{subfigure}[b]{0.47\textwidth}
        \centering
        \includegraphics[width=\textwidth]{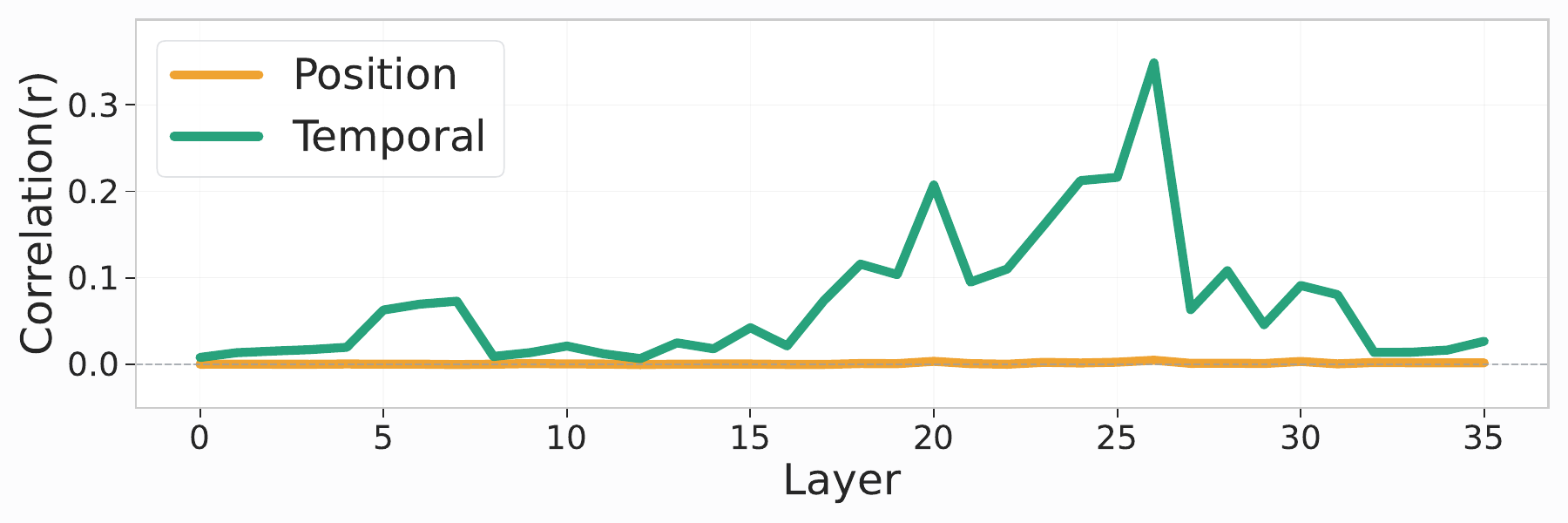} 
        \caption{RSA at anchor attribute token in AAVR task}
    \end{subfigure}
    \hfill 
    \begin{subfigure}[b]{0.47\textwidth}
        \centering
        \includegraphics[width=\textwidth]{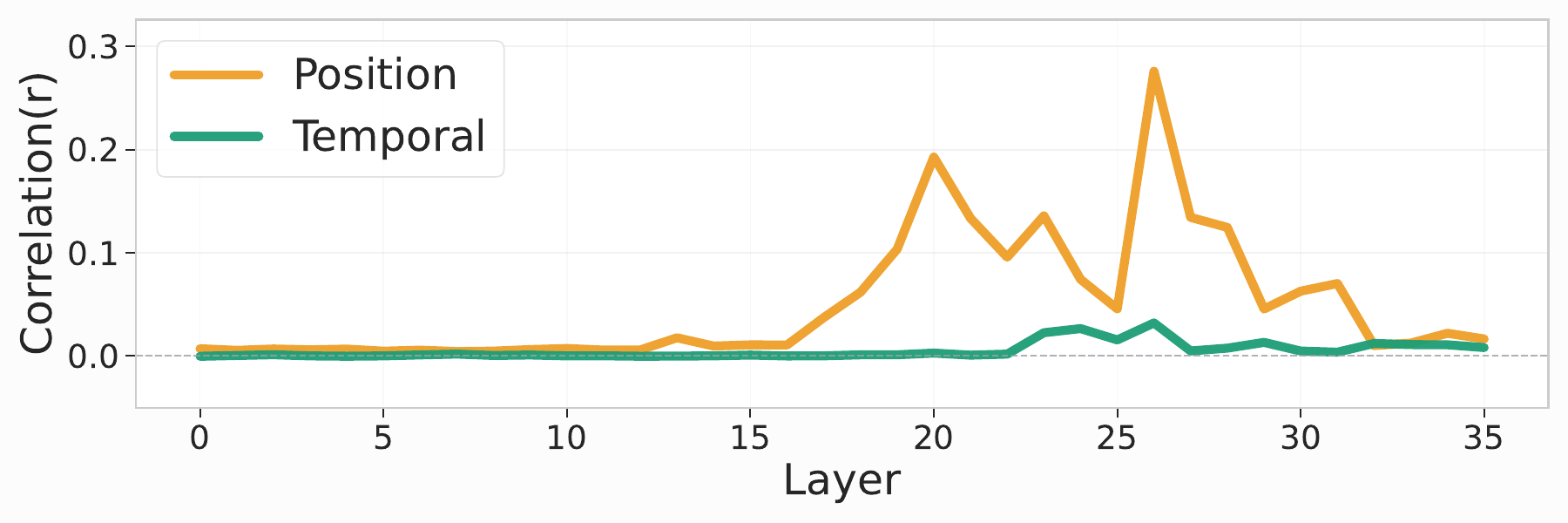} 
        \caption{RSA at anchor attribute token in VAAR task}
    \end{subfigure}
    \vspace{0.1cm} 
    \begin{subfigure}[b]{0.47\textwidth}
        \centering
        \includegraphics[width=\textwidth]{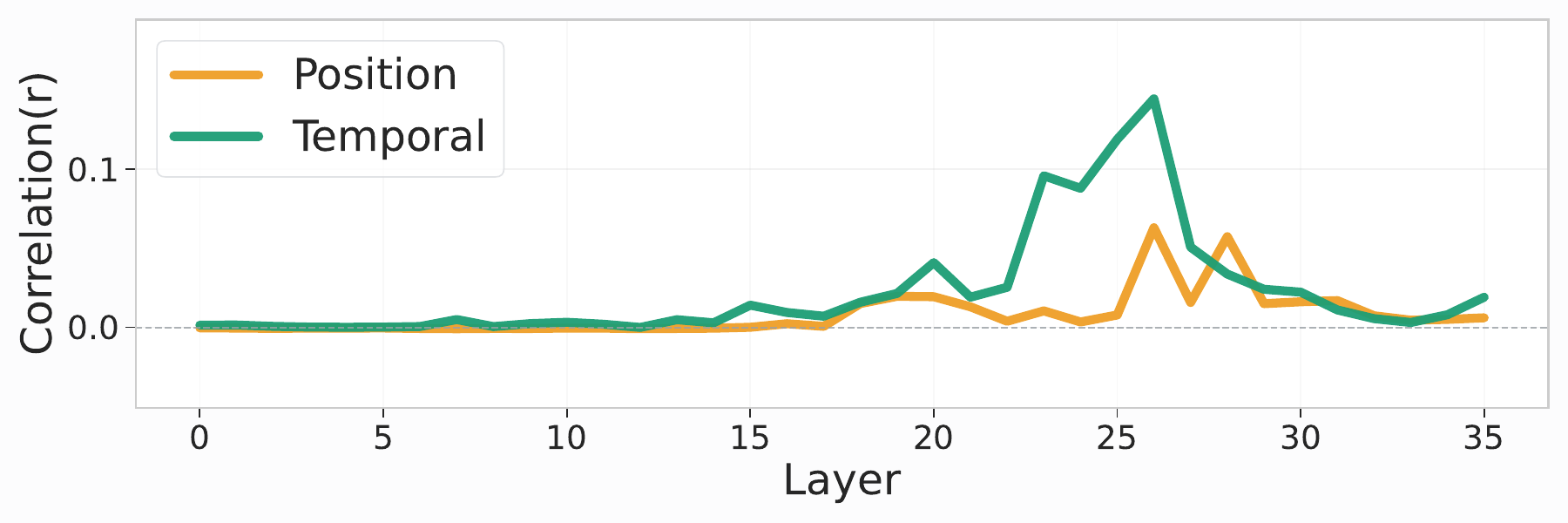} 
       \caption{RSA at last token in AAVR task}
    \end{subfigure}
    \hfill
    \begin{subfigure}[b]{0.47\textwidth}
        \centering
        \includegraphics[width=\textwidth]{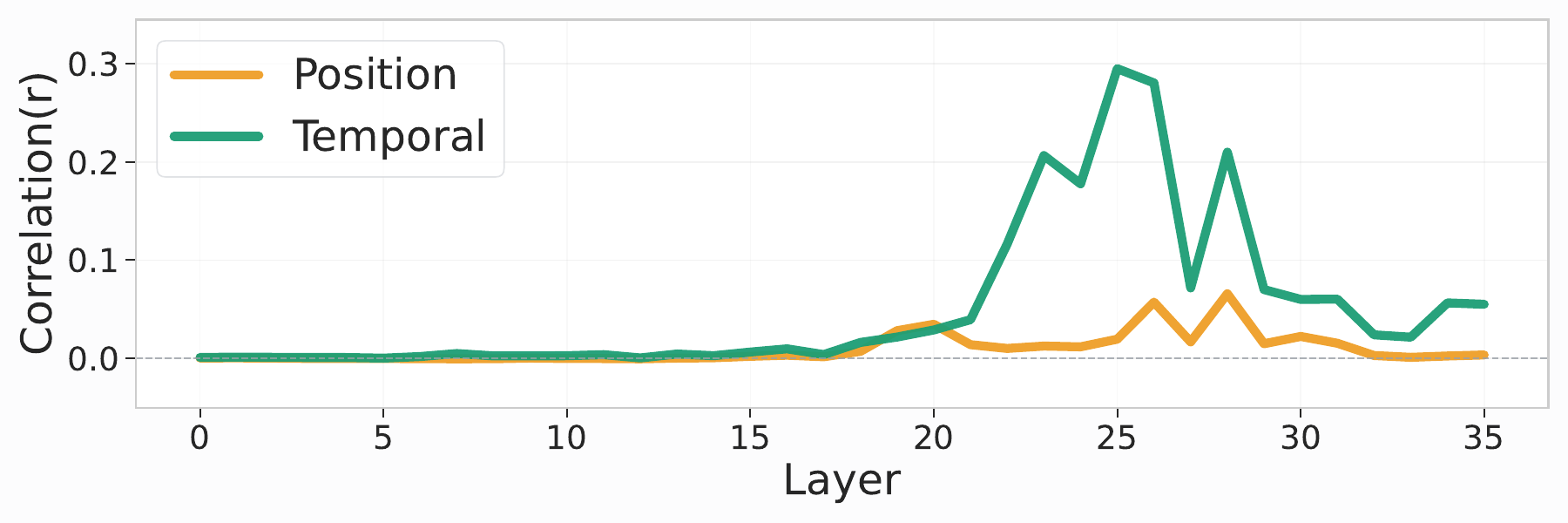} 
        \caption{RSA at last token in VAAR task}
    \end{subfigure}
    \caption{RSA results for Qwen2.5-Omni(3B) on real-world data.}
    \label{app_fig:qwen3b_real}
\end{figure}

\begin{figure}[t]
    \centering
    \begin{subfigure}[b]{0.47\textwidth}
        \centering
        \includegraphics[width=\textwidth]{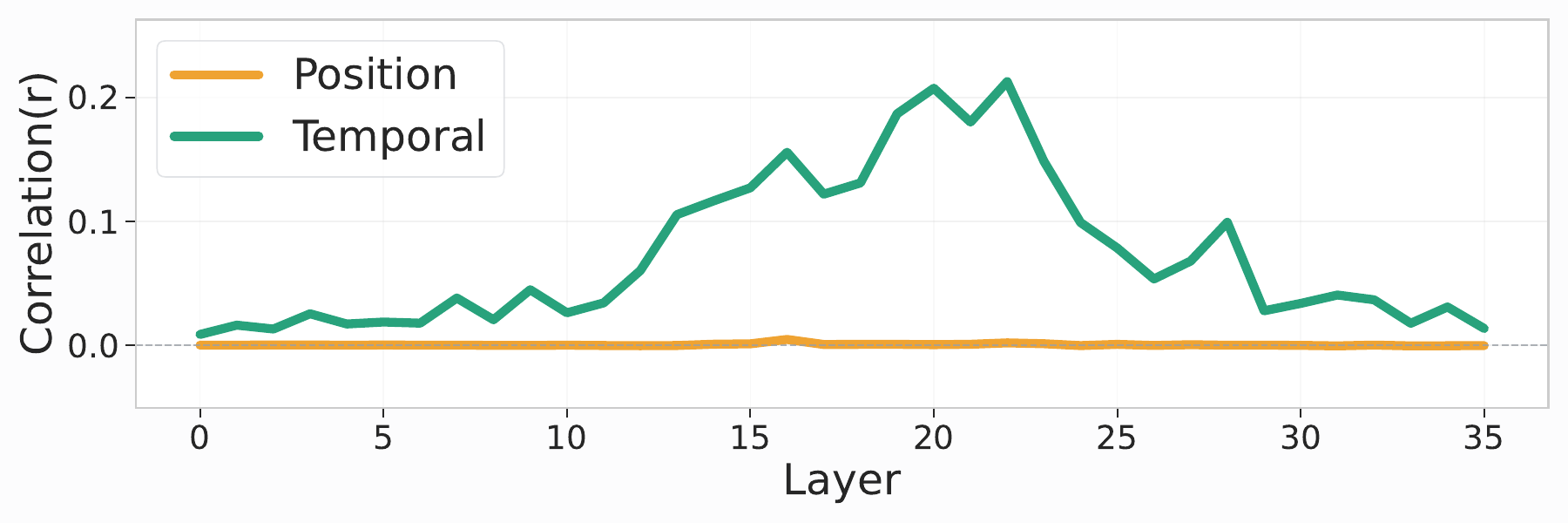}  
        \caption{RSA at anchor attribute token in AAVR task}
    \end{subfigure}
    \hfill 
    \begin{subfigure}[b]{0.47\textwidth}
        \centering
        \caption{RSA at anchor attribute token in VAAR task}
    \end{subfigure}
    \vspace{0.1cm} 
    \begin{subfigure}[b]{0.47\textwidth}
        \centering
        \includegraphics[width=\textwidth]{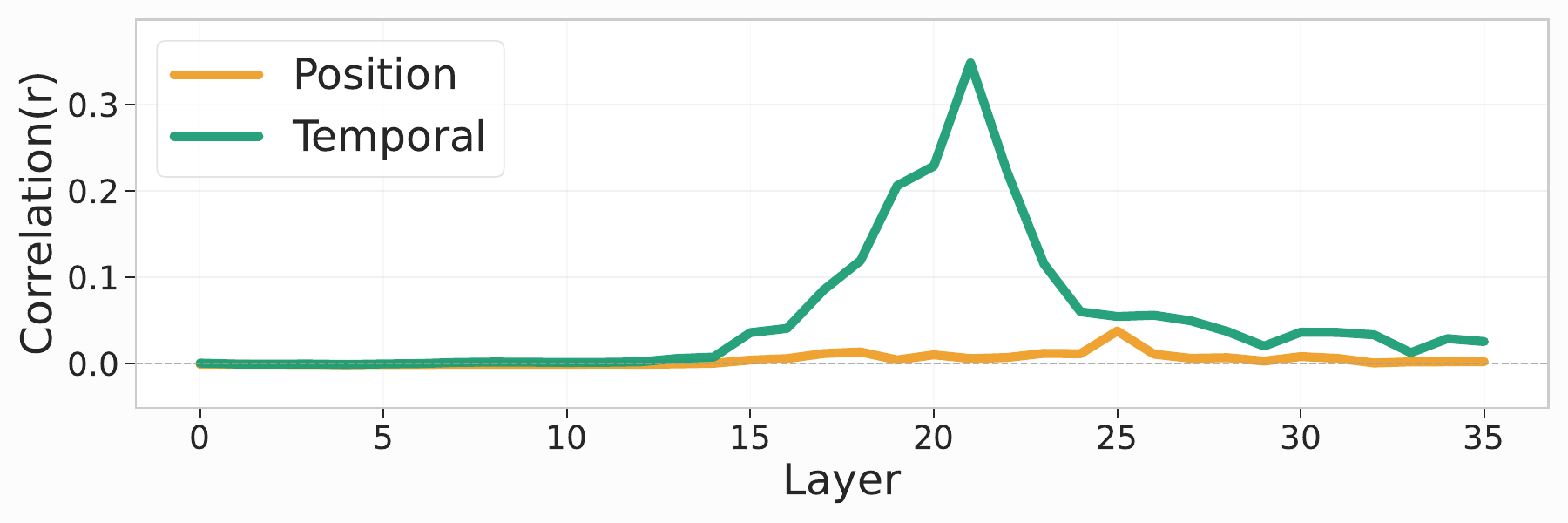}  
       \caption{RSA at last token in AAVR task}
    \end{subfigure}
    \hfill
    \begin{subfigure}[b]{0.47\textwidth}
        \centering
        \includegraphics[width=\textwidth]{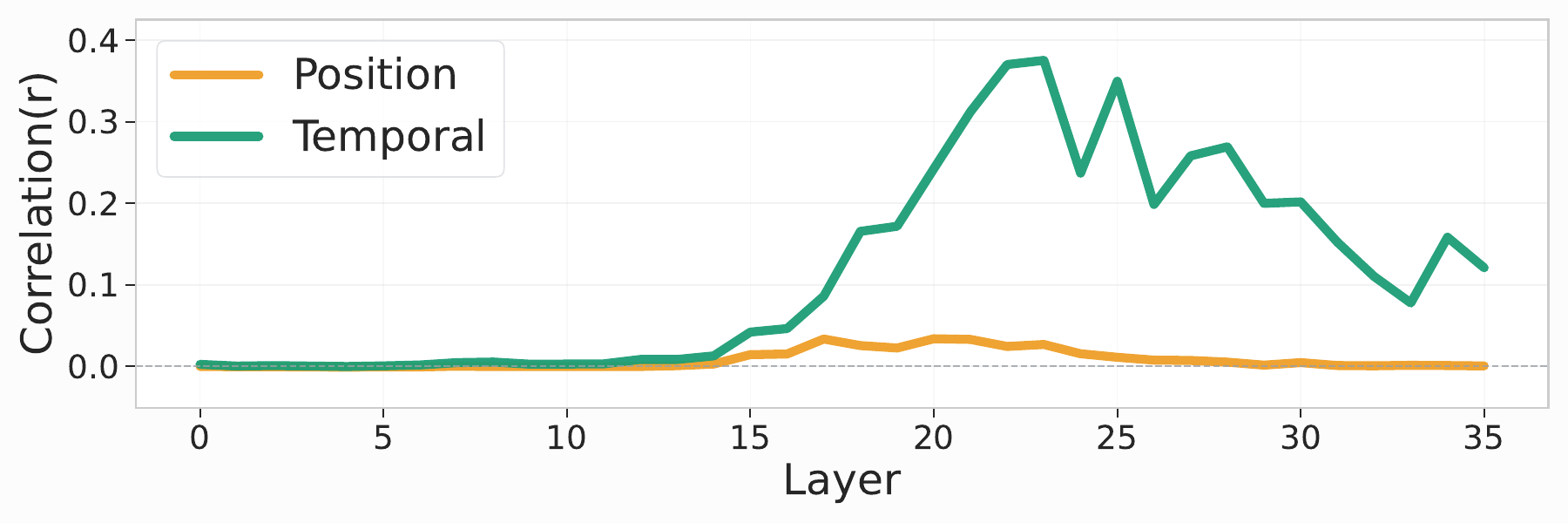}  
        \caption{RSA at last token in VAAR task}
    \end{subfigure}
    \caption{RSA results for MiniCPM-o-4.5(9B) on real-world data.}
    \label{app_fig:minicpm_real}
\end{figure}

\cref{app_fig:plus_real,app_fig:qwen7b_real,app_fig:qwen3b_real,app_fig:minicpm_real} present the results of replicating the real-world dataset experiments (\cref{sec:real}) on video-SALMONN2+(7B), Qwen2.5-Omni (7B, 3B), and MiniCPM-o-4.5, respectively. Overall, while the patterns are less pronounced than those seen in the toy dataset, the observed trends remain consistent with our proposed mechanism. In the AAVR case, the temporal ID exhibits a strong correlation at the anchor attribute token within the mid-to-late layers. Subsequently, at the last token—which is expected to encode the target entity from the opposite modality—the correlation with position IDs become stronger than at the anchor token, whereas the correlation with temporal IDs weakens. Conversely, in the VAAR case, while the temporal ID correlation at the anchor attribute token appears somewhat stronger in the early layers compared to the toy dataset, the position ID correlation remains distinctly dominant in the mid-to-late layers. Furthermore, at the last token, the temporal ID correlation becomes even stronger than at the anchor attribute token, whereas the position ID correlation noticeably weakens.


\section{Audio-Visual Prompting}
\label{app:audio_visual_prompting}
\subsection{Details of Training Free Prompting}
\label{app:training_Free_prompting}
To leverage the zero-shot reasoning capabilities without additional parameter updates, we employ a training-free prompting strategy that explicitly guides the model's spatial and temporal attention toward the active speaker by incorporating visual cues directly into the text prompt. We modify the original prompts according to the specific task: for Question Answering (QA) tasks, where precise grounding is crucial, we prepend the instruction, \textit{"At each moment, focus on the person with the red bounding box, as the red box marks the person that is currently speaking. [Original Prompt]"}. For Captioning tasks, we ensure narrative consistency by appending the note, \textit{"[Original Prompt] Note that the person with the red bounding box at each moment marks the person that is currently speaking."} By defining the red bounding box as a active speaker, this prompting mechanism minimizes identity ambiguity in multi-speaker scenarios and effectively aligns the model's linguistic output with the target speaker's actions and speech.

\newpara{Sensitivity to ASD}
\cref{tab:main} already uses imperfect annotations from a SOTA ASD model~\cite{nguyen2026laser} Even with two weaker ASD models~\cite{wang2024loconet, 10448124}. ASD-FT reaches 48.4\%/48.6\% on AVSpeaker---close to the 49.1\% with~\cite{nguyen2026laser} and well above vanilla 44.7\%---confirming robustness.

\subsection{Details of Finetuning}
\label{app:details_of_ft}
\subsubsection{Training Datasets}
\label{app:training_data}

\begin{figure}[t]
    \centering
    \begin{subfigure}{0.5\linewidth}
        \centering
        \includegraphics[width=\linewidth]{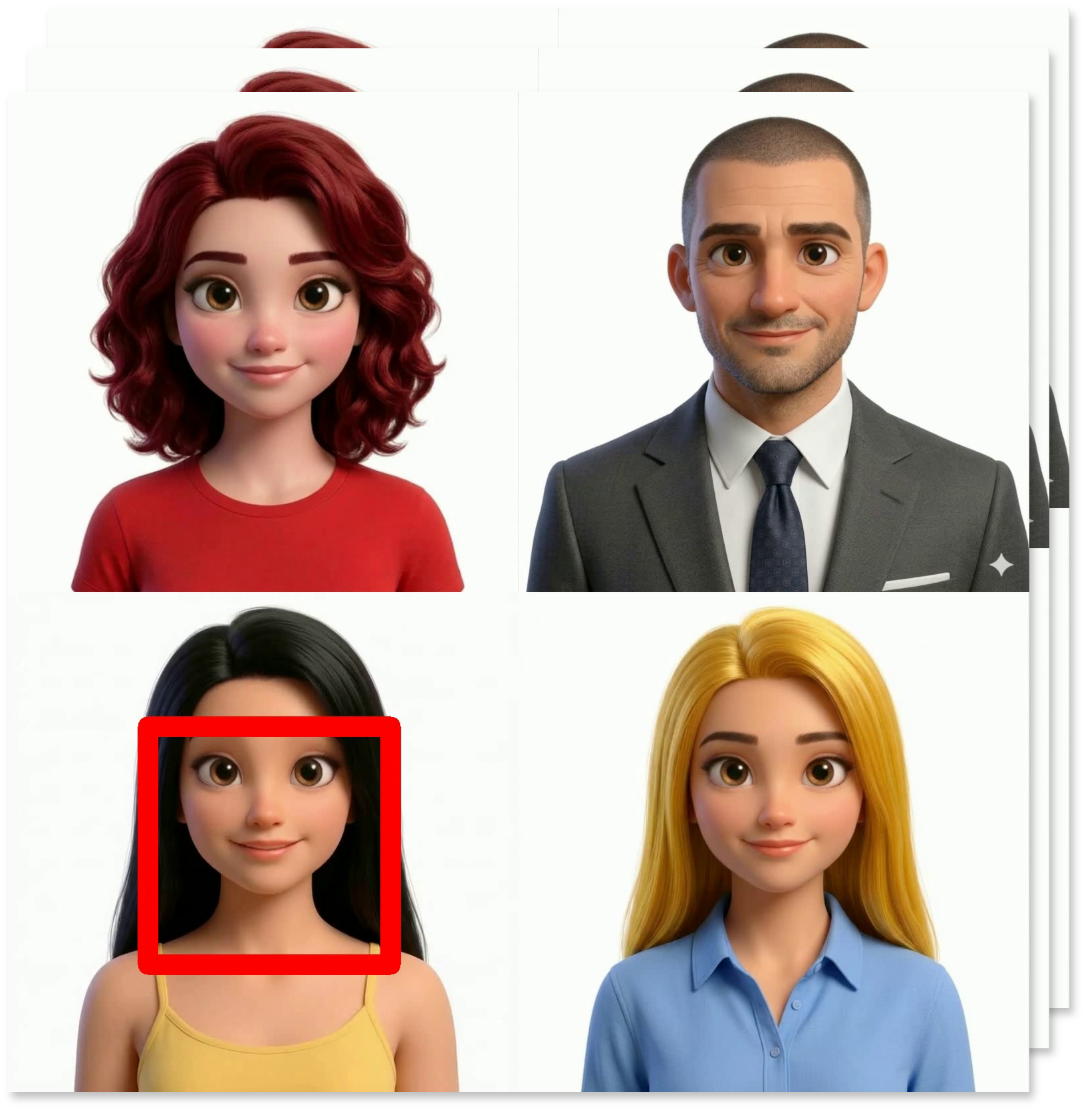}
        \caption{Four people talking single-shot video}
        \label{fig:four_single}
    \end{subfigure}
    \hfill
    \begin{subfigure}{0.48\linewidth}
        \centering
        \includegraphics[width=\linewidth]{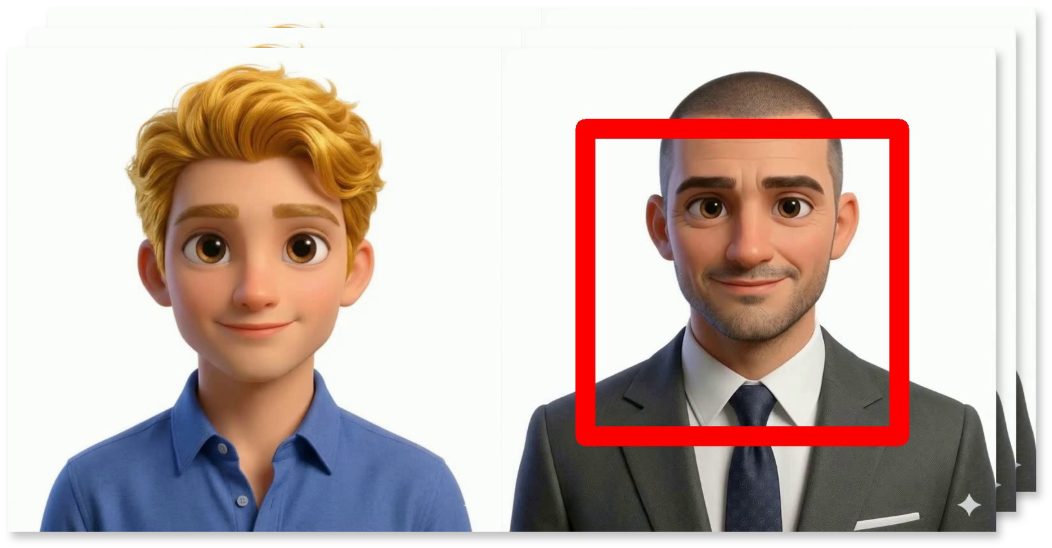}
        \caption{Two people talking single-shot video}
        \label{fig:two_single}
    \end{subfigure}
    \begin{subfigure}{0.48\linewidth}
        \centering
        \includegraphics[width=\linewidth]{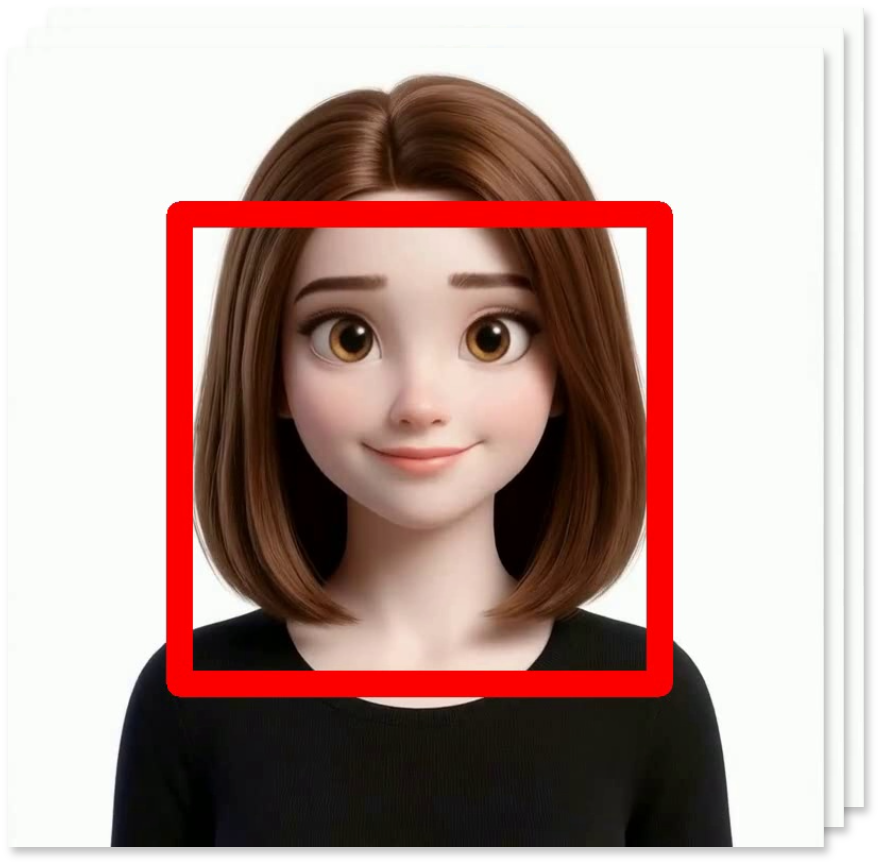}
        \caption{Two people talking multi-shot video}
        \label{fig:two_multi}
    \end{subfigure}
    \caption{Examples of the three video configurations in our training corpus. The active speaker at each moment is marked with a red bounding box.}
    \label{fig:training_dataset}
\end{figure}

We construct the training corpus from a synthetic, human-character video dataset in which one or more speakers each utter a single country name (\textit{Canada}, \textit{Germany}, \textit{Japan}, or \textit{Egypt}). The active speaker at every moment is highlighted by a red bounding box, providing an unambiguous visual--audio--text alignment supervision signal. As illustrated in \cref{fig:training_dataset}, the corpus contains three video configurations that vary along two axes: the number of people on screen and whether all speakers appear together in a single shot or are presented as a sequence of shots (multi-shot).

\paragraph{Asset pool.}
All clips are composed from a small, fixed asset pool, which makes per-axis ablation analyses tractable while still producing diverse compositions. The pool consists of (i) eight human \emph{characters} -- four female and four male -- whose reference images are generated with Gemini and used consistently throughout the dataset; (ii) four \emph{country names} that constitute the entire spoken vocabulary, namely \textit{Canada}, \textit{Germany}, \textit{Japan}, and \textit{Egypt}; and (iii) a set of TTS-generated \emph{voices} used to synthesize the spoken country names. Each video is built by sampling a layout of characters, a per-character country word, and a per-character voice from these pools, so that visual identity, lexical content, and acoustic identity are independently controlled.

\paragraph{Video configurations.}
\cref{tab:train_videos} summarizes the three configurations. The two-people single-shot setting (\cref{fig:two_single}) shows two speakers simultaneously in one continuous shot, with one word spoken per person. The four-people single-shot setting (\cref{fig:four_single}) extends this to four simultaneously visible speakers. The two-people multi-shot setting (\cref{fig:two_multi}) again involves two speakers, but they appear in separate consecutive shots rather than together, requiring the model to track identity across shot boundaries.

\begin{table}[h]
\centering
\small
\caption{Video configurations in the training set. Each unique video yields multiple training samples by pairing it with different question types and target utterances.}
\label{tab:train_videos}
\begin{tabular}{ccr}
\toprule
\# People & Shot type & \# Videos \\
\midrule
2 & Single-shot & 200 \\
4 & Single-shot & 100 \\
2 & Multi-shot  & 100 \\
\midrule
\multicolumn{2}{c}{Total} & 400 \\
\bottomrule
\end{tabular}
\end{table}

\paragraph{Question types.}
Each video is paired with three kinds of questions, yielding 2{,}040 training samples in total. AAVR, VAAR, \textbf{Captioning}(It requires a free-form summary that names every speaker by appearance and reports the word each one utters). \cref{tab:train_questions} reports per-setting counts.

\begin{table}[h]
\centering
\small
\caption{Distribution of training samples by video setting and question type. AAVR and VAAR are 2-way multiple choice for the two-people settings and 4-way multiple choice for the four-people setting; Captioning is open-ended.}
\label{tab:train_questions}
\begin{tabular}{lrrrr}
\toprule
Setting & AAVR & VAAR & Captioning & Total \\
\midrule
Two people, single-shot  & 400 & 400 & 13 & 813 \\
Four people, single-shot & 400 & 400 & 13 & 813 \\
Two people, multi-shot   & 200 & 200 & 14 & 414 \\
\midrule
Total                    & 1{,}000 & 1{,}000 & 40 & 2{,}040 \\
\bottomrule
\end{tabular}
\end{table}

\paragraph{Question examples.}
For all multiple-choice questions, the prompt is preceded by a fixed instruction explaining that the red bounding box marks the current speaker, and the model is asked to respond with a single option letter. Representative prompts and gold answers from each question type are shown below.

\begin{tcolorbox}[title={AAVR (4-way)}, colback=gray!5, colframe=black!50, fonttitle=\bfseries\small, fontupper=\small]
\textbf{Prompt:} Which person speaks the 2nd utterance in the clip (the word ``Egypt'')? Note that the person with the red bounding box at each moment marks the person that is currently speaking. \\
A. A man with black hair wearing a gray T-shirt \\
B. A long-haired blond woman dressed in blue \\
C. A man dressed formally in a suit with short hair \\
D. A brown-haired woman in a black shirt \\
\textbf{Answer:} B
\end{tcolorbox}

\begin{tcolorbox}[title={VAAR (2-way)}, colback=gray!5, colframe=black!50, fonttitle=\bfseries\small, fontupper=\small]
\textbf{Prompt:} What single word does a woman with long blond hair in a blue shirt speak? Note that the person with the red bounding box at each moment marks the person that is currently speaking. \\
A. Egypt \\
B. Japan \\
\textbf{Answer:} A
\end{tcolorbox}

\begin{tcolorbox}[title={Captioning (open-ended)}, colback=gray!5, colframe=black!50, fonttitle=\bfseries\small, fontupper=\small]
\textbf{Prompt:} Provide a caption for this video, including the people present and the words they speak. Note that the person with the red bounding box at each moment marks the person that is currently speaking. \\
\textbf{Answer:} A sequence of four speakers is shown, each contributing one word. It starts with a red-haired woman in a red shirt saying ``Japan''. A blond-haired man in a blue shirt then says ``Egypt''. Then, a woman with long yellow hair wearing a blue shirt continues with ``Germany''. Finally, a black-haired man in a gray shirt says ``Canada''.
\end{tcolorbox}

\subsubsection{Implementation Details}
\label{app:training_detail}
Due to its rapid convergence, video-SALMONN2+ (7B) was trained for 140 steps, whereas the remaining models were trained for 250 steps. Furthermore, because our training data is synthetic, we bypassed off-the-shelf active speaker detection models; instead, we directly generated the exact bounding boxes during the data synthesis process. Evaluation is performed with greedy decoding. The $\pm$ values reported in \cref{tab:main} reflect variance across multiple Gemini-judge evaluations of the same model outputs.

\subsection{Discussion on the Experimental Results}
\label{app:diss}

\newpara{Training-Free ASD Results} The effectiveness of training-free ASD relies heavily on an AVLLM's zero-shot visual grounding capability to interpret overlaid markers-such as red bounding boxes-as indicators of the active speaker. Qwen2.5-Omni successfully leverages these overlays, supported by its explicit reports on referring expression comprehension and grounding benchmarks~\cite{xu2025qwen25omnitechnicalreport}. Similarly, MiniCPM-o-4.5 effectively utilizes this approach, as its training incorporates a relevance-aware masking strategy specifically designed to encourage the model to focus on visually grounded content~\cite{MiniCPM}. Conversely, video-SALMONN2+ does not benefit from training-free ASD. Although built upon Qwen2.5-VL~\cite{Bai2025Qwen25VLTR}, its downstream optimization is heavily tailored toward holistic audio-visual comprehension, leaving the zero-shot interpretation of fine-grained visual markers largely unaddressed. This indicates that without specific provisions to maintain or elicit these grounding skills during adaptation, training-free ASD becomes ineffective. When this prerequisite is unmet, our lightweight fine-tuning (ASD-FT) successfully bridges the gap, effectively teaching the model to bind the visual marker to the active speaker.

\section{Illustrations}
\label{app:illu}
\subsection{Illustration of Symbolic Mechanism in VAAR task}
\label{app:vaar_illu}
\begin{figure}[t]
    \centering
    \includegraphics[width=0.5\textwidth]{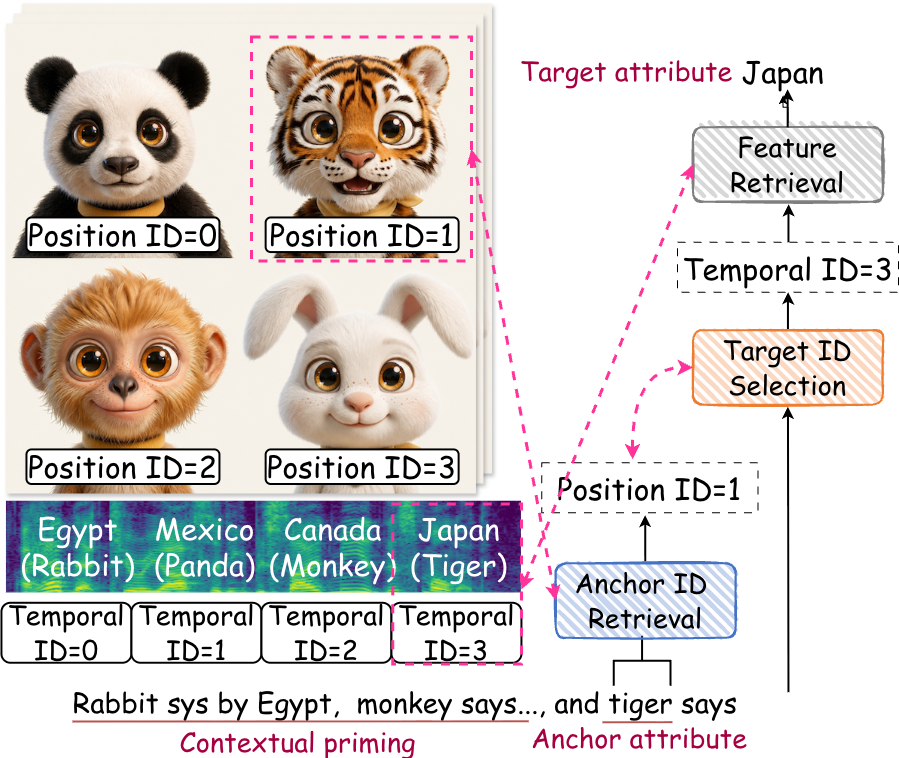}
    \caption{Illustration of trimodal symbolic binding mechanisms in VAAR task.}
    \label{fig:vaar}
\end{figure}

\cref{fig:vaar} illustrates how the proposed symbolic mechanism operates within the VAAR task. Unlike the AAVR task, the anchor attribute here is a visual attribute (e.g., the ``tiger''). Accordingly, during the anchor ID retrieval stage, the model retrieves the position ID, which serves as the binding ID for the visual modality. In the subsequent target ID selection stage, this position ID is mapped to its corresponding temporal ID (e.g., temporal ID = 1). Finally, during the feature retrieval stage, the model attends to the segment associated with this temporal ID to extract the target attribute's feature, which in this case is "Japan."

\subsection{Illustration of Causal Mediation Analysis for AAVR task}
\label{app:illu_cma}
\begin{figure}[htbp]
    \centering
    \includegraphics[width=1.0\textwidth]{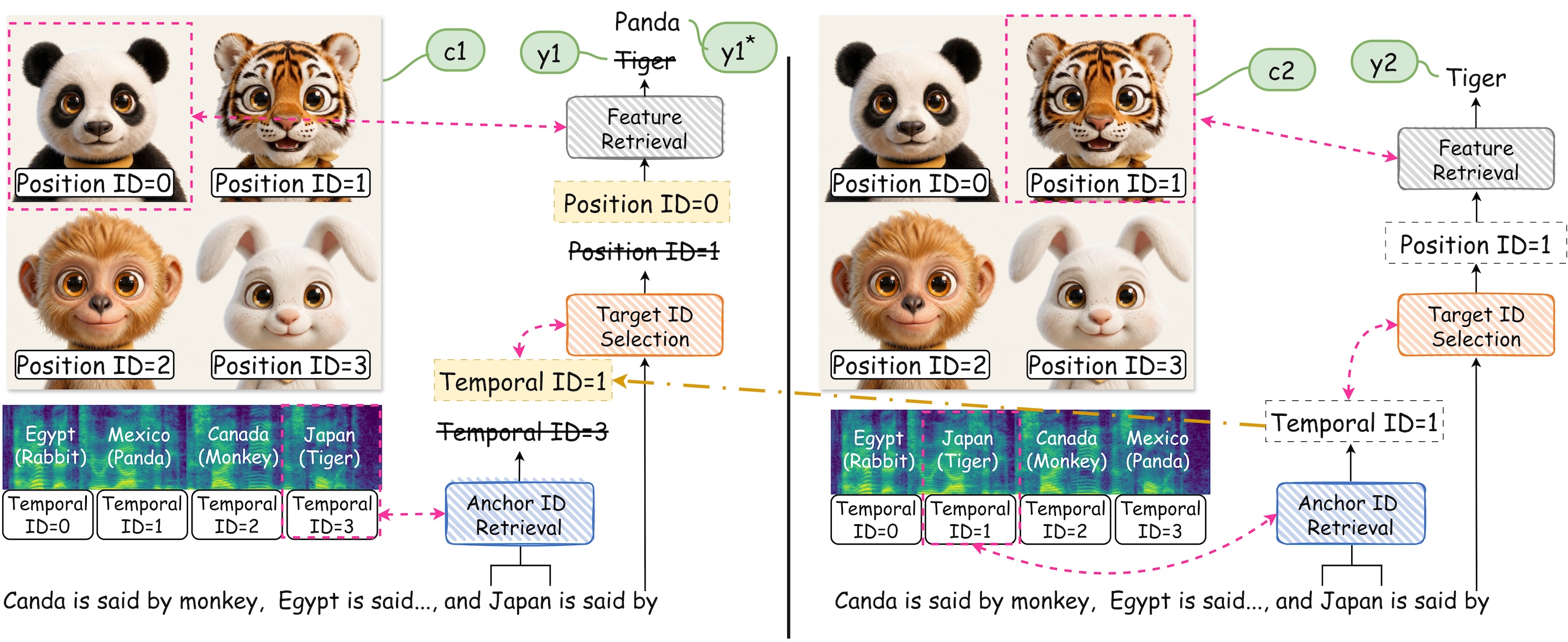}
    \caption{$c_2$=temporal ID manipulation, patching at the anchor ID retrieval stage}
    \label{fig:cma1_1}
\end{figure}

\begin{figure}[htbp]
    \centering
    \includegraphics[width=1.0\textwidth]{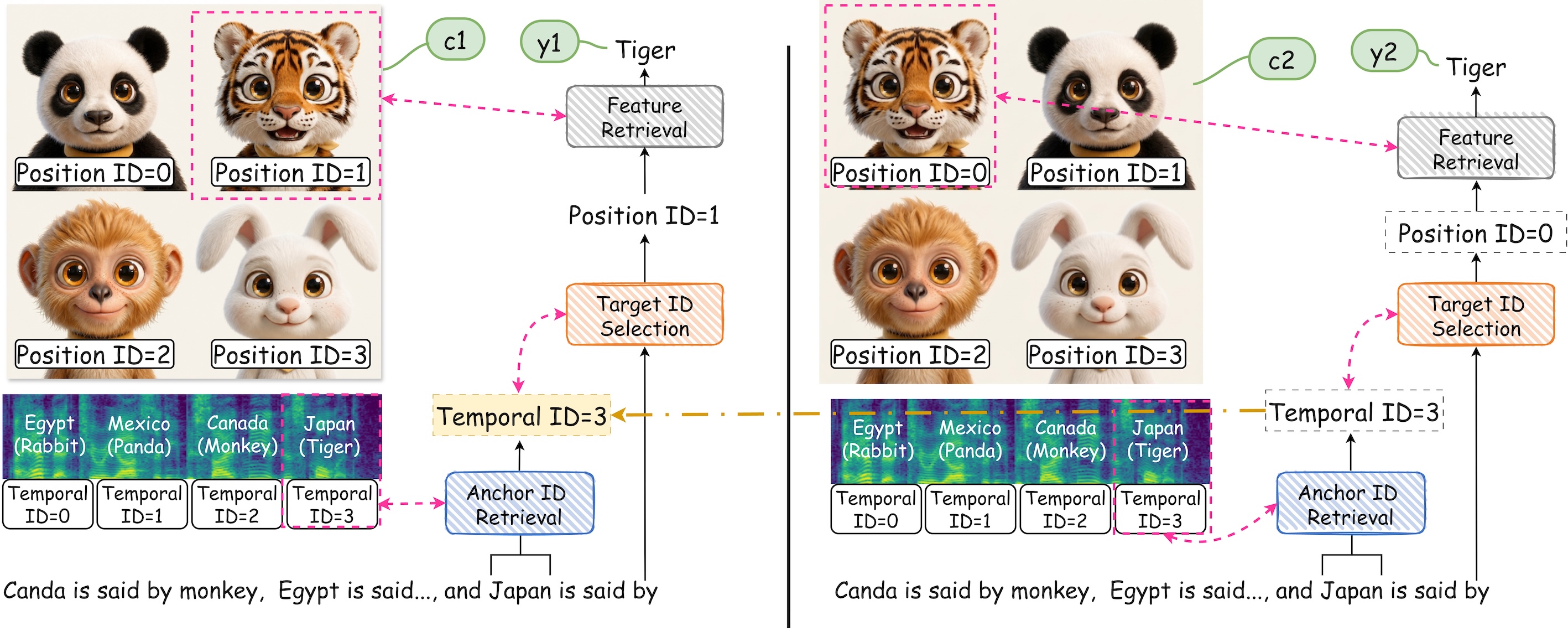}
    \caption{$c_2$=position ID manipulation, patching at the anchor ID retrieval stage}
    \label{fig:cma1_2}
\end{figure}

\begin{figure}[htbp]
    \centering
    \includegraphics[width=1.0\textwidth]{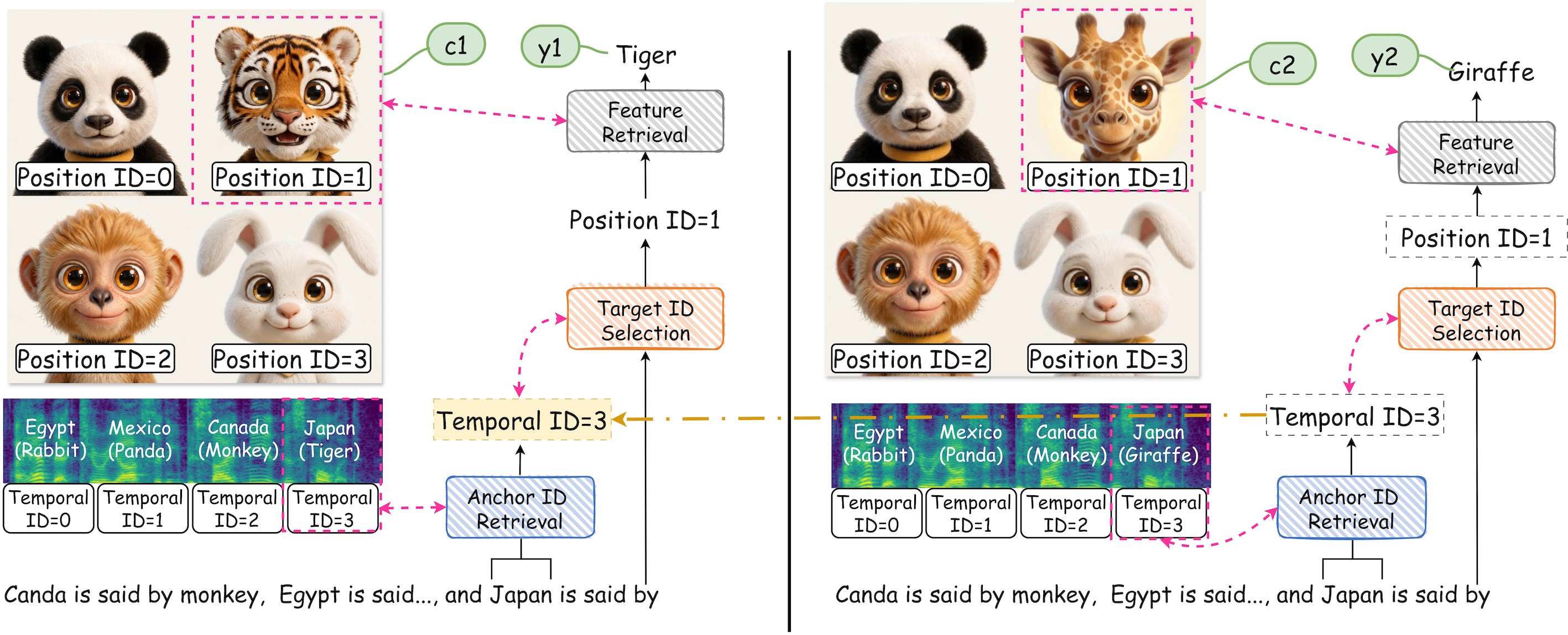}
    \caption{$c_2$=semantic content manipulation, Patching at the anchor ID retrieval stage}
    \label{fig:cma1_3}
\end{figure}

\begin{figure}[htbp]
    \centering
    \includegraphics[width=1.0\textwidth]{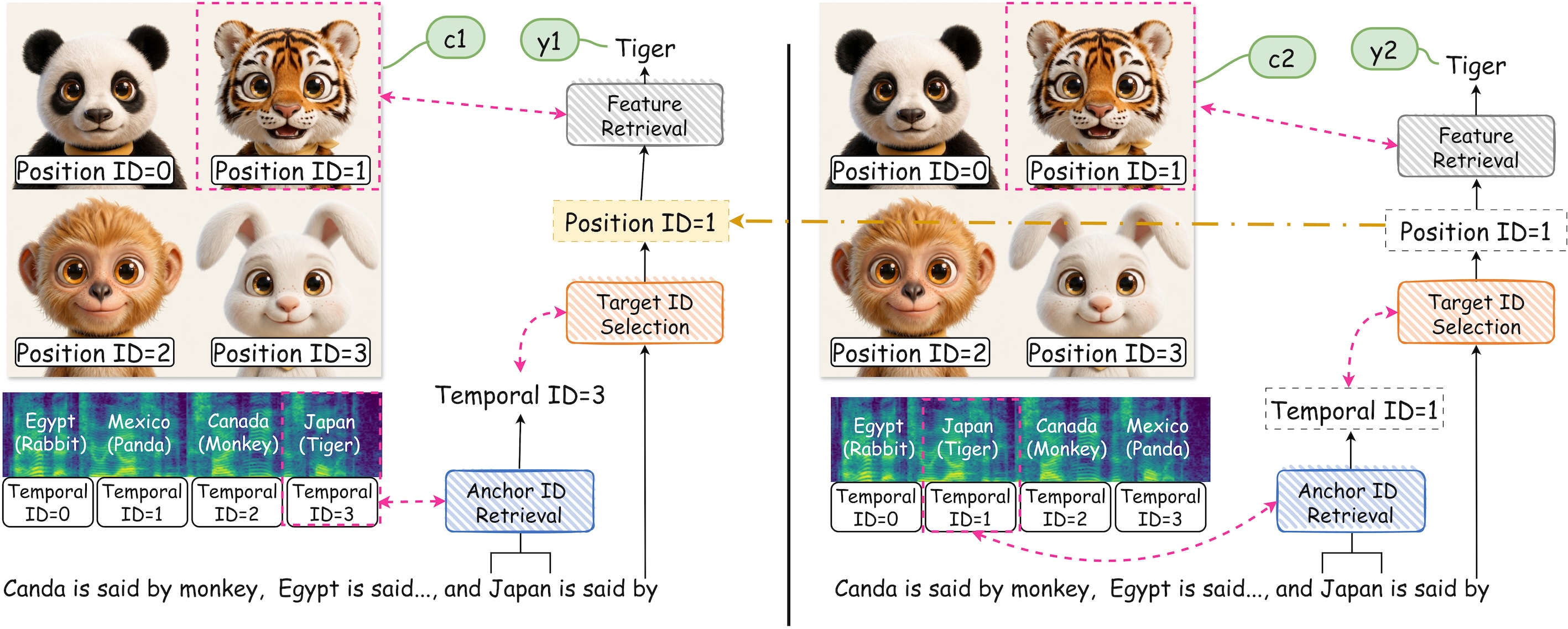}
    \caption{$c_2$=temporal ID manipulation, patching at the target ID selection stage}
    \label{fig:cma2_1}
\end{figure}

\begin{figure}[htbp]
    \centering
    \includegraphics[width=1.0\textwidth]{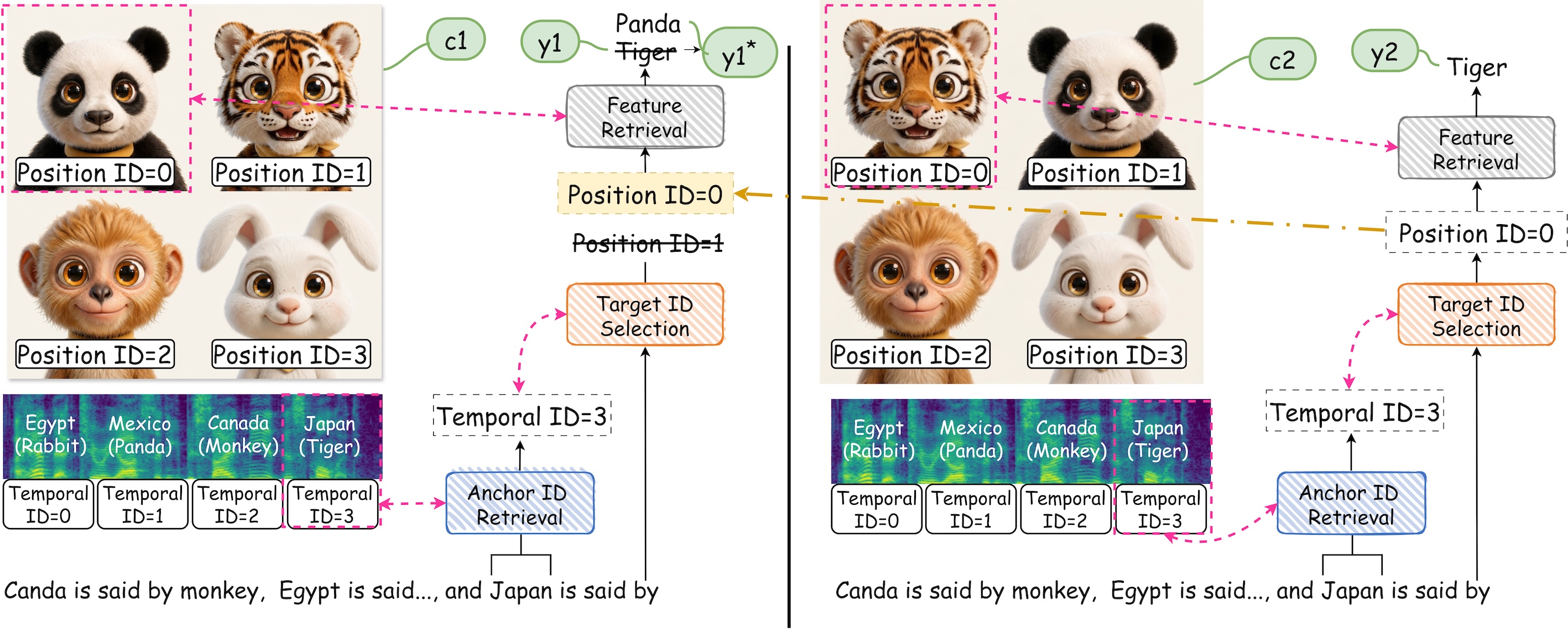}
    \caption{$c_2$=position ID manipulation, patching at the target ID selection stage}
    \label{fig:cma2_2}
\end{figure}

\begin{figure}[htbp]
    \centering
    \includegraphics[width=1.0\textwidth]{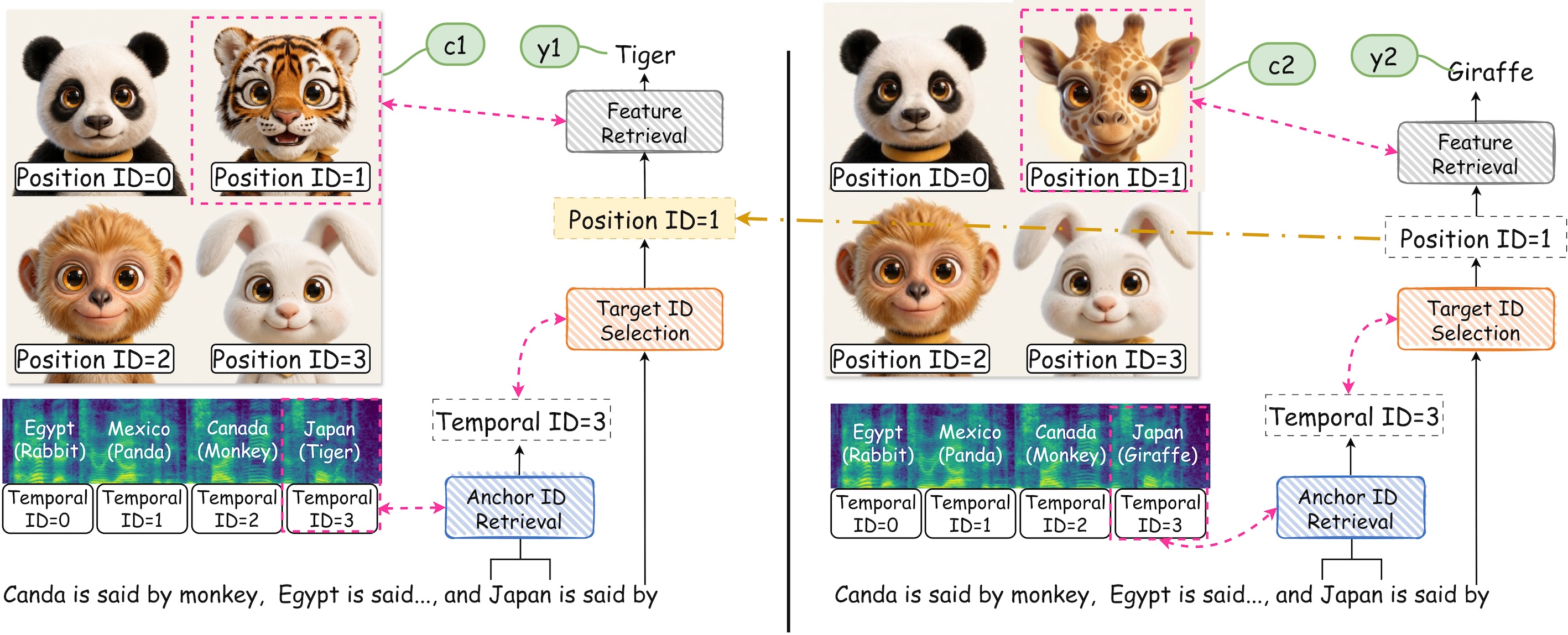}
    \caption{$c_2$=semantic content manipulation, patching at the target ID selection stage}
    \label{fig:cma2_3}
\end{figure}

\begin{figure}[htbp]
    \centering
    \includegraphics[width=1.0\textwidth]{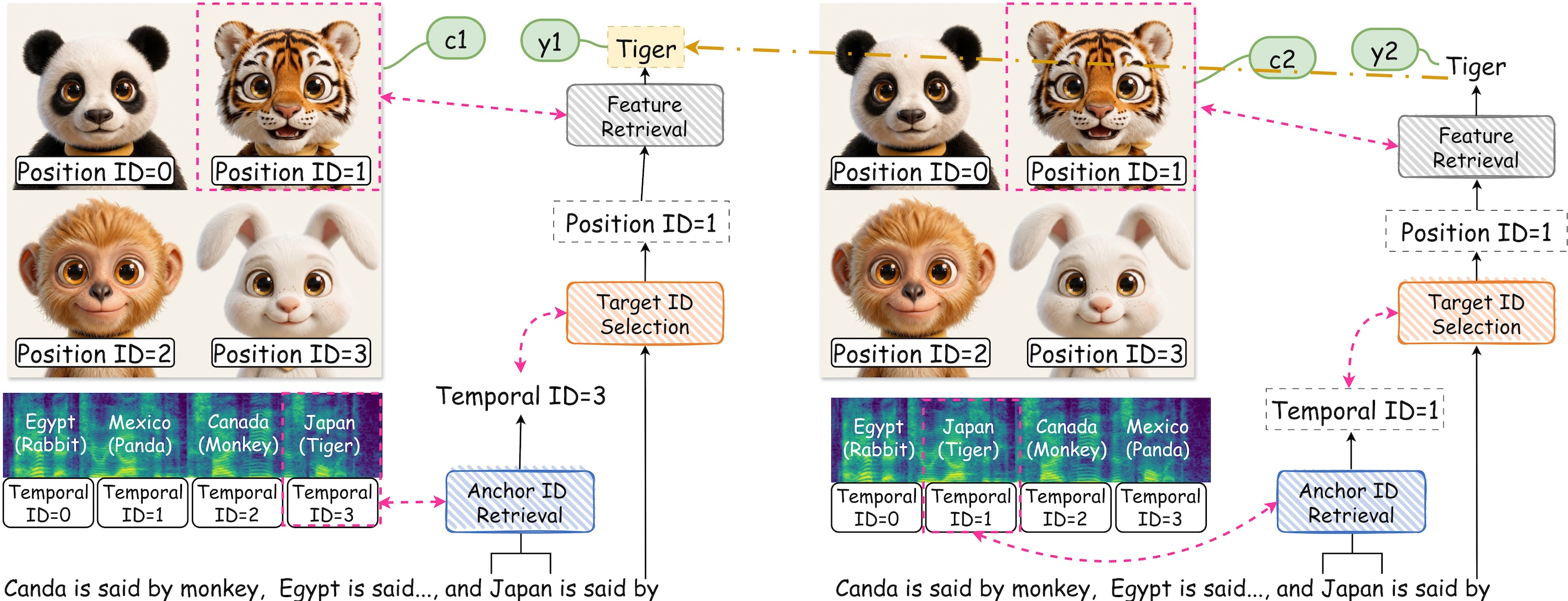}
    \caption{$c_2$=temporal ID manipulation, patching at the feature retrieval stage}
    \label{fig:cma3_1}
\end{figure}

\begin{figure}[htbp]
    \centering
    \includegraphics[width=1.0\textwidth]{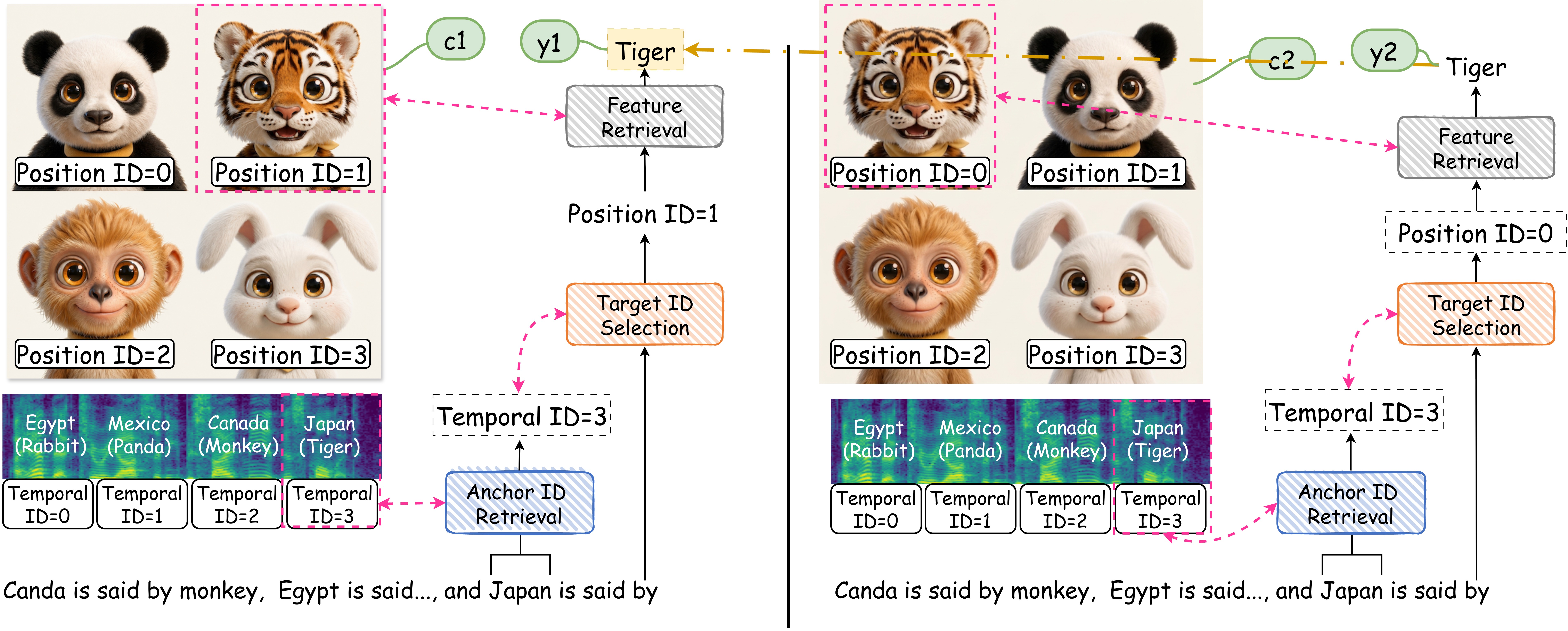}
    \caption{$c_2$=position ID manipulation, patching at the feature retrieval stage}
    \label{fig:cma3_2}
\end{figure}

\begin{figure}[htbp]
    \centering
    \includegraphics[width=1.0\textwidth]{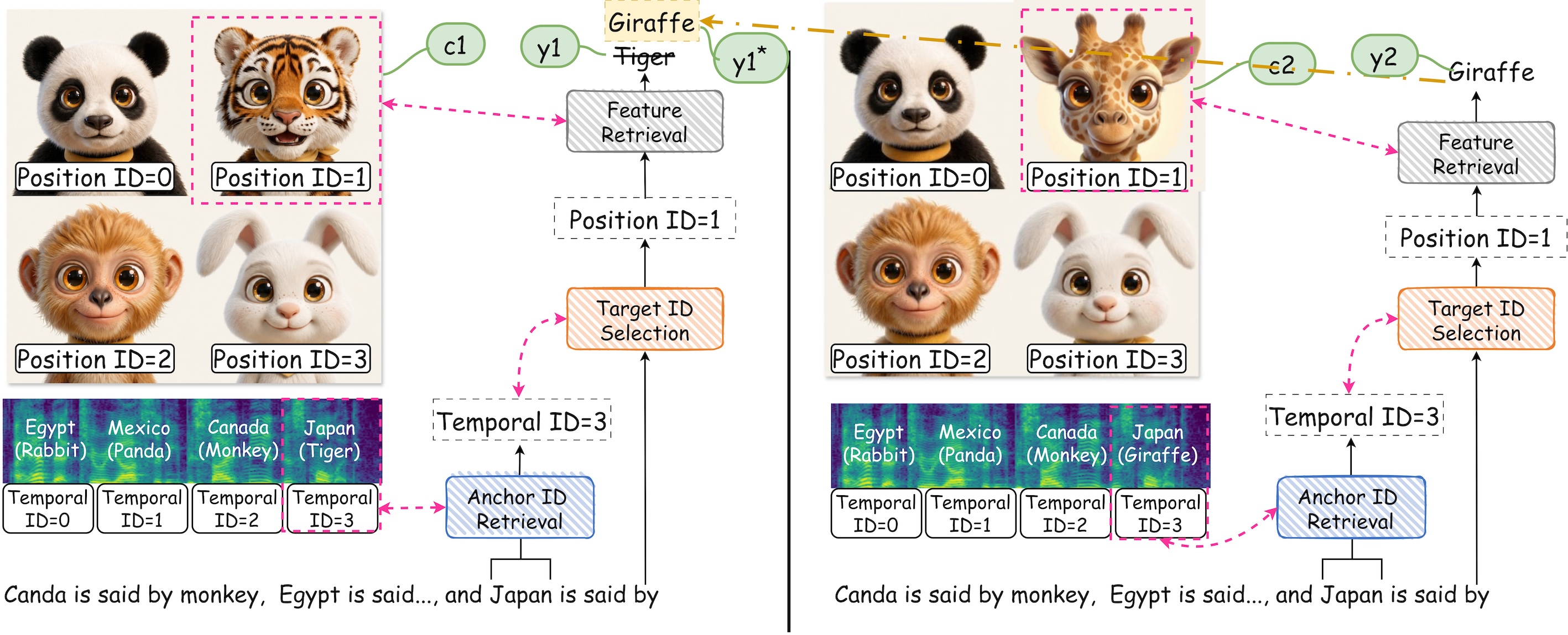}
    \caption{$c_2$=semantic content manipulation, patching at feature retrieval stage}
    \label{fig:cma3_3}
\end{figure}

The following figures illustrate the patching configurations used in our causal mediation analysis on the AAVR task. For each of the three-stage mechanisms-\emph{anchor ID retrieval}, \emph{target ID selection}, and \emph{feature retrieval}-we patch the residual stream with a $c_2$ obtained from one of three manipulations: temporal ID manipulation, position ID manipulation, or semantic content manipulation. This yields a $3 \times 3$ grid of patching configurations (\cref{fig:cma1_1}--\cref{fig:cma3_3}).

Our three-stage hypothesis makes a precise prediction about which manipulation should drive the model's output toward the counterfactual target $y_1^*$ at each stage. Specifically, we predict that:
\begin{itemize}
    \item Patching at the \textbf{anchor ID retrieval} stage (anchor attribute token at mid-late layers) flips the prediction to $y_1^*$ only when $c_2$ is constructed via \emph{temporal ID manipulation} (\cref{fig:cma1_1}), while the other two manipulations leave the output essentially unchanged (\cref{fig:cma1_2,fig:cma1_3}).
    \item Patching at the \textbf{target ID selection} stage (last prompt token at late layers) flips the prediction to $y_1^*$ only when $c_2$ is constructed via \emph{position ID manipulation} (\cref{fig:cma2_2}), with the other two manipulations being ineffective (\cref{fig:cma2_1,fig:cma2_3}).
    \item Patching at the \textbf{feature retrieval} stage (last prompt token at deepest layers) flips the prediction to $y_1^*$ only when $c_2$ is constructed via \emph{semantic content manipulation} (\cref{fig:cma3_3}), while temporal and position ID manipulations have no comparable effect (\cref{fig:cma3_1,fig:cma3_2}).
\end{itemize}

These predictions are consistent with the empirical results reported in the main text, supporting the view that each stage encodes a distinct, causally separable type of information.




\section{Qualitative Results}
\label{app:quali}
\begin{figure}[htbp]
    \centering
    \begin{subfigure}[t]{0.95\textwidth}
        \centering
        \includegraphics[width=\textwidth]{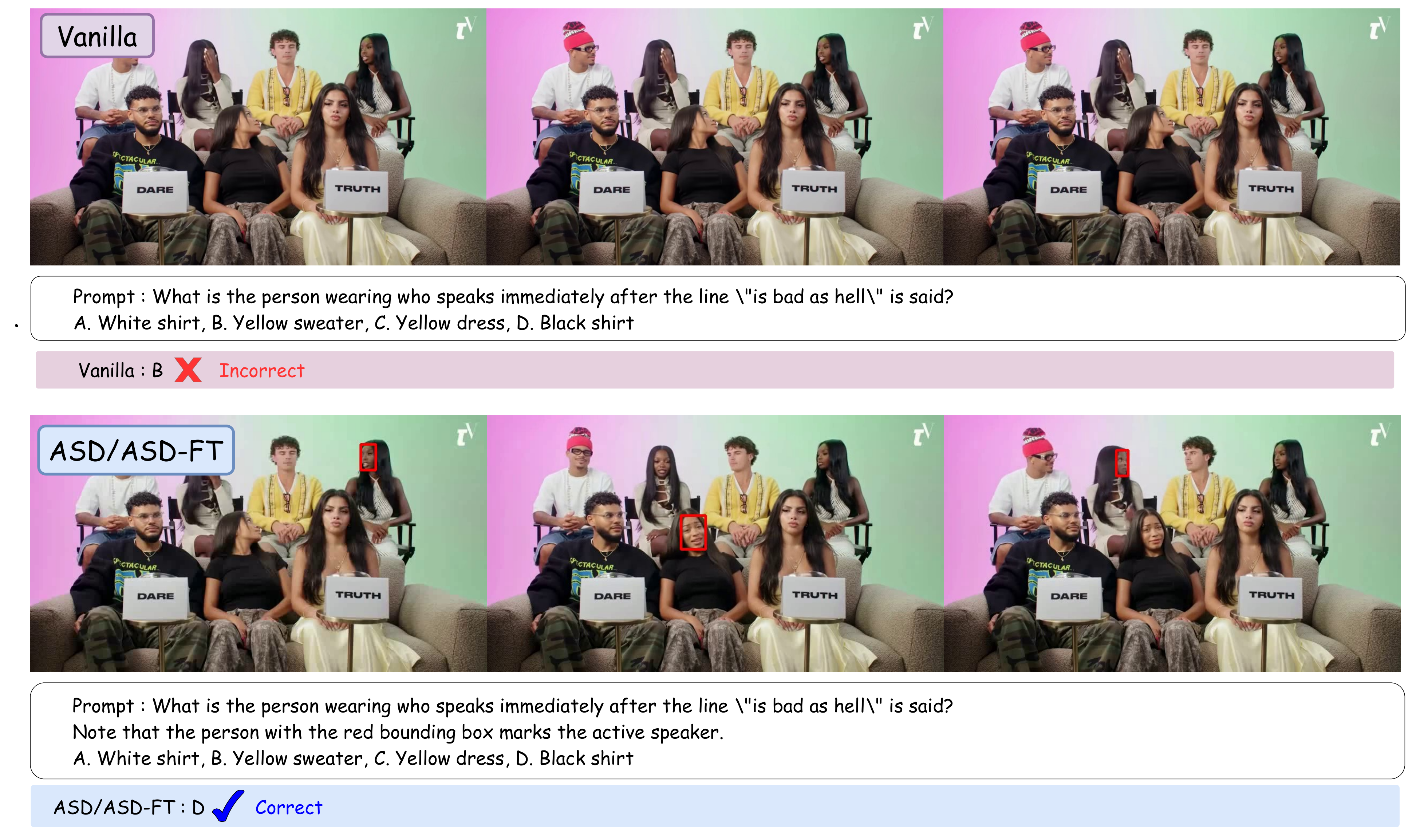}
        \label{fig:qual_qa1}
    \end{subfigure}
    \vspace{0.5em}
    \begin{subfigure}[t]{0.95\textwidth}
        \centering
        \includegraphics[width=\textwidth]{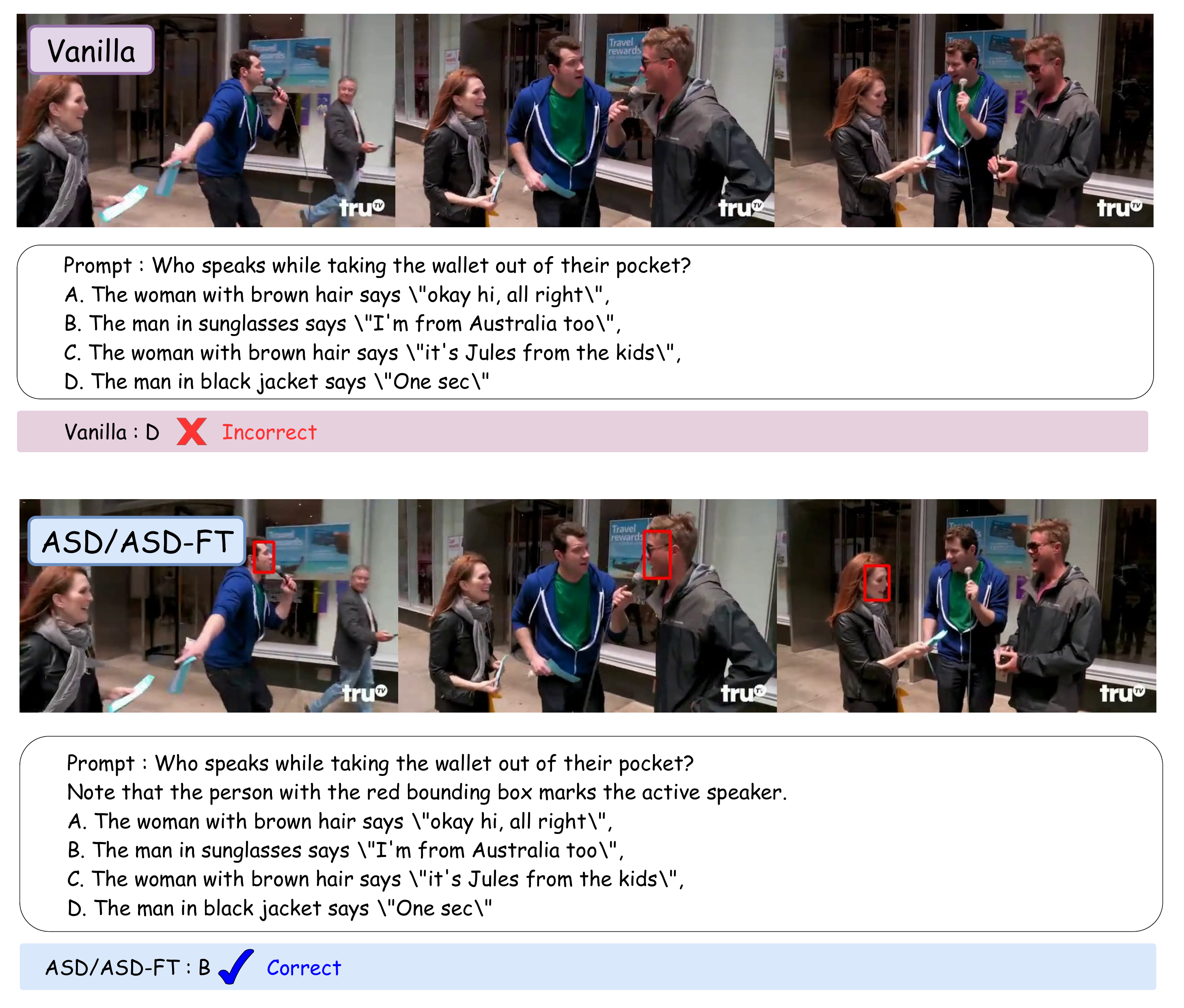}
        \label{fig:qual_qa2}
    \end{subfigure}
    \caption{\textbf{Qualitative comparison on AVSpeaker\cite{nguyen2025see} with Qwen2.5-Omni (7B).} The vanilla model selects the wrong answer, while the ASD/ASD-FT variant answers correctly.}
    \label{fig:qual_qa}
\end{figure}

\begin{figure}[htbp]
    \centering
    \includegraphics[width=0.85\textwidth]{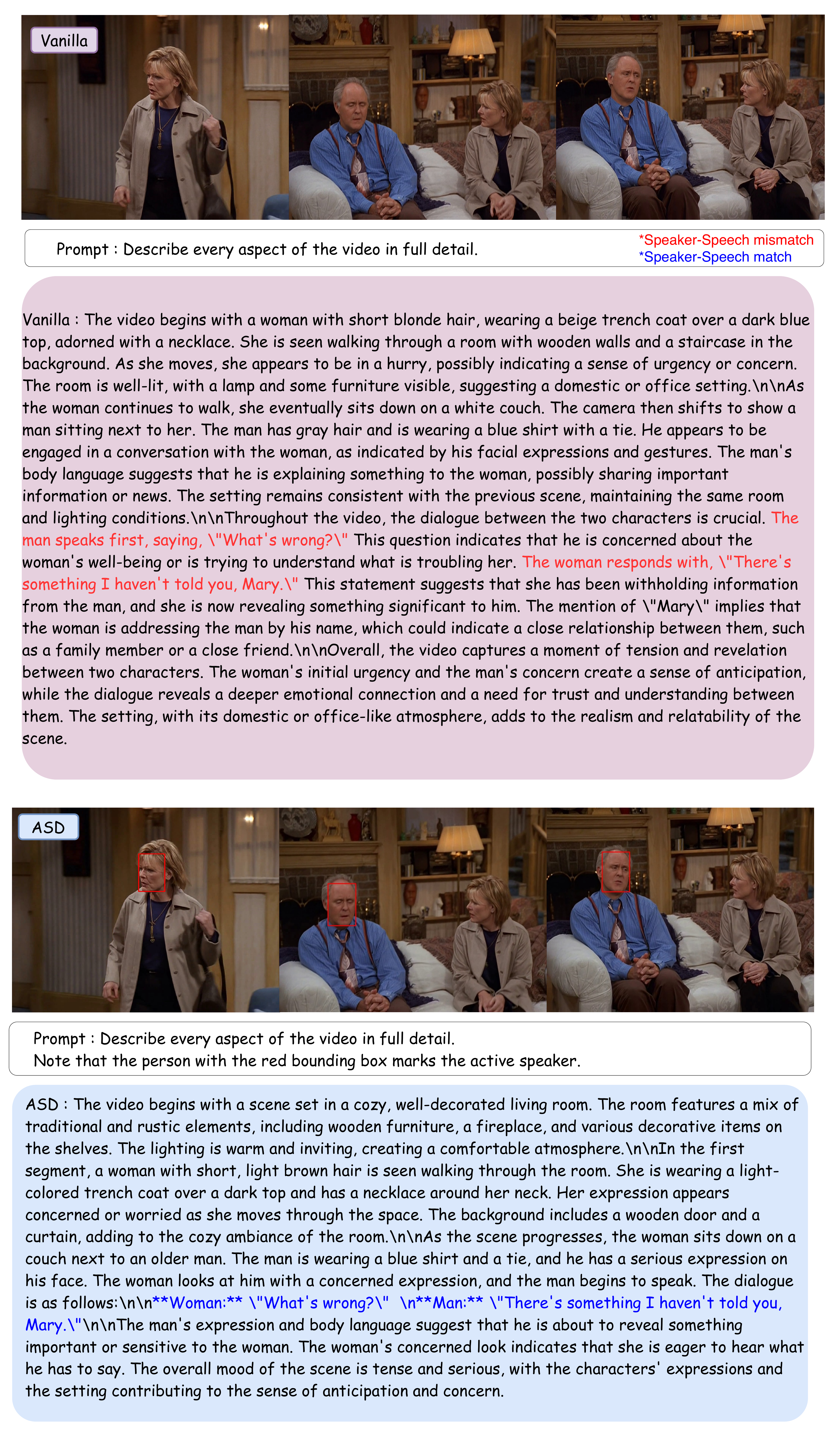}
    \caption{\textbf{Qualitative comparison on a multi-speaker captioning example from DiaDemBench~\cite{chen2026diadem} with Qwen2.5-Omni (7B).} The vanilla model produces speaker-speech mismatches, attributing utterances to the wrong characters in the scene. In contrast, the ASD variant generates a caption in which each utterance is consistently aligned with the correct speaker.}
    \label{fig:qual_caption}
\end{figure}

We present qualitative comparisons between the vanilla Qwen2.5-Omni (7B) and our ASD/ASD-FT variants on conversation-rich videos. As illustrated in \cref{fig:qual_qa}, the vanilla model frequently misattributes utterances to the wrong speaker on multi-speaker QA examples, reflecting the audio-visual binding bottleneck identified in our analysis. By overlaying a red bounding box on the active speaker, our method allows the model to correctly align each utterance with its corresponding speaker. This effect also extends to free-form captioning on DiaDemBench (\cref{fig:qual_caption}), where the ASD variant produces speaker--utterance pairings that are consistently faithful to the scene, demonstrating that the gains from audio-visual prompting generalize across both discriminative and generative settings.

\section{Limitations}
\label{app:limit}
While our study demonstrates generalizability to real-world datasets, our main analysis relies on synthetic toy datasets. This choice enables a controlled examination of the target phenomenon, and similar toy-dataset analyses have been widely adopted in prior work to isolate specific model behaviors~\cite{yang2025emergent, assouel2025visual}. Nevertheless, such datasets do not fully capture the diversity and complexity of real-world audiovisual scenes, and thus our conclusions should be viewed as a focused analysis of the phenomenon rather than an exhaustive evaluation across all practical conditions. Following~\cite{chen2026diadem}, we focus exclusively on the more challenging single-shot setting, where performance is typically worse than in multi-shot settings in which a single active speaker is often zoomed in per frame. As a result, we do not include an analysis of multi-shot scenarios. Additionally, while we propose a simple, efficient, and generalizable method that leverages off-the-shelf active speaker detection to mitigate the limitations of AVLLMs, our approach is not designed to modify or improve the intrinsic capabilities of the AVLLM itself. We hope that our findings serve as a stepping stone toward future work on training methodologies and model architectures that more directly address this issue.

\section{Computational Resource}
\label{app:compu}
We run most experiments on a machine equipped with an AMD EPYC 7513 32-core CPU and a single NVIDIA RTX A6000 GPU. For the WorldSense~\cite{hong2026worldsense} evaluation, which requires larger memory, we use an NVIDIA B6000 GPU.

\section{Social Impact}
\label{app:social}
Our work advances the ability of Audio-Visual LLMs to accurately resolve "who says what" in multi-speaker videos, which can benefit a range of socially valuable applications such as automatic captioning and transcription for the deaf and hard-of-hearing community, improved accessibility tools for educational and media content, and more reliable assistive technologies for analyzing conversational video. The mechanistic insights presented in this paper also contribute to the broader interpretability literature, helping researchers and practitioners better understand and diagnose failure modes in multimodal systems-an important step toward building more transparent and trustworthy AI. However, since our method depends on an off-the-shelf Active Speaker Detection model, any biases or demographic disparities in that module could propagate into downstream predictions, potentially affecting certain populations unequally. We encourage future work to pair these capabilities with appropriate consent frameworks, fairness audits across demographic groups, and dataset documentation practices to mitigate these risks.

\end{document}